\documentclass{article}

\usepackage{arxiv}

\usepackage[utf8]{inputenc} 
\usepackage[T1]{fontenc}    
\usepackage{hyperref}       
\usepackage{url}            
\usepackage{booktabs}       
\usepackage{amsfonts}       
\usepackage{nicefrac}       
\usepackage{microtype}      
\usepackage{graphicx}

\usepackage{mathtools,amssymb} 
\usepackage{textcomp,marvosym} 
\usepackage{bm, scalerel}
\usepackage{subcaption}

\DeclareMathOperator*{\E}{\mathbb{E}}

\usepackage[colorinlistoftodos]{todonotes}
\usepackage{algorithmic}
\usepackage{algorithm}

\usepackage{setspace}

\title{Dynamic planning in hierarchical active inference}

\author{
	Matteo Priorelli \\
	Institute of Cognitive Sciences and Technologies\\
	National Research Council of Italy, Padova\\
	Sapienza University of Rome, Italy \\
	\texttt{matteo.priorelli@gmail.com} \\
	\And
	Ivilin Peev Stoianov \\
	Institute of Cognitive Sciences and Technologies\\
	National Research Council of Italy\\
	Padova, Italy \\
	\texttt{ivilinpeev.stoianov@cnr.it} \\
}

\begin{document}
	
	\maketitle
	\setstretch{1.1}
	
	
	
	
	\begin{abstract}
		By dynamic planning, we refer to the ability of the human brain to infer and impose motor trajectories related to cognitive decisions. A recent paradigm, \textit{active inference}, brings fundamental insights into the adaptation of biological organisms, constantly striving to minimize prediction errors to restrict themselves to life-compatible states. Over the past years, many studies have shown how human and animal behaviors could be explained in terms of active inference -- either as discrete decision-making or continuous motor control -- inspiring innovative solutions in robotics and artificial intelligence. Still, the literature lacks a comprehensive outlook on effectively planning realistic actions in changing environments. Setting ourselves the goal of modeling complex tasks such as tool use, we delve into the topic of dynamic planning in active inference, keeping in mind two crucial aspects of biological behavior: the capacity to understand and exploit affordances for object manipulation, and to learn the hierarchical interactions between the self and the environment, including other agents. We start from a simple unit and gradually describe more advanced structures, comparing recently proposed design choices and providing basic examples. This study distances itself from traditional views centered on neural networks and reinforcement learning, and points toward a yet unexplored direction in active inference: hybrid representations in hierarchical models.
	\end{abstract}
	
	
	\section{Introduction}
	
	Three characteristics of the brain are relevant to tasks involving planning in changing environments, such as tool use. First, the brain's ability to maintain estimates not only of bodily states, but also of external physical variables in relation to the self. Studies have shown how the Posterior Parietal Cortex (PPC) of the monkey brain encodes objects with sensorimotor representations reflecting the body structure \cite{Breveglieri2015,Romero2014}. These representations are extremely useful for object manipulation since they account efficiently for the action possibilities provided by the object, also known as \textit{affordances} \cite{Yamanobe2017}. For instance, one might encode a cup in different ways (power versus precision grips) based on whether one wants to throw it or drink from it. Further, to act timely in a dynamic environment, the PPC can encode multiple objects in parallel during sequences of actions, even when there is a considerable delay between different subgoals \cite{Baldauf2008}.
	
	A second characteristic regards flexible and deep hierarchies. Hierarchical structures are so pervasive that they not only exist as causal relationships between physical properties of the environment, but are also inherent to how biological organisms act over it. Even the most complex kinematic structures of animals follow a rigid hierarchical strategy, whereby different limbs propagate from a body-centered reference frame. The hierarchical modularity of brain functional networks is widely recognized \cite{Meunier2009,Hilgetag2020}, as well as the representation of the body schema in somatosensory and motor areas \cite{Holmes2004}, and the organization of hierarchical motor sequences concerning parietal and premotor cortices \cite{Yokoi2019}. In particular, the body schema is not a static entity but changes in concurrence to the development of the human body during childhood and adolescence \cite{assaiante}. Surprisingly, the nervous system is able to relate external objects to the self in a way that, although not reflecting the actual causal relationships between the body and the environment, is the most suitable for better operating in a specific context. Physiological studies have demonstrated that, with extensive tool use, parietal and motor areas of the monkey brain gradually adapt to make room for the tool, increasing the length of the perceived limb \cite{lriki1996,Obayashi2001}. This adaptation is highly plastic, assimilating objects in a very short time \cite{Carlson2010} and inducing altered somatosensory representations of the body morphology that persist even after tool use \cite{Cardinali2009}.
	
	A third characteristic is the ability to construct a dynamic discretized plan based on continuous sensory evidence. Complex tasks involve decision-making, which the brain is known to realize via several methods \cite{Pezzulo2019planning}. Among them, one is particularly relevant: planning for deliberation, also known as \textit{vicarious trials and errors}, whereby an action is selected after several alternatives have been generated and evaluated \cite{Redish2016}. One of the most intriguing aspects of human planning is the capacity to imagine, or endogenously generate dynamic representations of future states, including potential trajectories and subgoals that bring to such states \cite{Stoianov2018a, Stoianov2022}. The hippocampus is a key neural structure known to support trajectory generation, although planning is accomplished in concert with other areas \cite{Redish2016}.
	
	How does the human brain capture the hierarchical organization and dynamics of the self and the environment to afford purposeful planning? One recent theory is that of \textit{predictive coding} \cite{Rao1999,Hohwy2013,Clark2016,Friston2009}, which has been attracting increasing interest in recent years and proposes itself as a unifying paradigm of cortical function. According to predictive coding, living beings make sense of the world by building an internal generative model that imitates the causal relationships of the external generative process. From a high-level hypothesis about the world, a cascade of neural predictions takes place, eventually leading to a low-level guess about sensory evidence. Comparing the model's guess with the sensorium triggers another cascade of \textit{prediction errors} that travel back to the deepest cortical levels. The model iteratively refines its structure until all the prediction errors are minimized, that is, until it correctly predicts what will happen next. This optimization differs from the more traditional view of deep learning, in that the message passing is local and what climbs up the hierarchy does not signal the detection of a feature, but how much the model is \textit{surprised} about its prediction. Besides having stimulated cognitive and neural studies under several circumstances \cite{Hohwy2020,Clark2013,Shipp2016,millidge2022predictive}, this theory has also influenced novel directions in machine learning: Predictive Coding Networks (PCNs) have been shown to generalize well to classification or regression tasks \cite{Ororbia2022b,salvatori2023braininspired}, with key advantages compared to neural networks and still approximating the backpropagation algorithm \cite{Whittington2017, Whittington2019, Millidge2022}.
	
	While predictive coding can elucidate illusions and visual phenomena such as binocular rivalry \cite{Hohwy2008}, it explains just the first (perceptual) half of the story. More specifically, it does not explain why interactions with the environment occur -- a process that results, considering the above example, in the monkey brain actively distorting its body schema during tool use. On this trail, a second innovative perspective has been proposed, aspiring to unveil a unified first principle not just on cortical function, but on the behavior of all living organisms. This perspective, called \textit{active inference} \cite{Friston2010, Friston2010a, Buckley2017, Parr2021}, is grounded on the same theoretical basis of predictive coding but further assumes two key aspects of biological behavior. First, that a living being does not maintain static hypotheses about the world but also constructs internal dynamics -- either as instantaneous trajectories or future states -- to anticipate the unfolding of events occurring at different timescales. Second, that these dynamic hypotheses can be fulfilled by movements. The latter assumption replaces models with \textit{agents}, conveying a somewhat counterintuitive but insightful implication: while perception lets the agent's hypothesis conform to the environment (as in predictive coding), action forces the environment to conform to the hypothesis -- by sampling those observations that make the hypothesis true. If such hypotheses or \textit{beliefs} correspond to desired states defined, e.g., by the phenotype, cycling between action and perception ultimately allows the agent to survive. This is the core of the so-called \textit{free energy principle}, which states that in order to maintain homeostasis, all organisms must constantly and actively minimize the difference between their sensory states and their expectations based on a small set of life-compatible options. Giving a practical example, if I believe to find myself with a tool in hand, I will try with all my strength to observe visual images of the tool in my hand; in doing this, a combined reaching and grasping action happens. This view distances itself from the stimulus-response mapping widely established in neuroscience, and evidence indicates that it could be more biologically plausible than optimal control and Reinforcement Learning (RL) \cite{Friston2009a,Friston2011opt, Adams2013,Brown2011}.
	
	Active inference implementations can be divided into two frameworks, which have been used to simulate human and animal behaviors under the two complementary aspects of motor control \cite{Priorelli2023g, Lanillos2021x, Taniguchi2023, Pezzato2020, mlp22, mannella2021active, AnilMeera2022} and decision-making \cite{Kaplan2018, Adams2013b,Friston2017, Proietti2023,Donnarumma2017}. In principle, active inference might be key for understanding how goal-directed behavior emerges in the human brain \cite{pezzulo2023neural}. For instance, relevant objects used for manipulation may gradually become part of one's identity through a closed loop between motor commands and sensory evidence, meaning that the boundary of the self from the environment increases whenever the agent predicts the consequences of its own movements \cite{Lanillos2020}. Additionally, active inference might lead to key advances with current artificial agents, taking forward a promising research area known as \textit{planning as inference} \cite{Toussaint2006,Toussaint2009,Botvinick2012}. The three characteristics delineated above are fundamental to designing active inference agents that can tackle real-life applications such as tool use. But how to combine them into a single view? In other words, how to perform dynamic planning with hierarchical structures of several objects?
	
	To answer this question, in this study we explore an alternative direction in active inference, i.e., toward hybrid computations in hierarchical systems. We analyze many design choices that have been applied in the motor control domain, with an in-depth look at object affordances, deep hierarchies, and planning with continuous signals. Asking ourselves how to model tool use, we start from a simple unit and construct richer modules that can be linked in a hierarchical fashion, exhibiting interesting high-level features. In Chapter 2, we consider a single-DoF agent and explore how to account for affordances and realize a multi-step behavior in continuous time only. In Chapter 3, we analyze the implications of combining different units in a single network, using more complex kinematic configurations and distinguishing between intrinsic and extrinsic dynamics. In Chapter 4, we describe the advantages of using discrete decision-making in continuous environments, focusing on hybrid structures and drawing some parallelisms between the two worlds. Finally, in the Discussion we elaborate on the benefits of addressing discrete and continuous representations together, and give a few suggestions for future work on this subject.
	
	
	\section{Modeling affordances}
	
	In this chapter, we explain the inference mechanisms of a basic unit in continuous time. We then discuss one by one the changes and features that we introduce, in order to achieve a multi-step behavior in simple tasks that do not require deep hierarchical modeling nor online replanning.
	
	The continuous-time active inference framework \cite{Friston2008,Friston2009a,Priorelli2023g} -- generally compared to the low-level sensorimotor loops -- makes use of generalized filtering \cite{Friston2010gen} to model instantaneous trajectories of the self and the environment; these trajectories are inferred by minimization of a quantity called \textit{variational free energy}, which is the negative of what in machine learning is known as the \textit{evidence lower bound}. Differently from optimal control, motor commands in active inference derive from proprioceptive predictions that are fulfilled by classical spinal reflex arcs \cite{Adams2013}. This eliminates the need for cost functions -- as the inverse model maps from proprioceptive (and not latent) states to actions -- and replaces a control problem with an inference problem \cite{Friston2011opt}.
	
	Modeling of objects in active inference has been recently done in the context of active object reconstruction \cite{ferraro2022disentangling, vanbergen2022objectbased,VandeMaele2022,vandemaele2023objectcentric} -- where an agent encoded independent representations for multiple elements, and used action to more accurately infer its dynamics; for simulating oculomotor behavior \cite{Adams2015b} -- where the dynamics of a target belief was biased by a hidden location; or for analyzing epistemic affordance \cite{Donnarumma2017}, i.e., the changes in affordance of different objects in relation to the agent's beliefs. In continuous time, such affordances can be expressed in intrinsic reference frames corresponding to potential agent's configurations, defining specific ways to interact with the objects. Manipulating these additional beliefs depending on the agent's intentions \cite{Priorelli2023} permits effectively operating in dynamic contexts, e.g., tracking a target with the eyes \cite{Adams2015b}, or grasping an object on the fly \cite{Priorelli2023e} and placing it at a goal position \cite{Priorelli2023d}.
	
	
	\subsection{\label{sec:reaching}A simple agent}
	
	\begin{figure}[h]
		\begin{subfigure}{0.6\textwidth}
			\centering
			\includegraphics[width=\textwidth]{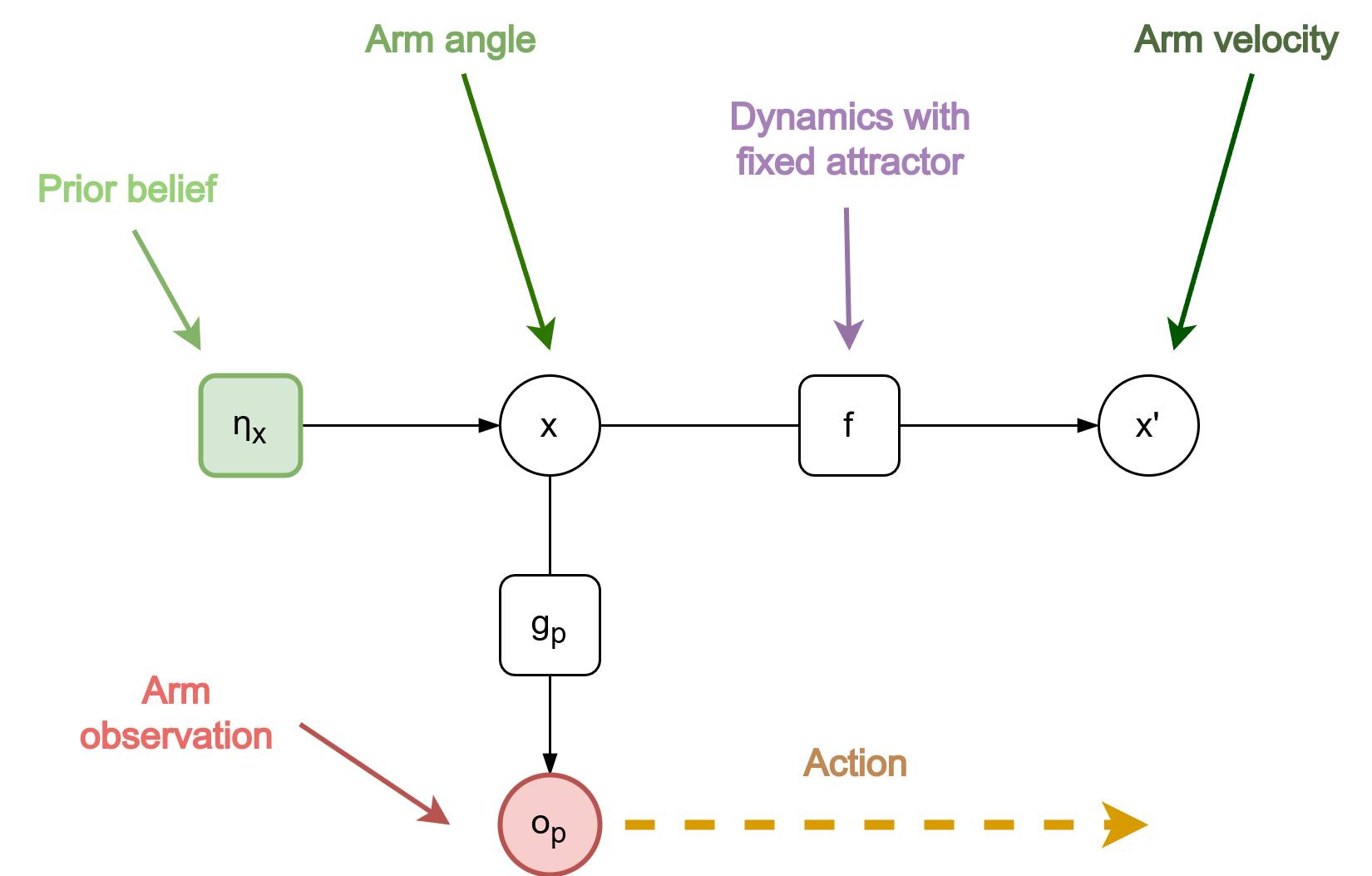}
			\caption{}
			\label{fig:1st_solution_new}
		\end{subfigure}
		\hfill
		\begin{subfigure}{0.32\textwidth}
			\centering
			\includegraphics[width=\textwidth]{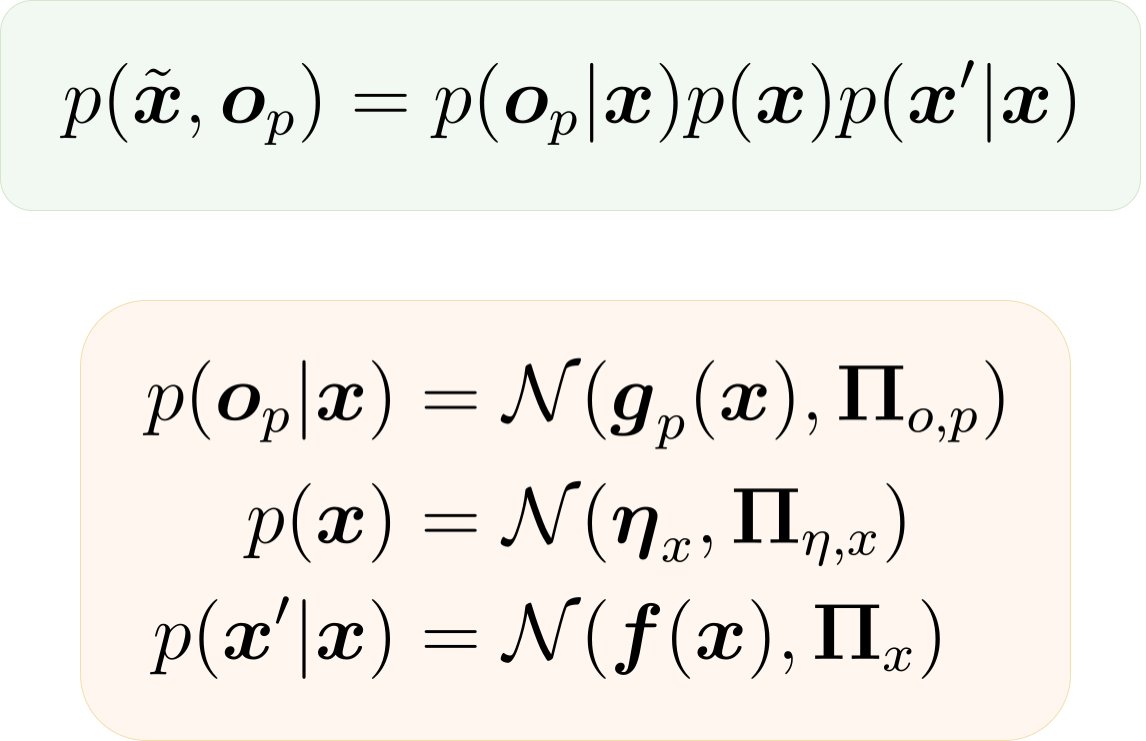}
			\vspace{3em}
			\caption{}
			\label{fig:gen_model_reaching}
		\end{subfigure}
		\caption{\textbf{(a)} Factor graph of a basic unit for static reaching. Variables and factors are indicated by circles and squares, respectively. Hidden states $\bm{x}$ (e.g., the arm angle) generate observations $\bm{o}_p$ (e.g., the arm proprioception) through the likelihood function $\bm{g}_p$, and their 1st derivatives $\bm{x}^\prime$ (e.g., the arm velocity) through a dynamics function $\bm{f}$. In contrast to optimal control, here action follows observation prediction errors arising from a simple attractor $\bm{\rho}$ embedded in the model dynamics, or from a prior belief $\bm{\eta}_x$ over the arm angle. \textbf{(b)} Agent's generative model.}
	\end{figure}
	
	The most elementary unit is represented in Figure \ref{fig:1st_solution_new}. This is the simplest formulation of a continuous-time active inference agent, where we kept only the key nodes. This allows us to easily describe a velocity-controlled dynamic system with the following likelihood $\bm{g}_p$ and dynamics $\bm{f}$:
	\begin{align}
		\begin{split}
			\bm{o}_p &= \bm{g}_p(\bm{x}) + \bm{w}_{o,p} \\
			\bm{x}^\prime &= \bm{f}(\bm{x}) + \bm{w}_x
		\end{split}
	\end{align}
	where $\bm{x}$ and $\bm{o}_p$ are respectively called hidden states and observations (the subscript $p$ indicates the proprioceptive domain), and the letter $w$ indicates noise terms sampled from Gaussian distributions. For simplicity, we considered just two temporal orders -- although all the features we elucidate in the following hold for a system of generalized coordinates \cite{Friston2010gen} -- and we defined a likelihood function only for a single temporal order. We assume that the corresponding generative model is factorized as in Figure \ref{fig:gen_model_reaching}, expressed in terms of \textit{precisions} (or inverse variances) $\bm{\Pi}$. Note that we introduced a prior $\bm{\eta}_x$ over the hidden states, which is not generally used in continuous-time formulations, but it is the key element connecting different levels in discrete-time active inference \cite{Friston2017a} or PCNs \cite{millidge2022predictive} -- as will be explained later. Also note that we used a generalized notation for instantaneous trajectories or paths, i.e., $\tilde{\bm{x}} = [\bm{x}, \bm{x}']$, where $\bm{x}$ will be indicated in the following as the 0th order, and $\bm{x}'$ as the 1st order.	We highlighted in green and red respectively the input and output of the unit, namely the prior $\bm{\eta}_x$ and the observations $\bm{o}_p$.
	
	Exact computation of the posterior $p(\tilde{\bm{x}} | \bm{o}_p)$ is unfeasible since the evidence requires marginalizing over every possible outcome, i.e., $p(\bm{o}_p)=\int{p(\tilde{\bm{x}}, \bm{o}_p) d\tilde{\bm{x}}}$. For this reason, estimation of hidden states $\tilde{\bm{x}}$ is carried out through a variational approach \cite{Blei2017}, e.g., by minimizing the difference between a properly chosen approximate posterior $q(\tilde{\bm{x}})$ and the true posterior. This difference is expressed in terms of a Kullback-Leibler (KL) divergence:
	\begin{equation} 
		\label{eq:KL}
		D_{KL}[q(\tilde{\bm{x}})||p(\tilde{\bm{x}}|\bm{o}_p)] = \int_{\tilde{\bm{x}}} q(\tilde{\bm{x}}) \ln \frac{q(\tilde{\bm{x}})}{p(\tilde{\bm{x}} | \bm{o}_p)} d\tilde{\bm{x}}
	\end{equation}
	The denominator $p(\bm{x}|\bm{o}_p)$ still depends on the marginal $p(\bm{o}_p)$, but the KL divergence can be rewritten in terms of the log evidence and a quantity known as the \textit{free energy} $\mathcal{F}$:
	\begin{equation}
		\label{eq:free_energy}
		\mathcal{F} = \E_{q(\tilde{\bm{x}})} \left[ \ln \frac{q(\tilde{\bm{x}})}{p(\tilde{\bm{x}},\bm{o}_p)} \right] = \E_{q(\tilde{\bm{x}})} \left[\ln \frac{q(\tilde{\bm{x}})}{p(\tilde{\bm{x}}|\bm{o}_p)} \right] - \ln p(\bm{o}_p)
	\end{equation}
	Since the KL divergence is always nonnegative, the free energy provides an upper bound on surprise, i.e., $\mathcal{F} \geq \ln p(\bm{o}_p)$. Hence, minimizing $\mathcal{F}$ achieves the dual objective of keeping surprise low while estimating the true distribution. Assuming that the approximate posterior can be factorized into independent contributions, and further assuming that each contribution is Gaussian -- i.e., $\bm{q}(\tilde{\bm{x}}) = \mathcal{N}(\tilde{\bm{\mu}}_x, \tilde{\bm{P}}_x)$, with generalized means $\tilde{\bm{\mu}}_x$ and precisions $\tilde{\bm{P}}_x$ -- the optimization process breaks down to the minimization of (precision-weighted) \textit{prediction errors} in terms of the approximate posterior -- see \cite{Parr2021} for more details:
	\begin{align}
		\begin{split}
			\label{eq:pred_error}
			\bm{\varepsilon}_{o,p} &= \bm{o}_p - \bm{g}_p(\bm{\mu}_x) \\
			\bm{\varepsilon}_{\eta,x} &= \bm{\mu}_x - \bm{\eta}_x \\
			\bm{\varepsilon}_x &= \bm{\mu}_x^\prime - \bm{f}(\bm{\mu}_x)
		\end{split}
	\end{align}
	Then, the inference of the means $\tilde{\bm{\mu}}_x = [\bm{\mu}_x, \bm{\mu}_x']$ (also called \textit{beliefs}) of the posterior over the hidden states is reduced to the following message passing:
	\begin{equation}
		\label{eq:hidden_update}
		\dot{\tilde{\bm{\mu}}}_x = \begin{bmatrix} \dot{\bm{\mu}}_x \\ \dot{\bm{\mu}}^{\prime}_x \end{bmatrix} = \mathcal{D} \tilde{\bm{\mu}}_x - \partial_{\tilde{x}} \mathcal{F} = \begin{bmatrix} \bm{\mu}_x^\prime - \bm{\Pi}_{\eta,x} \bm{\varepsilon}_{\eta,x} + \partial_{x} \bm{g}_p^T \bm{\Pi}_{o,p} \bm{\varepsilon}_{o,p} + \partial_{x} \bm{f}^T \bm{\Pi}_x \bm{\varepsilon}_x \\ \\ - \bm{\Pi}_x \bm{\varepsilon}_x \end{bmatrix}
	\end{equation}
	where $\mathcal{D}$ is an operator that shifts every derivative by one, i.e., $\mathcal{D} \tilde{\bm{\mu}}_x = [\bm{\mu}_x^\prime, \bm{0}]$. This term arises because the generative model maintains a belief not over a static point, but over a dynamic trajectory, and only when the motion of the mean $\dot{\tilde{\bm{\mu}}}_x$ equals the mean of the motion $\mathcal{D} \tilde{\bm{\mu}}_x$, is the free energy minimized. In short, the inferential process does not involve matching a state (as in PCNs) but tracking a path \cite{Ramstead2023}. Unpacking Equation \ref{eq:hidden_update}, we note that the 0th order is subject to a forward error from the prior, a backward error from the likelihood, and a backward error from the dynamics function. On the other hand, the 1st order is only subject to the latter but in the form of a forward error. The belief is then updated via gradient descent, i.e., $\tilde{\bm{\mu}}_{x,t+1} = \tilde{\bm{\mu}}_{x,t} + \Delta_t \dot{\tilde{\bm{\mu}}}_x$, where $\Delta_t$ is a time constant.
	
	How can this agent perform a simple reaching movement? As highlighted in Figure \ref{fig:1st_solution_new}, we can encode the arm angle and velocity as generalized hidden states. We will talk later about the relation between proprioceptive and exteroceptive domains; for now, we consider a single DoF that has a univocal mapping between the joint angle of the arm and the Cartesian position of the hand. Indicating the target to reach by $\bm{\rho}$, we can define the following dynamics function:
	\begin{equation}
		\label{eq:rho_dyn}
		\bm{f}(\bm{x}) = \bm{\rho} - \bm{x}
	\end{equation}
	expressing a simple attractor toward the target \cite{Lanillos2020,Sancaktar2020,Oliver2021,Meo2021,Rood2020}. These dynamics do not exist in the actual generative process, and it is indeed this discrepancy that forces the environment to conform to the agent's beliefs. Specifically, Equation \ref{eq:rho_dyn} means that the agent thinks its hand will be pulled toward the target with a strength proportional to the precision $\bm{\Pi}_x$. In fact, the attractor affects the belief update through the dynamics prediction error $\bm{\varepsilon}_x$, expressing a difference between the estimated velocity $\bm{\mu}_x^\prime$ and the one predicted by the agent through the dynamics function $\bm{f}$. Note that this error appears in both temporal orders: in brief, $\bm{\varepsilon}_x$ imposes a trajectory at the 1st order which in turn affects the 0th order directly through $\bm{\mu}_x^\prime$, and indirectly through the gradient $\partial_{x} \bm{f}$.
	
	\begin{figure}[h]
		\begin{subfigure}{0.4\textwidth}
			\centering
			\includegraphics[width=\textwidth]{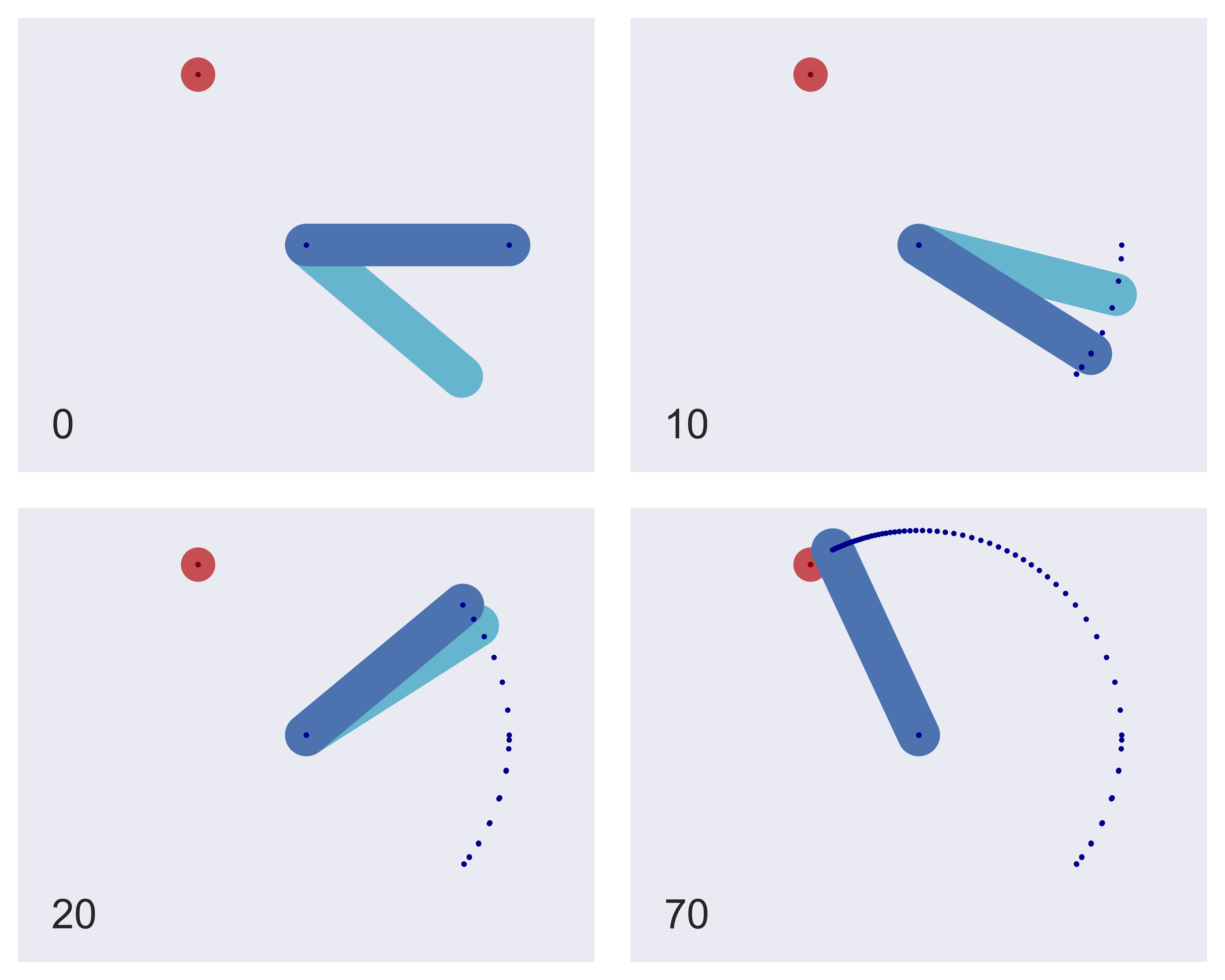}
			\vspace{1em}
			\caption{}
			\label{fig:frames_reaching}
		\end{subfigure}
		\hfill
		\begin{subfigure}{0.58\textwidth}
			\centering
			\includegraphics[width=\textwidth]{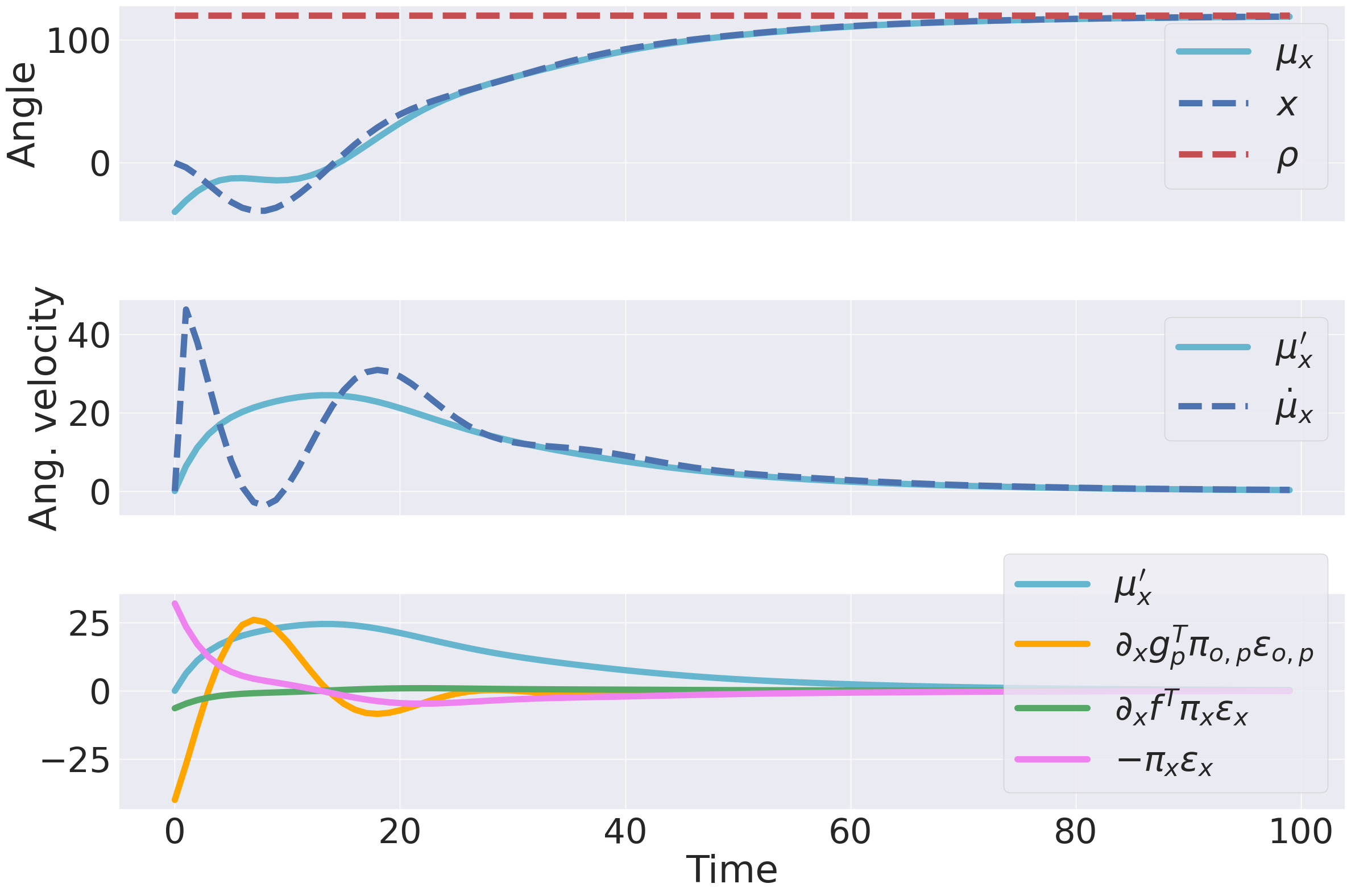}
			\caption{}
			\label{fig:dynamics_reaching}
		\end{subfigure}
		\caption{\textbf{(a)} In this task, the agent (a single DoF) has to reach a target angle represented by the red circle. Estimated and real arms are displayed in cyan and blue, respectively. Here, $\bm{\pi}_{\eta,x} = 0$, $\bm{\rho} = 120$°, and $\bm{\mu}_x$ was initialized to $-40$°. The time step is indicated in the bottom left corner of each frame. Since the belief was initialized at a negative value, the likelihood initially pulls the arm toward the wrong direction before adapting to the dynamics attractor. \textbf{(b)} The top graph shows the evolution of the real angle $\bm{x}$, its belief $\bm{\mu}_x$, and the target angle $\bm{\rho}$. The middle graph shows the evolution of the belief of the velocity $\bm{\mu}_x^\prime$ and the belief derivative $\dot{\bm{\mu}}_x$. The bottom graph shows the evolution of all the components that comprise the belief update: the belief of the velocity $\bm{\mu}_x^\prime$, the likelihood gradient $\partial_{x} \bm{g}^T \bm{\Pi}_{o,p} \bm{\varepsilon}_{o,p}$, the dynamics gradient $\partial_{x} \bm{f}^T \bm{\Pi}_x \bm{\varepsilon}_x$, and the weighted dynamics prediction error $-\bm{\Pi}_x \bm{\varepsilon}_x$. The latter has been plotted to compare its magnitude with the other components, although affecting the 1st temporal order.}
		\label{fig:example_reaching}
	\end{figure}
	
	The interactions between these quantities can be better understood from Figure \ref{fig:example_reaching}, showing a reaching movement with the defined dynamics function and the trajectories of the agent's generative model. Here, the belief is subject to two different forces: a likelihood gradient pushing it toward what it is currently perceiving (i.e., the real angle), and the other components that steer it toward the biased dynamics (i.e., the target angle $\bm{\rho}$). Note how in the third plot, of the three components that comprise the belief update, the backward error $\partial_{x} \bm{f}^T \bm{\Pi}_x \bm{\varepsilon}_x$ has the smallest amplitude. While the exact interactions arising from the dynamics prediction error have yet to be analyzed, in the following we assume that goal-directed behavior is achieved through the forward error at the 1st order $-\bm{\Pi}_x \bm{\varepsilon}_x$. An alternative would be to directly control the backward error without maintaining a belief of the 1st order \cite{Sancaktar2020}, which however requires taking a gradient into account and may be more challenging when defining appropriate attractors. Finally see how, in the middle plot, the agent tries at every instant to minimize the difference between $\bm{\mu}_x^\prime$ and $\dot{\bm{\mu}}_x$, thus tracking the actual path of the hidden states.
	
	But how does the agent move in practice? As mentioned in the Introduction, action is the other side of the coin of the free energy principle, through which the agent samples those observations conforming to its prior beliefs. In fact, in addition to the perceptual inference typical of predictive coding, active inference assumes that organisms minimize free energy also by interacting with the environment; this minimization breaks down to an even simpler update that only depends on (proprioceptive) prediction errors $\bm{\varepsilon}_{o,p}$. Since these prediction errors are generated from the agent's belief, this means that whenever the latter is biased toward some preferred state, movement naturally follows. There is thus a delicate balance between perception -- in which prediction errors climb up the hierarchy to bring the belief closer to the observations -- and action -- in which prediction errors are suppressed at a low level so that the observations are brought closer to their predictions. However, there is an open issue regarding how active inference should be practically realized in continuous time. A few studies demonstrated that using exteroceptive information directly for computing motor commands could result in smoother movements and resolution of visuo-proprioceptive conflicts \cite{Friston2010,mlp22,Priorelli2023}, and in fact some robotic implementations effectively used this approach \cite{Sancaktar2020,Oliver2021}. However, evidence seems to indicate that motor commands are generated by suppression of proprioceptive signals only \cite{Adams2013,Friston2011opt}, which is already in the intrinsic reference frame needed for movement and thus results in easier inverse dynamics. For this reason, in the following we assume that movements are realized by minimizing the free energy with respect to proprioceptive prediction errors:
	\begin{equation}
		\label{eq:motor}
		\dot{\bm{a}} = - \partial_{a} \mathcal{F} = - \partial_{a} \bm{g}_{p}^T \bm{\Pi}_{o,p} \bm{\varepsilon}_{o,p}
	\end{equation}
	where $\partial_{a} \bm{g}_p$ performs an inverse dynamics from proprioceptive predictions to motor commands $\bm{a}$, likely to be implemented by classical spinal reflex arcs. As a last note, the actions can also depend on multiple orders -- velocity, acceleration, and so on -- allowing more efficient movement and control \cite{Baioumy2020,9937202,Bos2022,Meera2021}, but since it is beyond our scope, we only minimize the 0th order. Nonetheless, 1st-order movements -- e.g., maintaining a constant velocity -- are still possible by specifying appropriate dynamics of the hidden states.
	
	
	\subsection{\label{sec:tracking}Tracking objects}
	
	\begin{figure}[h]
		\begin{subfigure}{0.55\textwidth}
			\centering
			\includegraphics[width=\textwidth]{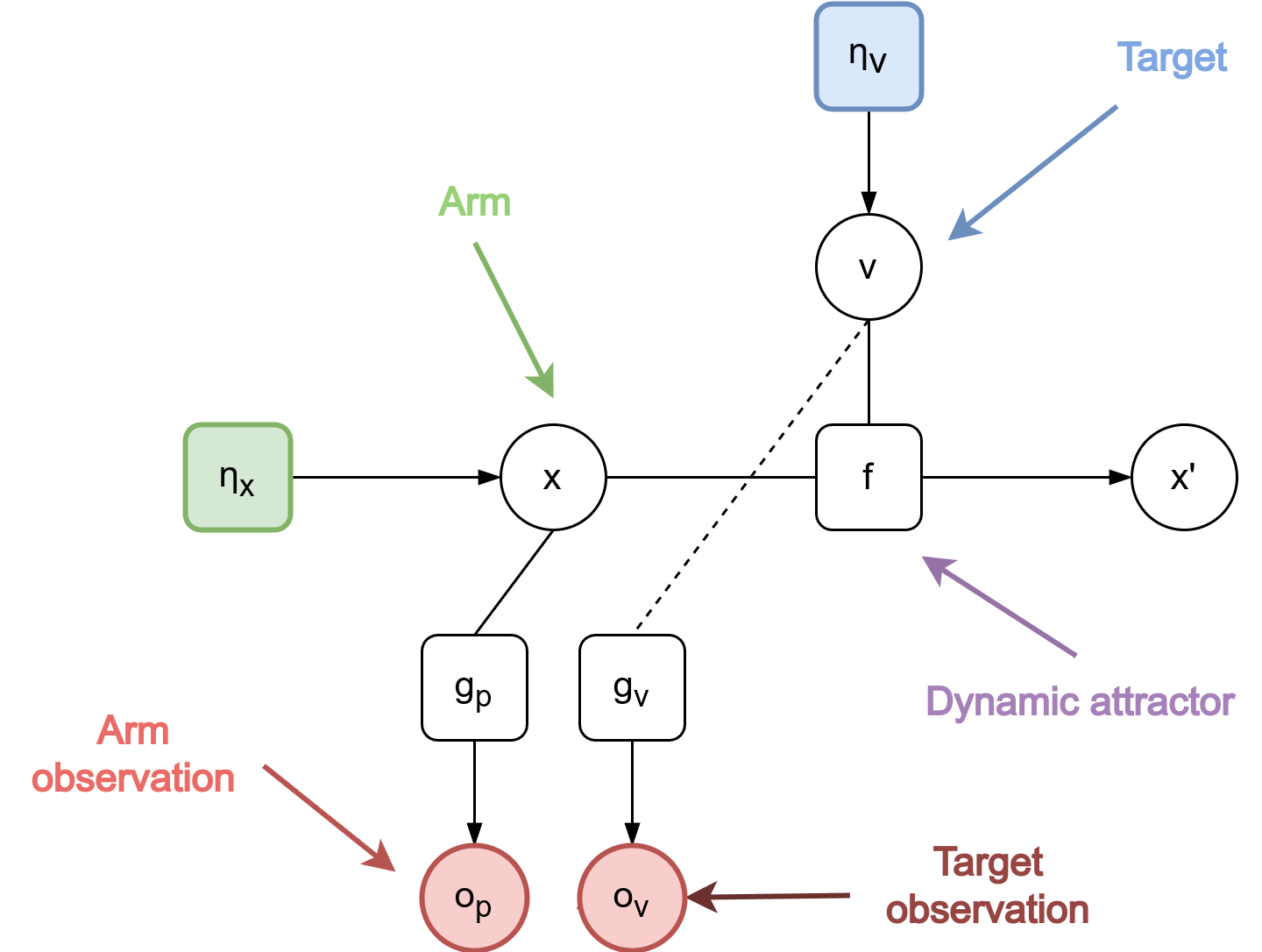}
			\caption{}
			\label{fig:2nd_solution_new}
		\end{subfigure}
		\hfill
		\begin{subfigure}{0.43\textwidth}
			\centering
			\includegraphics[width=\textwidth]{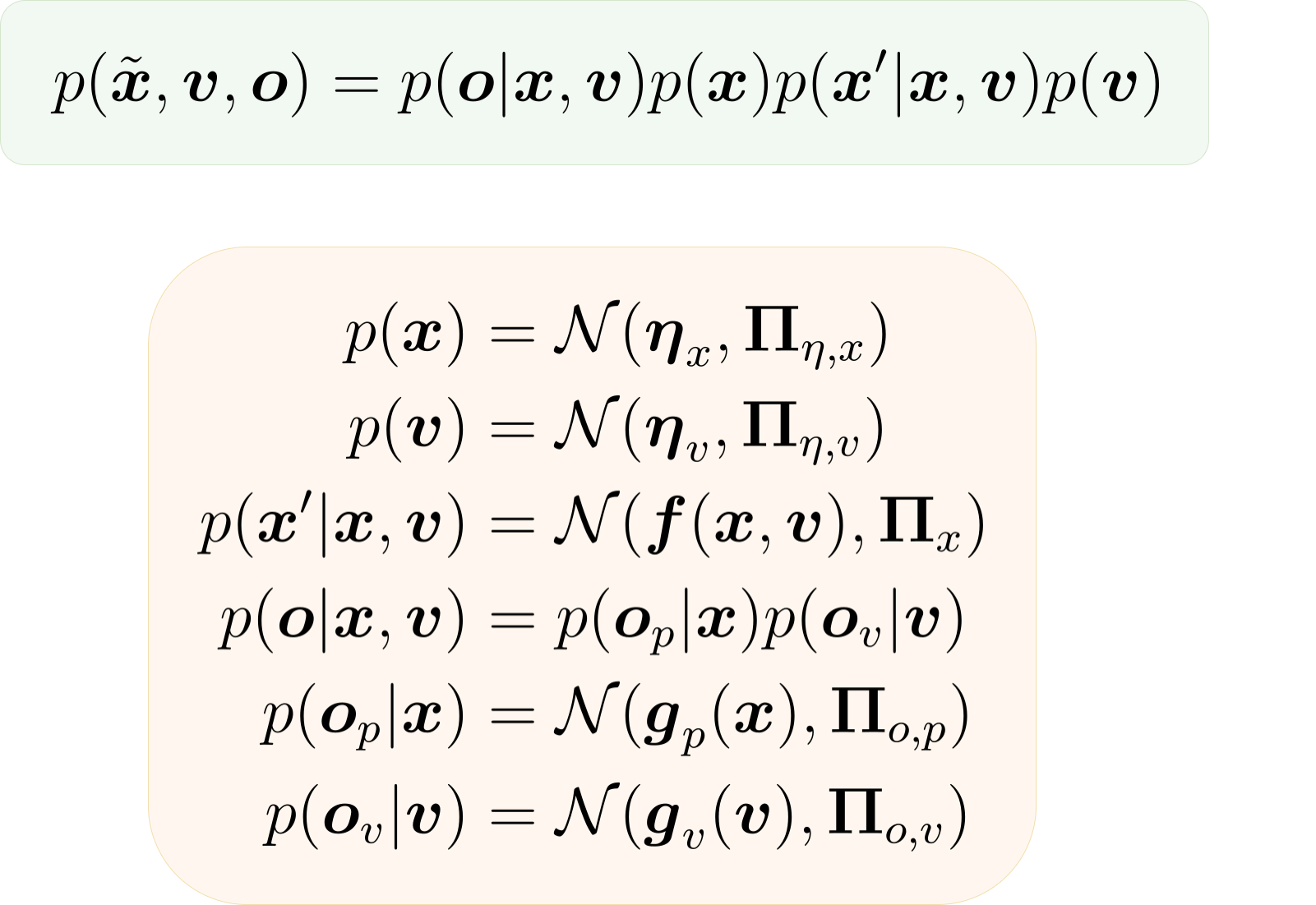}
			\vspace{1em}
			\caption{}
			\label{fig:gen_model_tracking}
		\end{subfigure}
		\caption{\textbf{(a)} The target is now encoded in the hidden causes $\bm{v}$, generating a dynamic attractor for object tracking. In fact, both hidden states and hidden causes generate predictions through proprioceptive and visual likelihood functions $\bm{g}_p$ and $\bm{g}_v$, and both concur in estimating the 1st-order hidden states $\bm{x}^\prime$. \textbf{(b)} Agent's generative model.}
	\end{figure}
	
	The simple agent defined in the previous section can only realize fixed trajectories embedded in the dynamics function, so how can it track moving objects? This is usually done by introducing a key concept in active inference, the \textit{hidden causes} $\bm{v}$, which link hierarchical levels and specify how the dynamics function evolves. In the active inference literature of motor control, they are also used to encode the target to be reached \cite{Friston2010,Parr2018c,Adams2015b,Pio-Lopez2016}, as depicted in Figure \ref{fig:2nd_solution_new}. Considering the target as a causal variable for the hidden states and sensory observations makes sense from an active perspective whereby ``it is an object I want to reach that generates my movements''. Now, the agent's generative model becomes that of Figure \ref{fig:gen_model_tracking}. Note that there are two priors, one over the hidden states and another over the hidden causes, respectively denoted by $\bm{\eta}_x$ and $\bm{\eta}_v$. Also, both dynamics and likelihood functions depend on the hidden causes, and we assumed a further factorization for the likelihood, where $\bm{o}_p$ and $\bm{o}_v$ denote the arm and target observations, respectively. For simplicity, we assume that the visual likelihood function $\bm{g}_v$ is a simple identity generating the target angle. It is through the connection between hidden causes and observations that the agent can operate in dynamic environments. In fact, we can define the following dynamics function:
	\begin{equation}
		\label{eq:attr}
		\bm{f}(\bm{x}, \bm{v}) = \bm{v} - \bm{x}
	\end{equation}
	\begin{figure}
		\begin{subfigure}{0.37\textwidth}
			\centering
			\includegraphics[width=\textwidth]{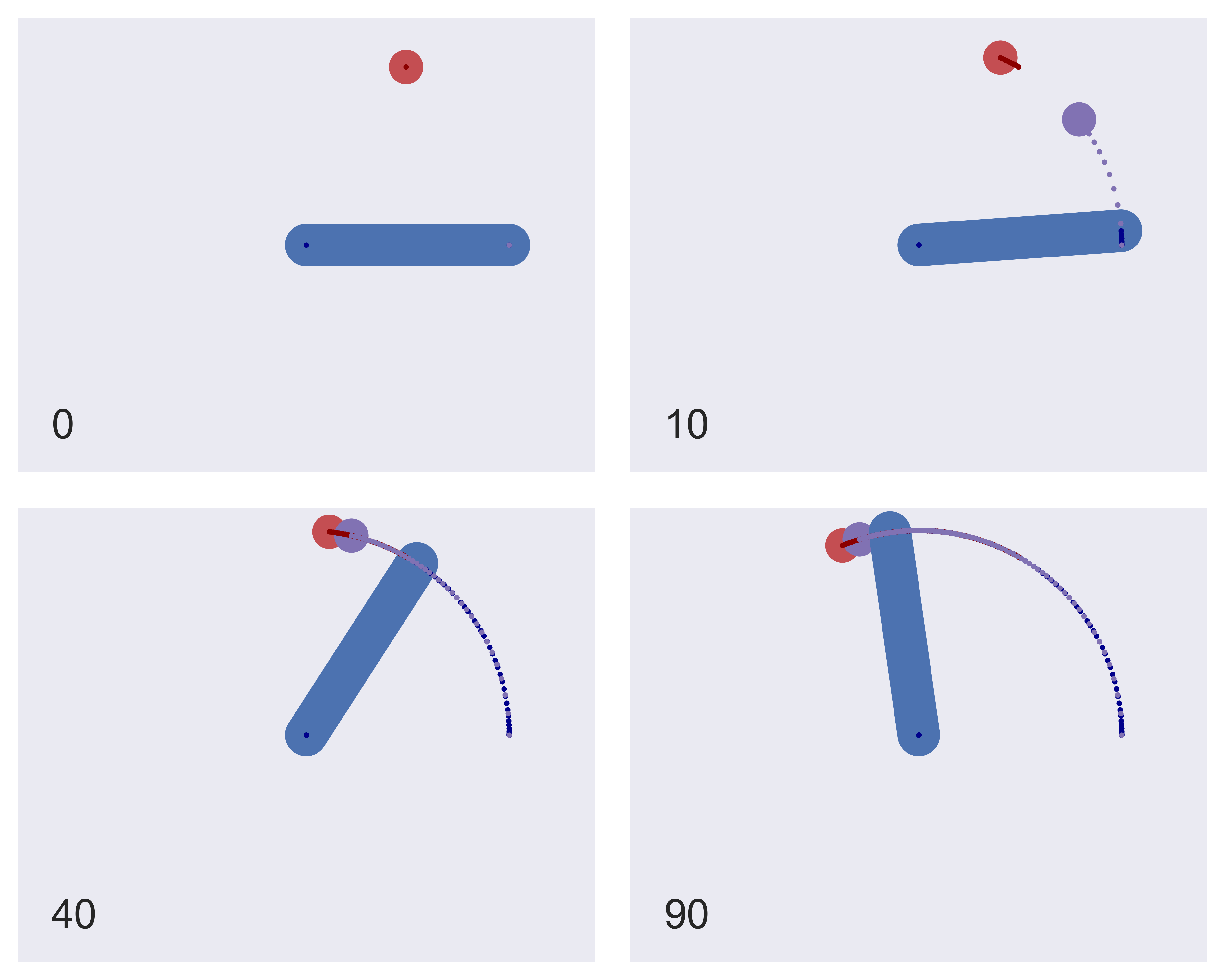}
			\vspace{1em}
			\caption{}
			\label{fig:frames_tracking}
		\end{subfigure}
		\hfill
		\begin{subfigure}{0.6\textwidth}
			\centering
			\includegraphics[width=\textwidth]{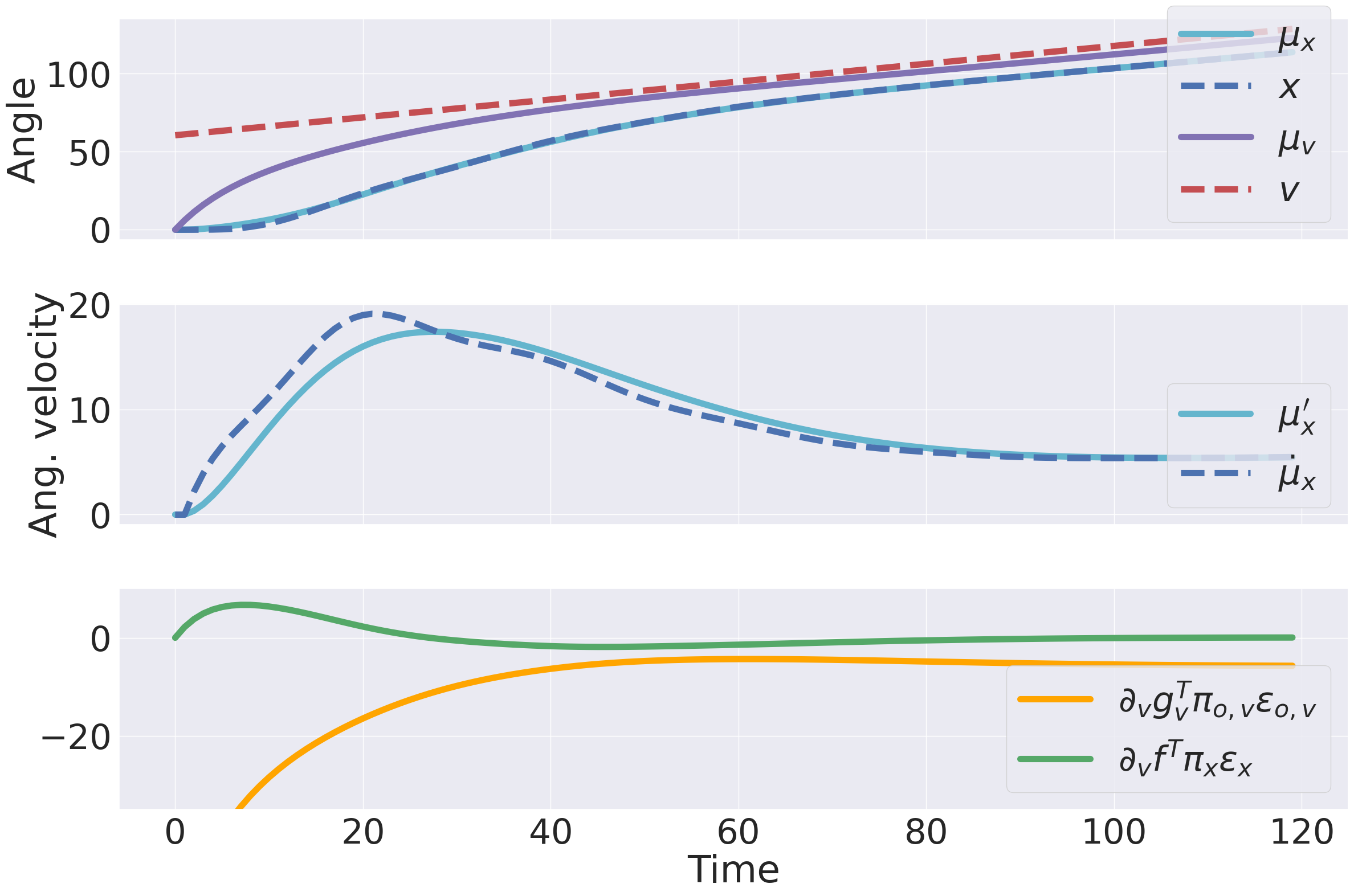}
			\caption{}
			\label{fig:dynamics_tracking}
		\end{subfigure}
		\caption{\textbf{(a)} In this task, the agent has to track a target angle rotating at a constant velocity. Estimated and real targets are displayed in purple and red, respectively. Here, $\bm{\Pi}_{\eta,x} = 0$, $\bm{\Pi}_{\eta,v} = 0$, $\bm{v}$ was initialized to $60$°, and both $\bm{\mu}_x$ and $\bm{\mu}_v$ were initialized to $0$°. Here, the belief of the hidden causes pulls the belief of the hidden states with it while approaching the real target angle. \textbf{(b)} The top graph shows the evolution of the real angle $\bm{x}$, its belief $\bm{\mu}_x$, the target angle $\bm{v}$, and its belief $\bm{\mu}_v$. The middle graph shows the evolution of the belief of the velocity $\bm{\mu}_x^\prime$ and the belief derivative $\dot{\bm{\mu}}_x$, as before. The bottom graph shows the evolution of all the components that comprise the hidden causes update: the likelihood gradient $\partial_{v} \bm{g}_v^T \bm{\Pi}_{o,v} \bm{\varepsilon}_{o,v}$, and the dynamics gradient $\partial_{v} \bm{f}^T \bm{\Pi}_x \bm{\varepsilon}_x$. Note how, in the middle plot, the estimated 1st temporal order stabilizes to a non-zero value as the agent rotates with a constant angular velocity.}
		\label{fig:example_tracking}
	\end{figure}
	where we just replaced the static target $\bm{\rho}$ with the hidden causes. As for the hidden states, we define a (Gaussian) approximate posterior over the hidden causes $\bm{q}(\bm{v}) = \mathcal{N}(\bm{\mu}_v, \bm{P}_v)$, with mean $\bm{\mu}_v$ and precision $\bm{P}_v$. Then, the mean is updated according to:
	\begin{equation}
		\label{eq:v_update}
		\dot{\bm{\mu}}_v = - \partial_{v} \mathcal{F} = - \bm{\Pi}_{\eta,v} \bm{\varepsilon}_{\eta,v} + \partial_{v} \bm{g}_v^T \bm{\Pi}_{o,v} \bm{\varepsilon}_{o,v} + \partial_{v} \bm{f}^T \bm{\Pi}_x \bm{\varepsilon}_x
	\end{equation}
	where we defined the following observation and prior prediction errors in terms of the approximate posterior:
	\begin{align}
		\begin{split}
			\bm{\varepsilon}_{o,v} &= \bm{o}_v - \bm{g}_v(\bm{\mu}_v) \\
			\bm{\varepsilon}_{\eta,v} &= \bm{\mu}_v - \bm{\eta}_v
		\end{split}
	\end{align}
	As evident, the hidden causes are subject to a prior prediction error, a backward dynamics error, and a backward likelihood error -- similar to the update of the hidden states, except that this inference is over a state and not a path.	Via the backward likelihood error, the agent can correctly estimate the target configuration whenever it moves, as shown in the tracking simulation of Figure \ref{fig:example_tracking}. Concerning the dynamics prediction error, it can now flow into two different pathways: specifically, the role of the gradients $\partial_{x} \bm{f}$ and $\partial_{v} \bm{f}$ are to respectively infer the state and the cause that may have generated a particular velocity; their actual role will be clear in Chapter 4.
	
	
	\subsection{\label{sec:affordances}Intention modulation and object affordances}
	
	Although capable of operating in dynamic contexts, the last approach still reproduces a simple scenario in which a target has no internal dynamics and always has the role of a cause for a hidden state. In other words, it does not permit modeling realistic tasks such as a pick-and-place operation, where an object is first the cause of a reaching and grasping movement, but then it becomes the consequence of another cause such as a goal position, resulting in a placing movement; critically, it does not allow to model a task wherein not only the dynamics of the self, but also the dynamics of the target must be learned (e.g., if a moving object should be grasped on the fly, the agent should infer its trajectory to anticipate where it will fall).
	
	It follows that to operate in a complex environment, the agent must (i) maintain complete representations for each entity that it wants to interact with, and (ii) flexibly assign causes and consequences for the next movement depending on the current context -- in a similar way to policies in discrete models, as will be explained later. Therefore, we first encode multiple environmental entities as \textit{potential body configurations} in the hidden states, i.e., $\bm{x} = [\bm{x}_0, \bm{x}_1, \dots, \bm{x}_N]$, where $\bm{x}_0$ is the actual body configuration (as before), and $N$ is the number of entities \cite{Priorelli2023}. Consequently, the factorized likelihood function generates a proprioceptive observation for the first component $\bm{x}_0$, and visual observations for each entity:
	\begin{equation}
		\bm{o} = [\bm{o}_p, \bm{o}_{v,1}, \dots, \bm{o}_{v,N}] = [\bm{g}_p(\bm{x}_0), \bm{g}_v(\bm{x}_1), \dots, \bm{g}_v(\bm{x}_N)]
	\end{equation}
	Here, visual observations are assumed to be in the Cartesian domain, so the visual likelihood function $\bm{g}_v$ generates the hand positions of the potential configurations through forward kinematics. This structure is similar to the previous model, except that the target is now embedded in the hidden states along with the hand, and that there is no connection between hidden causes and observations. We could define a similar factorization for the hidden causes and dynamics function, such that each entity would have an independent dynamics biased by a specific cause (e.g., where the arm or the target will be in the future); however, this is of limited use in a pick-and-place operation that demands interaction between entities. We hence compute an intentional state with a single function, such as:
	\begin{equation}
		\label{eq:intention}
		\bm{i}_m(\bm{x}) = \bm{W}_m \bm{x} + \bm{b}_m
	\end{equation}
	The weights $\bm{W}_m$ perform a linear transformation of the hidden states that combines every entity, while the bias $\bm{b}_m$ imposes a static configuration \cite{Priorelli2023b}. Equation \ref{eq:intention} can be realized through simple neural connections, wherein the weights are encoded as synaptic strengths and the bias represents the threshold needed to fire a spike, and both can be known a priori or learned via sensory evidence. An error is then computed between this intentional state and the current one:
	\begin{equation}
		\bm{e}_{i,m} = \bm{i}_m(\bm{x}) - \bm{x}
	\end{equation}
	This vector has the same role as the attractor of Equation \ref{eq:attr}, but now it points toward a function of the hidden states. Finally, we define the following dynamics function:
	\begin{equation}
		\label{eq:dyn}
		\bm{f}_m(\bm{x}, v_m) = v_m \bm{e}_{i.m}
	\end{equation}
	multiplying the error by a single-value hidden cause $v_m$. Thus, the latter is not intended as an explicit trajectory prior over the hidden states (e.g., encoding where my arm will be in the future), whose role is now delegated to the bias $\bm{b}_m$; but as an attractor gain, whereby a high value implies a strong force toward the future state. Since $\bm{i}_m$ is used to define a path for the hidden states aiming to produce a desired configuration, we call it an \textit{intention}.
	
	\begin{figure}[h]
		\begin{subfigure}{0.55\textwidth}
			\centering
			\includegraphics[width=\textwidth]{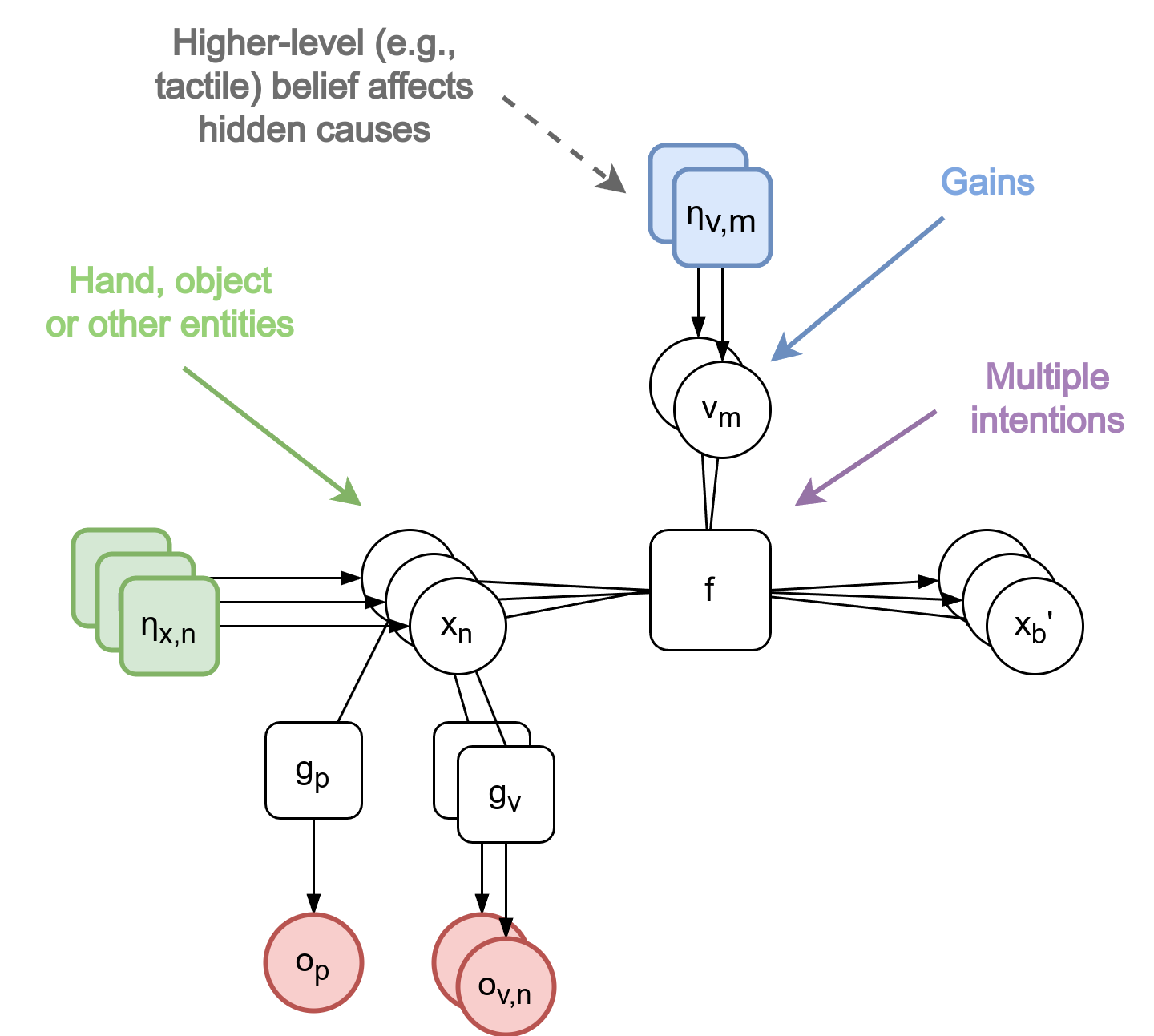}
			\caption{}
			\label{fig:affordances}
		\end{subfigure}
		\hfill
		\begin{subfigure}{0.42\textwidth}
			\centering
			\includegraphics[width=\textwidth]{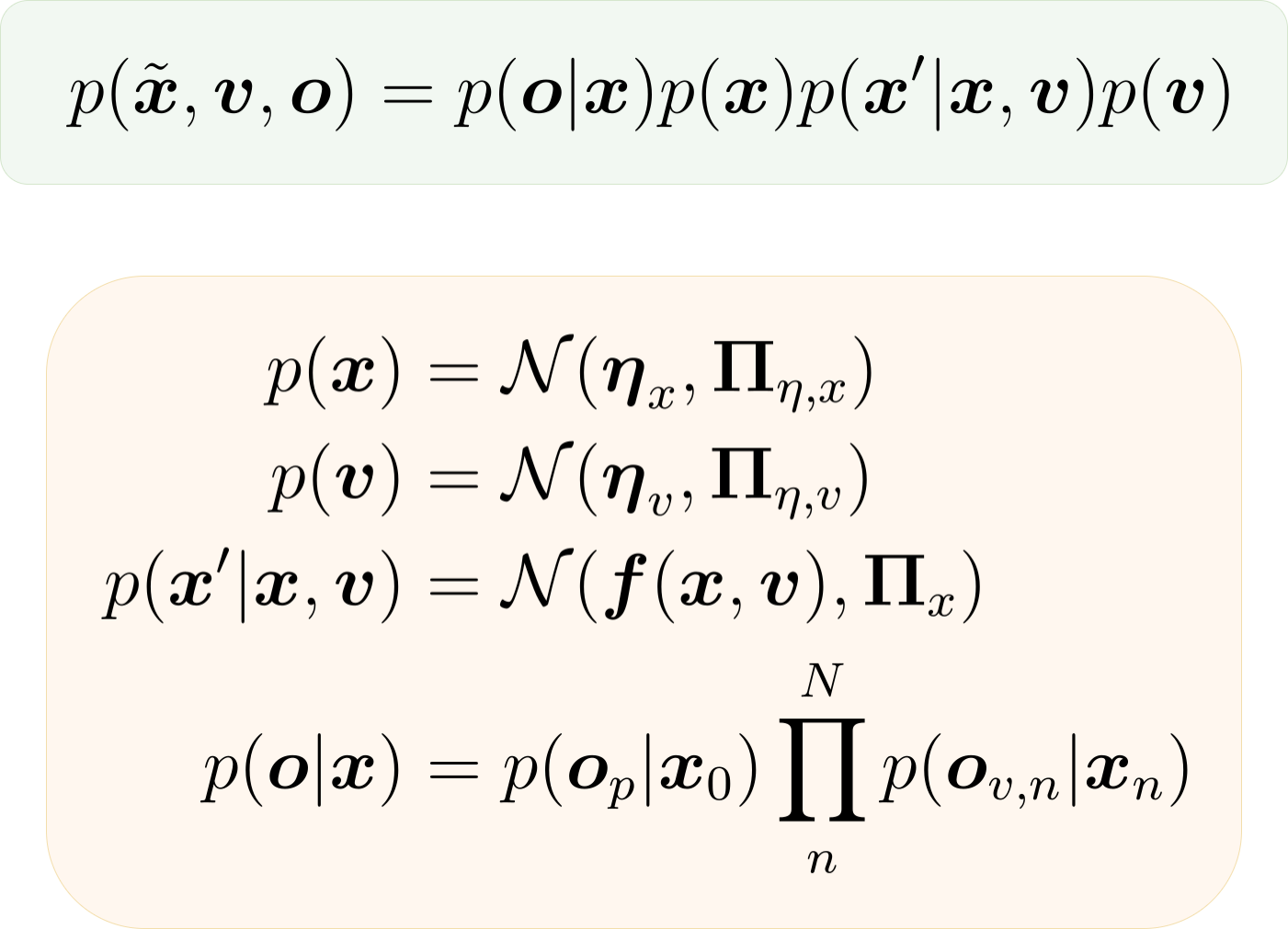}
			\vspace{1em}
			\caption{}
			\label{fig:gen_model_modulation}
		\end{subfigure}
		\caption{\textbf{(a)} Factor graph of the unit with object affordances. Hidden states are factorized into independent components that encode the actual bodily states and potential configurations related to the objects. The first component $\bm{x}_0$ generates proprioceptive predictions, while the successive components generate visual predictions of the objects. Every hidden cause $v_m$ now defines an attractor gain expressing the strength of an agent's intention (encoded as a distinct evolution of the world). These are combined to produce a trajectory $\bm{\eta}_{x}^\prime$, comprising all the body configurations $\bm{x}_n$. The transition between intentions can be achieved by a higher-level prior, e.g., a belief of tactile sensations. The weights $\bm{W}_m$ of the intention can be used, e.g., to track moving objects, while the bias $\bm{b}_m$ realizes a static configuration. See \cite{Priorelli2023,Priorelli2023d} for more details. \textbf{(b)} Agent's generative model.}
	\end{figure}
	
	The dynamics function of Equation \ref{eq:dyn} is not composed of segregated pathways as for the likelihood, but affects all the environmental entities at once -- e.g., it computes a trajectory for the arm depending on the target. The steps performed by the agent during a reaching movement are the following: (i) the 0th order imposes a trajectory to the 1st order and generates a sensory prediction; (ii) the 0th order infers the consequences of its predictions, hence it is now biased toward both the intentional state and the observation; (iii) a proprioceptive prediction is generated from this new biased position, eventually driving action. This approach can be seen as a generalization of \cite{Adams2015b} where, in the context of oculomotor behavior, the target and the center of gaze were encoded as hidden states, each with their own dynamics and attracted by a hidden location. Although limited compared to non-linear dynamics functions (e.g., obstacle avoidance can be realized via repulsive potentials \cite{Priorelli2023b}), with the specific form defined above -- along with the hidden states factorization -- there is a high flexibility for complex interactions. Further, interpreting the hidden causes as a gain still makes sense from an active inference perspective because what is represented at a higher level is the intention to move at the target, while the target location is inferred at a lower level.
	
	Taken alone, considering a hidden cause as an attractor gain may not seem so helpful. However, as depicted in Figure \ref{fig:affordances}, we can combine $M$ intentions in the following way:
	\begin{align}
		\begin{split}
			\label{eq:intention_dyn}
			\bm{\eta}_{x}^\prime &= \bm{f}(\bm{x}, \bm{v}) = \sum_m^M \bm{f}_m(\bm{x}, v_m) = \sum_m^M v_m (\bm{i}_m(\bm{x}) - \bm{x})
		\end{split}
	\end{align}
	In short, the agent entertains $M$ distinct dynamic hypotheses of how the world may evolve. Trajectories $\bm{f}_m(\bm{x}, v_m)$ are separately computed from each intention $\bm{i}_m$ and with their respective gains; then, the final trajectory $\bm{\eta}_{x}^\prime$ is found by combining all of them. The reason why we used the prior notation for the trajectory predictions will be clear in Chapter 4. Since each attractor $\bm{e}_{i,m}$ is proportional to its hidden cause $v_m$, the latter lends itself to a parallelism with the policies of discrete models, as will be explained later: if $v_m$ is set to $1$ and all the others to $0$, the hidden states will be subject only to intention $m$; conversely, if multiple hidden causes are active, the hidden states will be pulled toward a combination of the corresponding intentions. This means that the hidden causes act both as attractor gains -- expressing the absolute strength by which the belief is steered toward the desired direction -- and as intention modulators -- defining the relative strength between each trajectory. As a result, we have an additional modulation that combines with the dynamics precision $\bm{\Pi}_x$; their interactions will be explained in Chapter 4.	The resulting behavior is similar to the Lotka-Volterra dynamics used in \cite{Friston2011b}. In the latter, the agent revisits in sequence points in attractor space linked to specific locations, and the defined dynamics ensure that only one attractor is active at any time. This permits modeling itinerant movements such as handwriting or walking.
	
	\begin{figure}[h]
		\begin{subfigure}{0.38\textwidth}
			\centering
			\includegraphics[width=\textwidth]{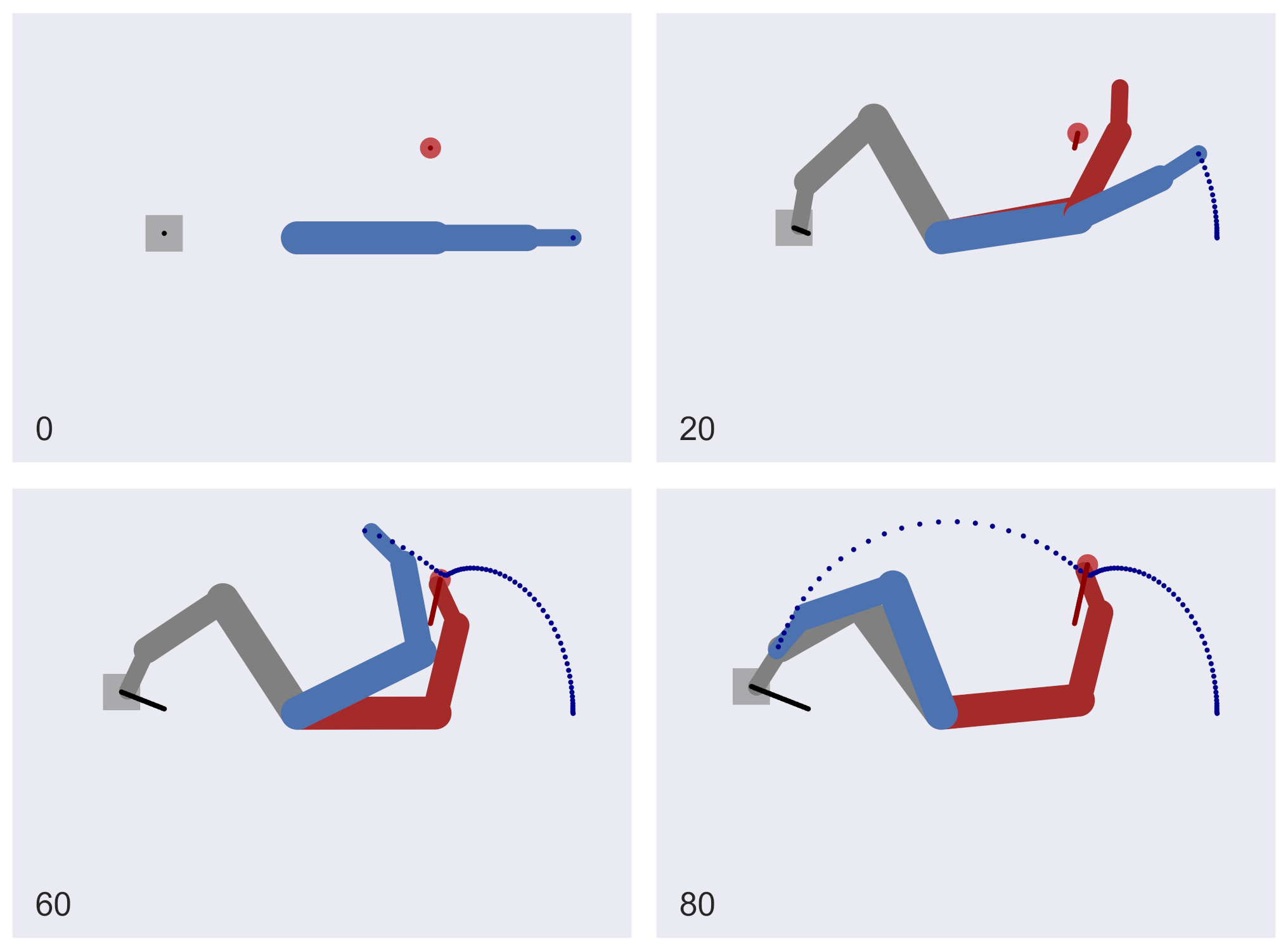}
			\vspace{2em}
			\caption{}
			\label{fig:frames_affordances}
		\end{subfigure}
		\hfill
		\begin{subfigure}{0.6\textwidth}
			\centering
			\includegraphics[width=\textwidth]{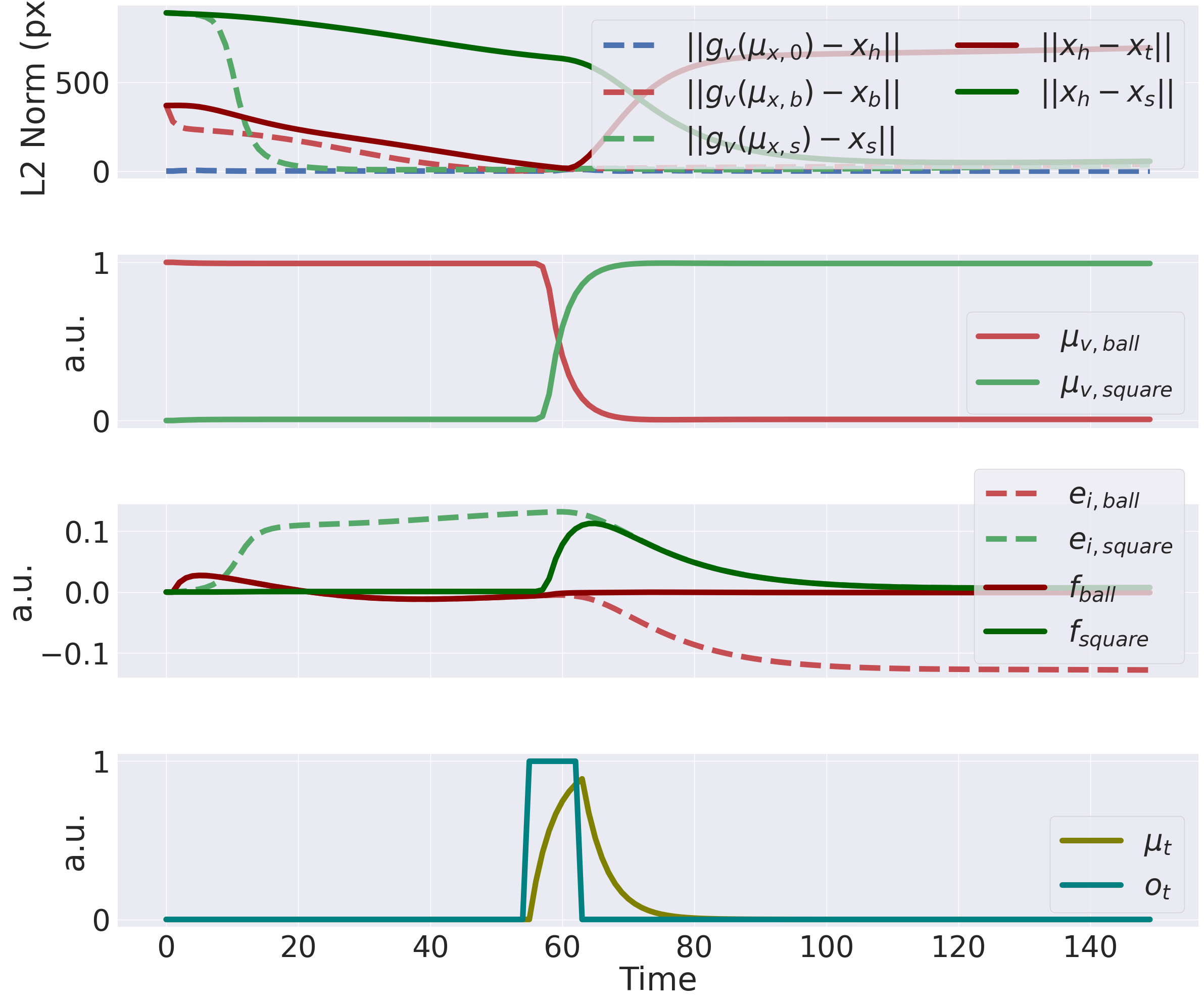}
			\caption{}
			\label{fig:dynamics_modulation}
		\end{subfigure}
		\caption{\textbf{(a)} In this task, the agent has to first touch the moving ball (in red) and then track the moving square (in grey). The transition is done by a tactile belief. Note how throughout the task, the agent maintains potential body configurations related to the two objects. See \cite{Priorelli2023,Priorelli2023d} for more details. \textbf{(b)} The first graph shows the difference between the estimated and real hand positions (dotted blue), the difference between the potential hand position related to the ball and the actual ball position (dotted red), the difference between the potential hand position related to the square and the actual square position (dotted green), the difference between the real hand and ball positions (solid dark red), and between the real hand and square positions (solid dark green). The second graph shows the evolution of the hidden causes associated with the two intentions, $\mu_{v,ball}$ and $\mu_{v,square}$. The third graph shows the evolution of the attractors $\bm{e}_{i,ball}$ and $\bm{e}_{i,square}$, and the dynamics functions $\bm{f}_{ball}$ and $\bm{f}_{square}$, regarding the first joint angle. The last graph shows the evolution of the tactile observation $\bm{o}_t$, and its belief $\bm{\mu}_{t}$.}
		\label{fig:example_modulation}
	\end{figure}
	
	The generative model is shown in Figure \ref{fig:gen_model_modulation}. The update rule for the 0th-order hidden states becomes:
	\begin{equation}
		\dot{\bm{\mu}}_x = \bm{\mu}_x^\prime - \bm{\Pi}_{\eta,x} \bm{\varepsilon}_{\eta,x} + \partial_{x} \bm{g}_p^T \bm{\Pi}_{o,p} \bm{\varepsilon}_{o,p} + \sum_n^N \partial_{x} \bm{g}_v^T \bm{\Pi}_{o,v,n} \bm{\varepsilon}_{o,v,n} + \partial_{x} \bm{f}^T \bm{\Pi}_x \bm{\varepsilon}_x
	\end{equation}
	In particular, the dynamics prediction error:
	\begin{equation}
		\label{eq:eps_x}
		\bm{\varepsilon}_x = \bm{\mu}_x^\prime - \bm{\eta}_{x}^\prime
	\end{equation}
	realizes an average trajectory that the agent predicts for the current situation. This approach is effective for two reasons. First, it allows defining a composite movement in terms of simpler subgoals, which can be tackled separately; this can be helpful, for instance, if one has to analyze the behavior of an agent when subject to two or more opposing priors \cite{Priorelli2023}. Second, a simple multi-step behavior without planning can be achieved \cite{Priorelli2023d}, wherein one just needs to modulate the hidden causes. Transitions between continuous trajectories could then be realized by higher-level priors, e.g., a belief of tactile sensations. Third, and most importantly, it permits maintaining in parallel potential body configurations related to the objects to be manipulated -- thus providing efficient transitions between movements -- and encoding the objects according to their affordances and the agent's intentions (e.g., grasping a cup by the handle or with the whole hand). These features are illustrated in the simulation of Figure \ref{fig:example_modulation}, showing a two-step reaching task with moving objects.
	
	
	\section{Hierarchical models}
	
	So far, we have discussed several units with two kinds of inputs -- a prior over the hidden states and a prior over the hidden causes -- and one kind of output -- the 0th-order observation. In this chapter, we focus on combining such units in a single network to achieve a more advanced and efficient control. For this, we will make use of the first input, leaving the discussion about the second to the next chapter.
	
	In hierarchical active inference models, units are arranged in layers so that the output of one layer provides input to the subordinate layers. This architecture permits representing complex data, such as convolutions models or nonlinear time series \cite{Friston2008}. In motor control, a (deep) hierarchy of goals integrates control and motivational streams of the brain \cite{Pezzulo2018}. For robotics, a hierarchical kinematic model can be designed in continuous time, wherein each unit encodes a certain Degree of Freedom (DoF) in intrinsic and extrinsic reference frames \cite{Priorelli2023b}. This permits realizing advanced movements that involve simultaneous coordination of multiple limbs, e.g., moving with a glass in hand. This hierarchical structure can be generalized to perform homogeneous transformations between reference frames, e.g., perspective projections \cite{Priorelli2023c}.
	
	
	\subsection{\label{sec:ie}Intrinsic and extrinsic causes}	
	
	The last unit presented affords a multi-step behavior in continuous time that accounts for object affordances and, to some extent, for dynamic elements of the environment. However, it can only estimate body configurations, while in real-life applications we constantly plan movements in the spatial domain. Further, it only generates visual predictions of an object related to a single DoF (e.g., the hand), while we generally deal with much more complex kinematic structures with different branches such as the human body. As in optimal control, continuous-time active inference considers three reference frames and two inversions: an \textit{extrinsic} signal (e.g., encoding the Cartesian position of a target) is first transformed in an \textit{intrinsic} signal (e.g., encoding the joint angles configuration corresponding to the hand at the target) through \textit{inverse kinematics}, which is in turn converted to the actual motor control signals (e.g., joint torques) through \textit{inverse dynamics} \cite{Todorov2004}. These two processes are also attributed to the human brain \cite{Floegel2023,Vallar1999,Hinman2019}, but there is a substantial difference between optimal control and active inference regarding how they unfold in practice. As mentioned in the previous chapter, in active inference the motor commands are replaced by proprioceptive prediction errors that are suppressed through spinal reflex arcs \cite{Adams2013}. As a consequence, inverse dynamics becomes easier because action is put aside and the agent has just to know the mapping from proprioceptive states to motor commands -- see Equation \ref{eq:motor}.
	
\begin{figure}[h]
		\begin{subfigure}{0.42\textwidth}
			\centering
			\includegraphics[width=\textwidth]{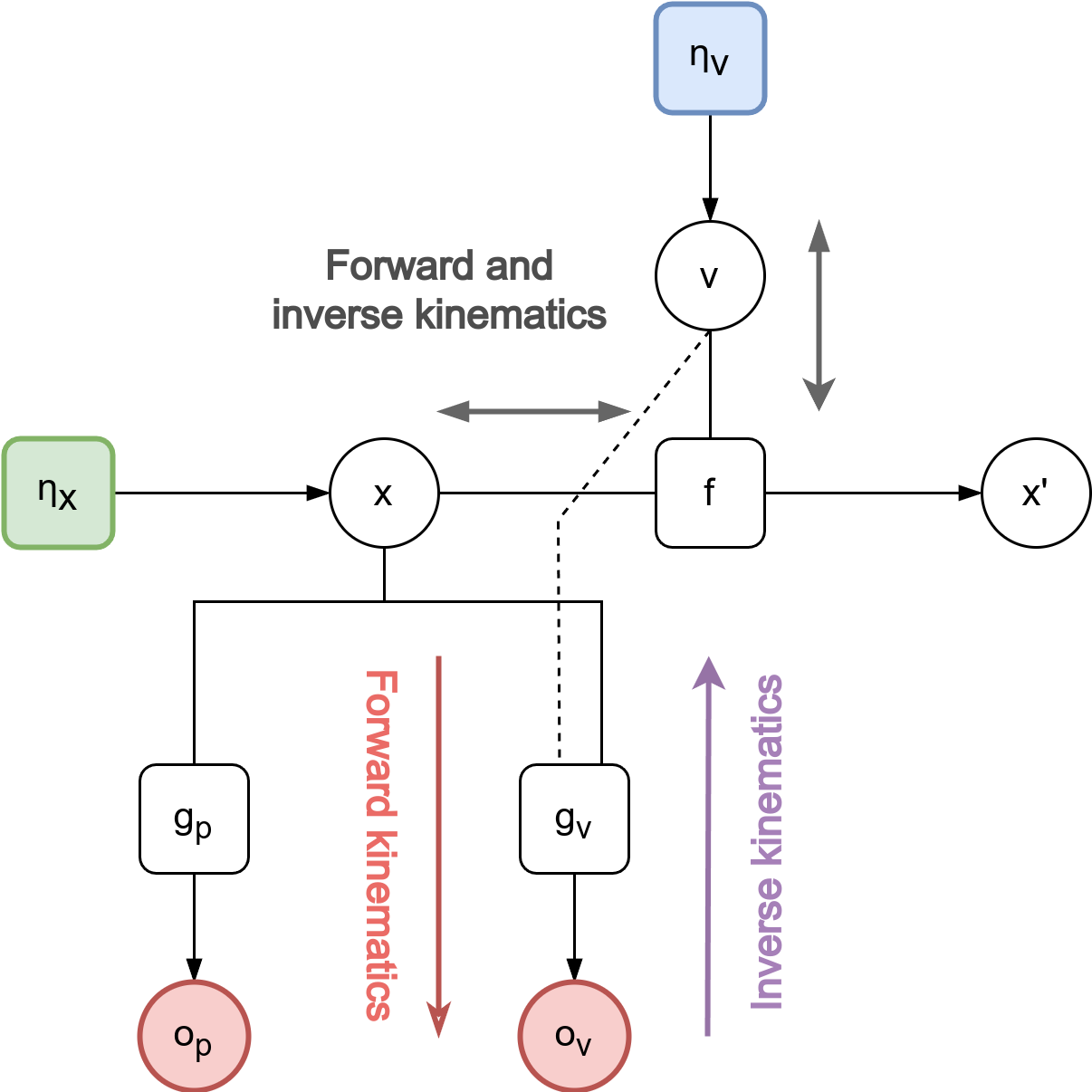}
			\caption{}
			\label{fig:duplicated}
		\end{subfigure}
		\hfill
		\begin{subfigure}{0.52\textwidth}
			\centering
			\includegraphics[width=\textwidth]{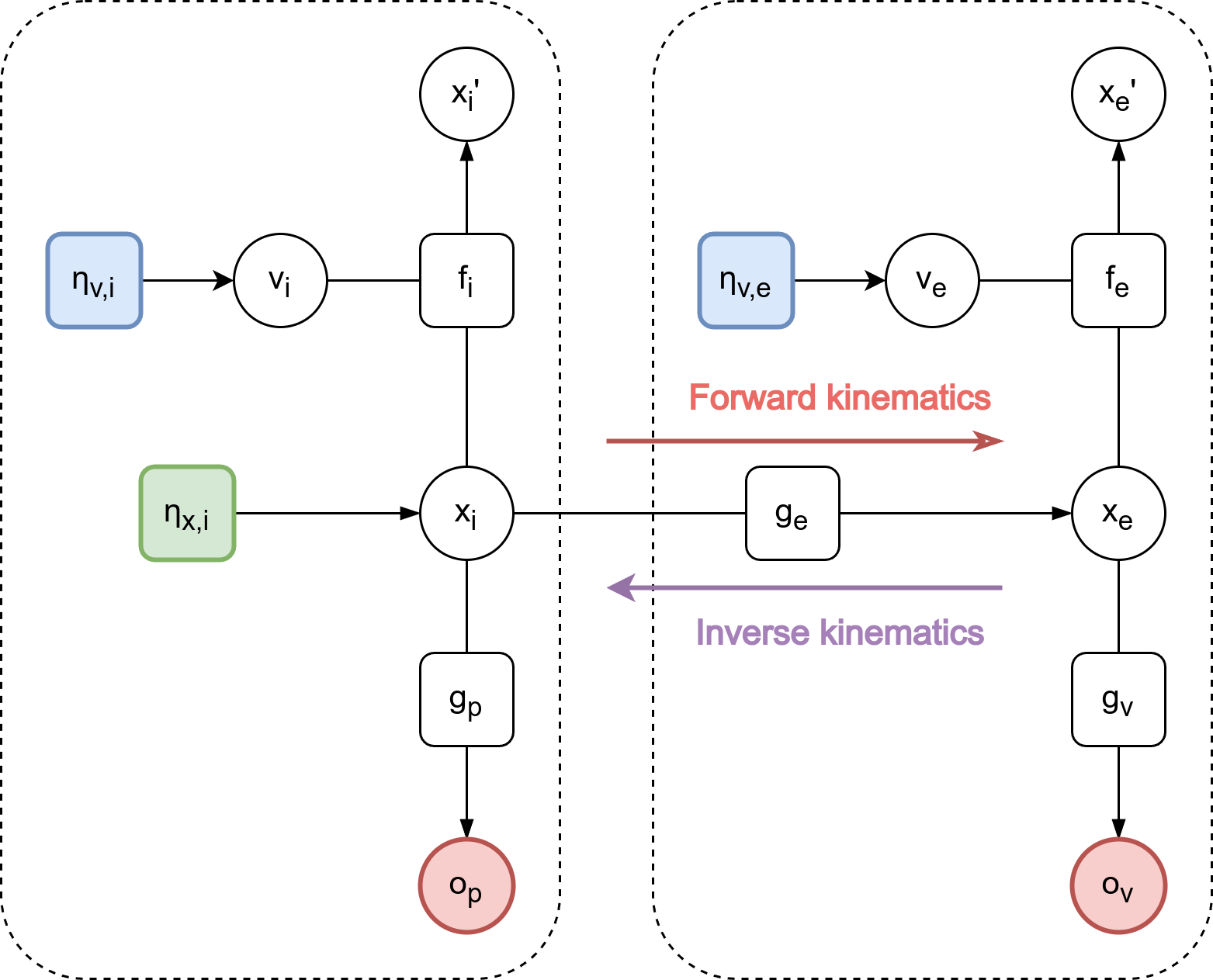}
			\caption{}
			\label{fig:inv_kin}
		\end{subfigure}
		\begin{subfigure}{1.3\textwidth}
			\vspace{2em}
			\hspace{1em}
			\includegraphics[width=\textwidth]{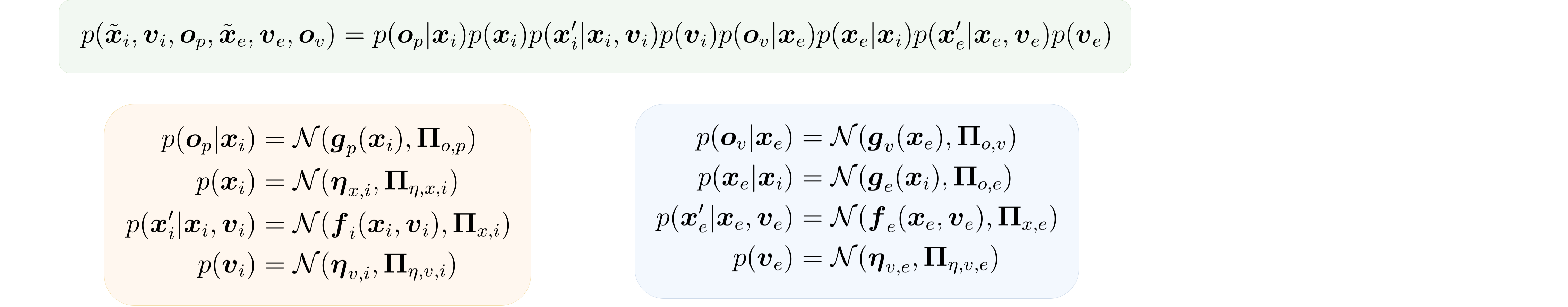}
			\caption{}
			\label{fig:gen_model_ie}
		\end{subfigure}
		\caption{\textbf{(a)} Factor graph of an active inference model commonly used for kinematics. Hidden causes of a single level represent a target to reach, while hidden states define the joint angles of the kinematic chain, generating proprioceptive predictions through $\bm{g}_p$ and visual predictions through $\bm{g}_v$. Forward and inverse kinematics are duplicated, further requiring the likelihood $\bm{g}_v$ to be embedded into the dynamics function. \textbf{(b)} Factor graph of an alternative hierarchical model for kinematics. Two different units (left and right blocks) encode information about arm joint angles and target position, respectively generating proprioceptive and visual predictions. The two levels are linked by the likelihood function $\bm{g}_e$ performing forward kinematics. Inverse kinematics and goal-directed behavior arise naturally through inference; moreover, both levels can express their own dynamics, affording more advanced control. \textbf{(c)} Generative model of (b).}
		\label{fig:kin_models}
	\end{figure}

	But what about inverse kinematics? Recall the perspective that we mentioned in the previous chapter, i.e., that ``it is an object I want to reach that generates my movements''. Turning optimal control upside down, active inference posits that action is driven by the proprioceptive consequences (e.g., changes in muscle lengths) of extrinsic causes (e.g., a limb position) \cite{Friston2011opt}. Intuitively, one could model an extrinsic movement as in Figure \ref{fig:duplicated}, i.e., with the following dynamics and likelihood functions:
	\begin{align}
		\begin{split}
			\label{eq:duplicated}
			\bm{f}(\bm{x},\bm{v}) &= \bm{J}^T (\bm{v} - \bm{T}(\bm{x})) \\
			\bm{g}_p(\bm{x}) &= \bm{x} \\
			\bm{g}_v(\bm{x},\bm{v}) &= \begin{bmatrix} \bm{T}(\bm{x}) & \bm{v} \end{bmatrix}
		\end{split}
	\end{align}
	where $\bm{x}$ are the arm joint angles, $\bm{v}$ is the target position to reach, $\bm{T}$ is the forward kinematics returning the hand position, and $\bm{J}$ is its Jacobian matrix.
	
	The visual likelihood function $\bm{g}_v$ generates visual predictions for the hand and the target through forward kinematics and identity map, respectively. For goal-directed behavior, first an error between the target and hand positions is generated; then, an inverse kinematic model is embedded directly into the dynamics function, e.g., a Jacobian transpose or a pseudoinverse  \cite{Sancaktar2020,Oliver2021,Meo2021,Lanillos2020, Friston2010,Friston2011b,Pio-Lopez2016}. In other words, an extrinsic reference frame is inverted to generate an intrinsic state, which is in turn transformed again in the first domain to be compared with visual observations. As a result, forward and inverse kinematics are performed twice, once in the dynamics function and once in the forward and backward passes of visual inference, i.e., when backpropagating the visual prediction error $\bm{\varepsilon}_{o,v}$:
	\begin{equation}
		\label{eq:inv_kin}
		\partial \bm{g}_v^T \bm{\varepsilon}_{o,v} = \begin{bmatrix}
			\bm{J} \\ \bm{1}
		\end{bmatrix} (\bm{o}_v - \begin{bmatrix}
			\bm{T}(\bm{x}) & \bm{v}
		\end{bmatrix})
	\end{equation}
	If the predictions are not temporarily stored, this requires increased computational demand and memory. Additionally, there is an issue regarding biological plausibility: using sensory-level attractors within the dynamics function means that a unit is aware of part of the likelihood prediction -- generally assumed to go all down to the sensorium -- and its inverse mapping, which are lower-level features. Finally, with the model in Figure \ref{fig:duplicated} the agent cannot easily express paths in extrinsic coordinates needed, e.g., for realizing linear or circular motions, or for imposing constraints in both intrinsic and extrinsic domains such as when walking with a glass in hand.
	\begin{figure}[t]
		\centering
		\begin{subfigure}{0.28\textwidth}
			\centering
			\includegraphics[width=\textwidth]{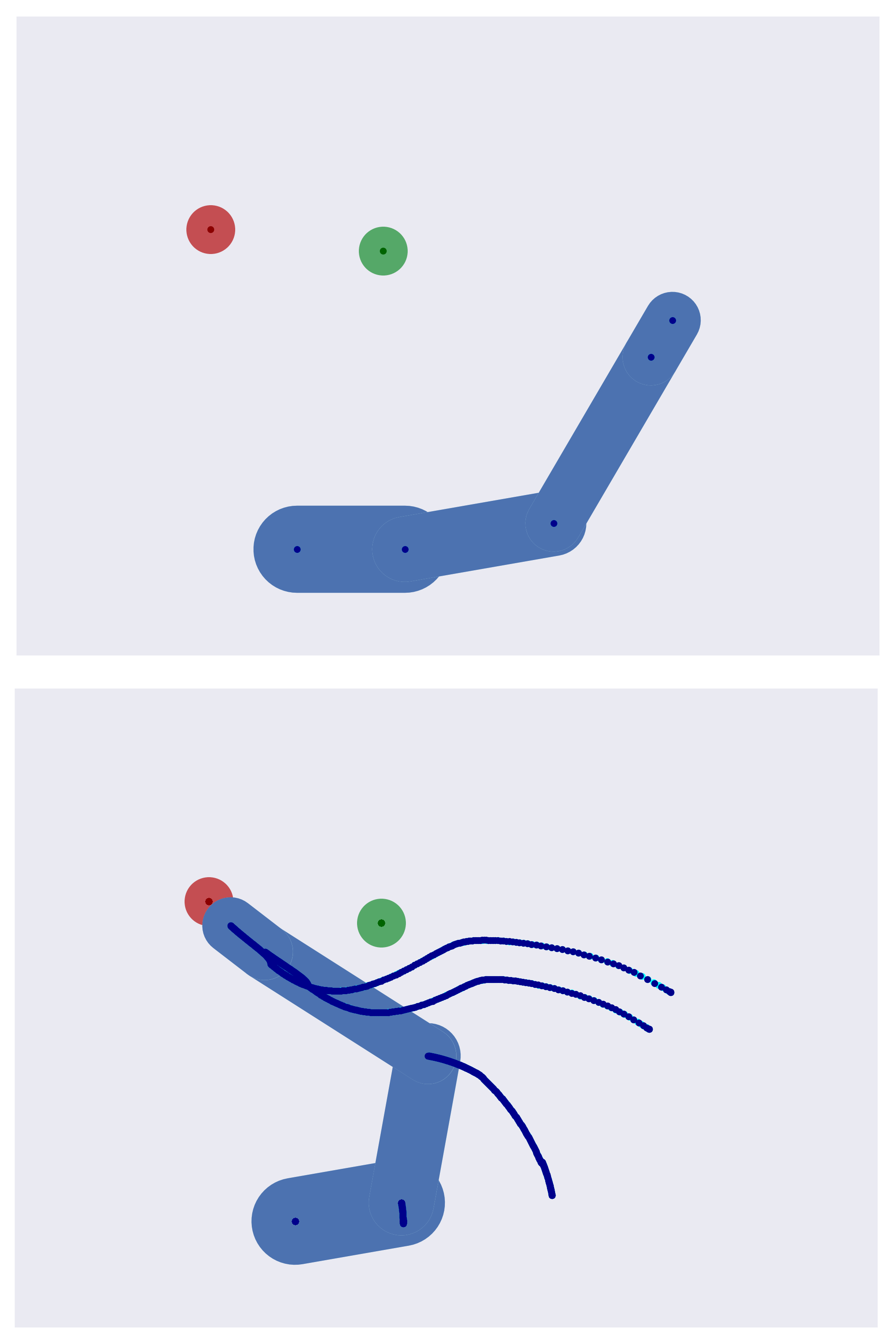}
			\caption{}
			\label{fig:example_reachavoid}
		\end{subfigure}
		\begin{subfigure}{0.28\textwidth}
			\centering
			\includegraphics[width=\textwidth]{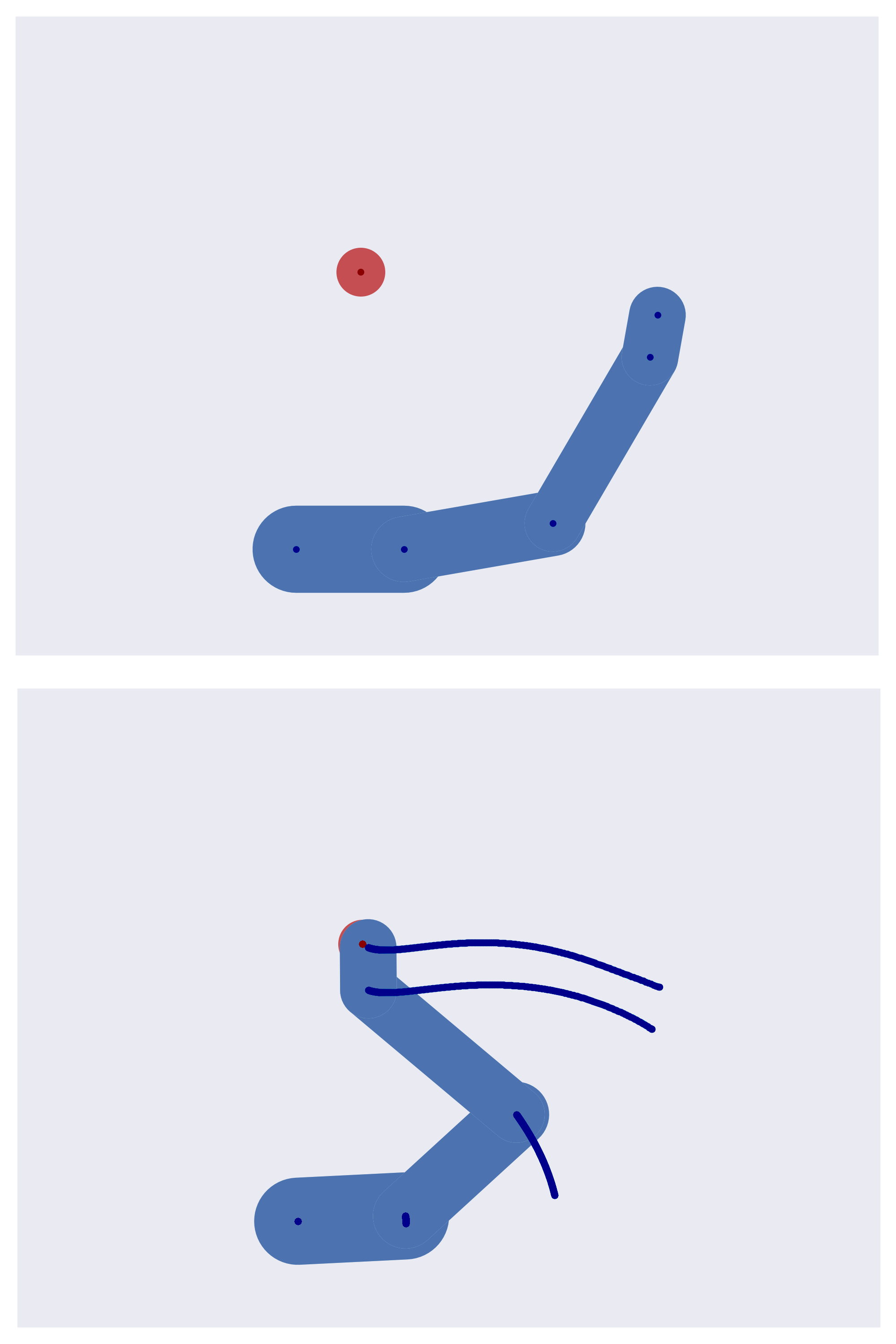}
			\caption{}
			\label{fig:example_maintain}
		\end{subfigure}
		\begin{subfigure}{0.28\textwidth}
			\centering
			\includegraphics[width=\textwidth]{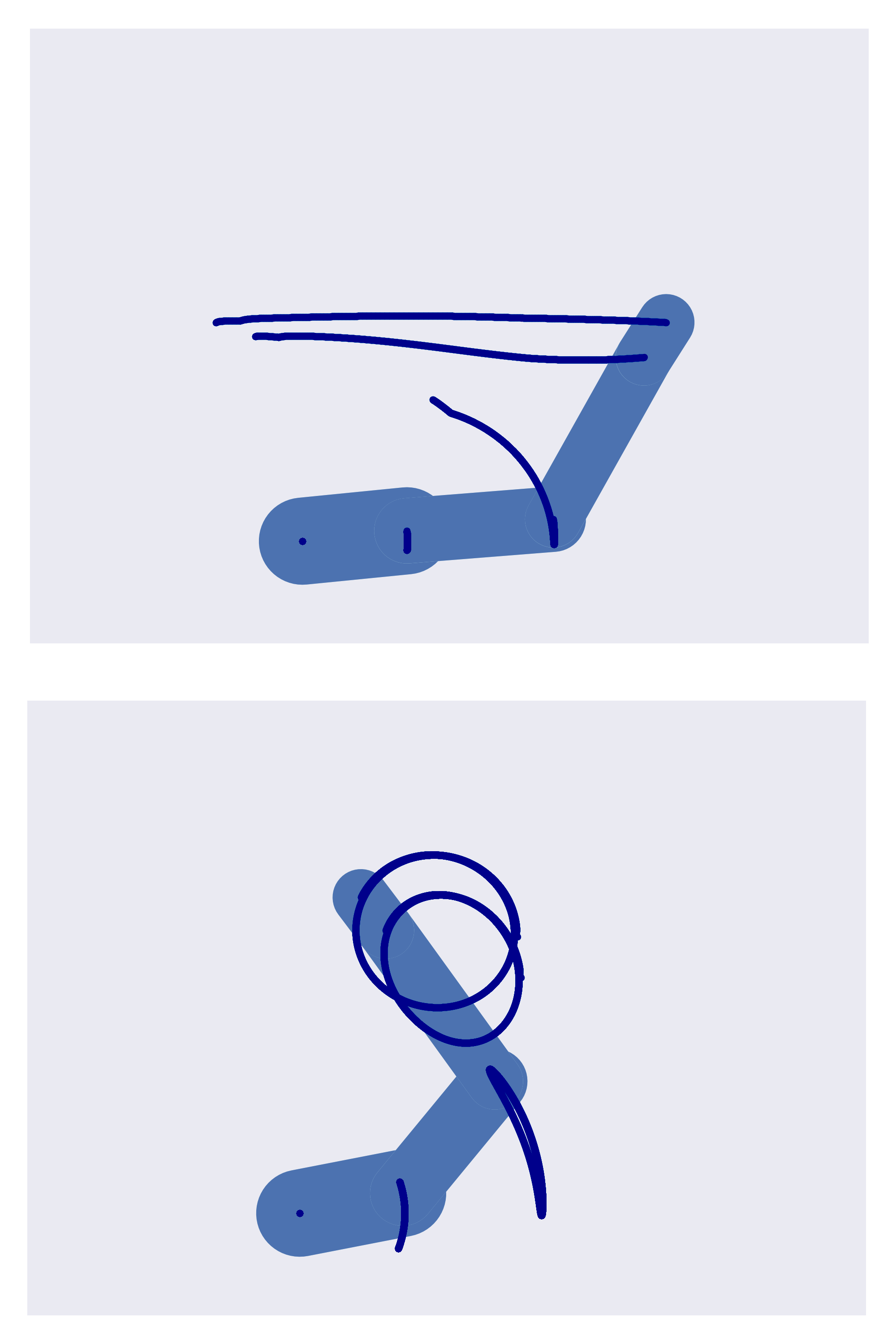}
			\caption{}
			\label{fig:example_circular}
		\end{subfigure}
		\caption{\textbf{(a)} In this task, the agent (a 4-DoF arm) has to reach the red target while avoiding the green obstacle; this is possible by specifying two (attractive and repulsive) functions at the extrinsic level. \textbf{(b)} In this task, the agent has to reach the red target while maintaining the same hand orientation (such as when walking with a glass in hand); this is possible by combining intrinsic and extrinsic constraints. \textbf{(c)} In this task, the agent has to perform linear (top) and circular (bottom) motions, possible by defining attractors at the 1st temporal order of the extrinsic hidden states. See \cite{Priorelli2023b} for more details.}
		\label{fig:example_extrinsic}
	\end{figure}
	
	We can instead exploit forward and inverse kinematics of Equation \ref{eq:inv_kin} and follow the natural flow of the generative process to avoid duplicated computations, as displayed in Figure \ref{fig:inv_kin}. The alternative generative model is displayed in Figure \ref{fig:gen_model_ie}. This model relies on two hierarchical levels, where an intrinsic unit (encoding the arm joint angles) is placed at the top and generates predictions through forward kinematics for an extrinsic unit (encoding the Cartesian position of the target) \cite{Priorelli2023b}:
	\begin{equation}
		\bm{x}_e = \bm{g}_e(\bm{x}_i) + \bm{w}_{o,e} = \bm{T}(\bm{x}_i) + \bm{w}_{o,e}
	\end{equation}
	while the visual likelihood becomes a simple identity map, i.e., $\bm{g}_v(\bm{x}_e) = \bm{x}_e$. The goal-directed behavior from the dynamics function of Equation \ref{eq:duplicated} arises naturally via backpropagation of the \textit{extrinsic prediction error} $\bm{\varepsilon}_{o,e}$:
	\begin{equation}
		\label{eq:new_kin}
		\partial \bm{g}_e^T \bm{\varepsilon}_{o,e} = \bm{J}^T (\bm{\mu}_{x,e} - \bm{T}(\bm{\mu}_{x,i}))
	\end{equation}
	Having a complete unit that deals with extrinsic information -- which is thus not embedded into the hidden causes of the intrinsic unit -- allows the agent to specify its dynamics, leading to an efficient decomposition between intrinsic and extrinsic attractors -- i.e., $\bm{f}_i(\bm{x}_i, \bm{v}_i)$ and $\bm{f}_e(\bm{x}_e, \bm{v}_e)$ -- and between proprioceptive and visual observations -- as exemplified in the simulations of Figure \ref{fig:example_extrinsic}. Note the similarity of Equation \ref{eq:new_kin} with Equations \ref{eq:duplicated} and \ref{eq:inv_kin}: if in the model of Figure \ref{fig:duplicated} we had two different forward and inverse kinematics either for goal-directed behavior or for predicting current observations, in the new model of Figure \ref{fig:inv_kin} what is compared with the observations already contains a bias toward intentional states, without the need for sensory-level attractors within the dynamics function. The extrinsic prediction error of Equation \ref{eq:new_kin} becomes zero only when the extrinsic hidden states $\bm{x}_e$ match the predictions of the intrinsic unit; if such predictions are not met, $\bm{\varepsilon}_{o,e}$ will flow through the hierarchy eventually generating an action. The update rules for the 0th orders of the intrinsic and extrinsic hidden states are the following:
	\begin{align}
		\dot{\bm{\mu}}_{x,i} &= \bm{\mu}_{x,i}^\prime - \bm{\Pi}_{\eta,x,i} \bm{\varepsilon}_{\eta,x,i} + \partial_{x_{i}} \bm{g}_p^T \bm{\Pi}_{o,p} \bm{\varepsilon}_{o,p} + \partial_{x_{i}} \bm{g}_e^T \bm{\Pi}_{o,e} \bm{\varepsilon}_{o,e} + \partial_{x_{i}i} \bm{f}_i^T \bm{\Pi}_{x,i} \bm{\varepsilon}_{x,i} \\
		\dot{\bm{\mu}}_{x,e} &= \bm{\mu}_{x,e}^\prime - \bm{\Pi}_{o,e} \bm{\varepsilon}_{o,e} + \partial_{x_{e}} \bm{g}_v^T \bm{\Pi}_{o,v} \bm{\varepsilon}_{o,v} + \partial_{x_{e}} \bm{f}_e^T \bm{\Pi}_{x,e} \bm{\varepsilon}_{x,e}
	\end{align}
	Although the generative model follows the forward flow of optimal control, the relationship between proprioceptive consequences and extrinsic causes peculiar to active inference still holds because the kinematic inversion regards a high-level process that manipulates abstract (intrinsic or extrinsic) representations, and both of them concur to generate low-level proprioceptive states. As Adams and colleagues note, ``\textit{The key distinction is not about mapping from desired states in an extrinsic (kinematic) frame to an intrinsic (dynamic) frame of reference, but the mapping from desired states (in either frame) to motor commands}'' \cite{Adams2013}. Having said this, there is a significant difference between the two models represented in Figure \ref{fig:kin_models}, which can be compared to the two supervised learning modes of predictive coding \cite{millidge2022predictive}: a forward mode that fixes the latent states to the labels and the observations to the data can generate highly accurate images of digits, while the inverse classification task is more difficult as there is no univocal mapping between labels and data; instead, a backward mode that fixes the latent states to the data and the observations to the labels achieves high performances on classification but falls short when generating images. Based on this, we can interpret the model of Figure \ref{fig:duplicated} as a backward mode that would rapidly generate a proper kinematic configuration with the hand at the target, but that would hardly infer from proprioception the hand position needed to plan movements. Conversely, we can interpret the model of Figure \ref{fig:inv_kin} as a forward mode that would generate with high accuracy the hand position, but that would find it difficult to infer the kinematic configuration needed to actually realize movement.
	
	
	\subsection{\label{sec:hier}A module for iterative transformations}
	
\begin{figure}
		\begin{minipage}[b]{\textwidth}
			\centering
			\subfloat
			[]
			{\label{fig:hierarchies_cont}\vspace{0em}\includegraphics[height=0.26\textheight]{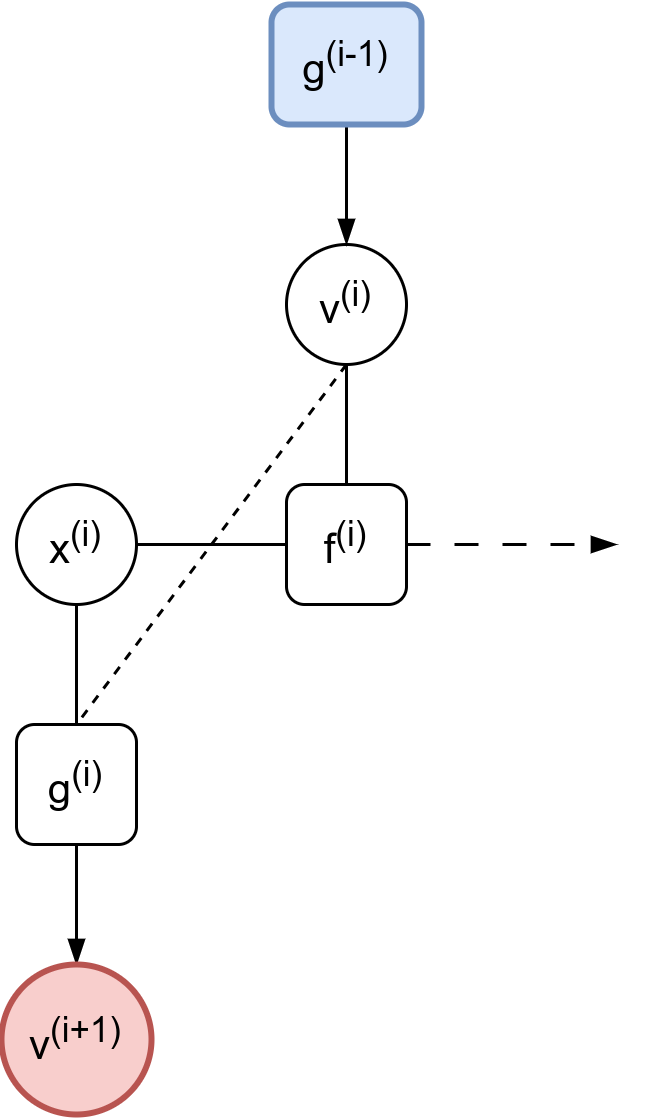}}
			\hspace{2em}
			\subfloat
			[]
			{\label{fig:hierarchies_disc}\vspace{0em}\includegraphics[height=0.26\textheight]{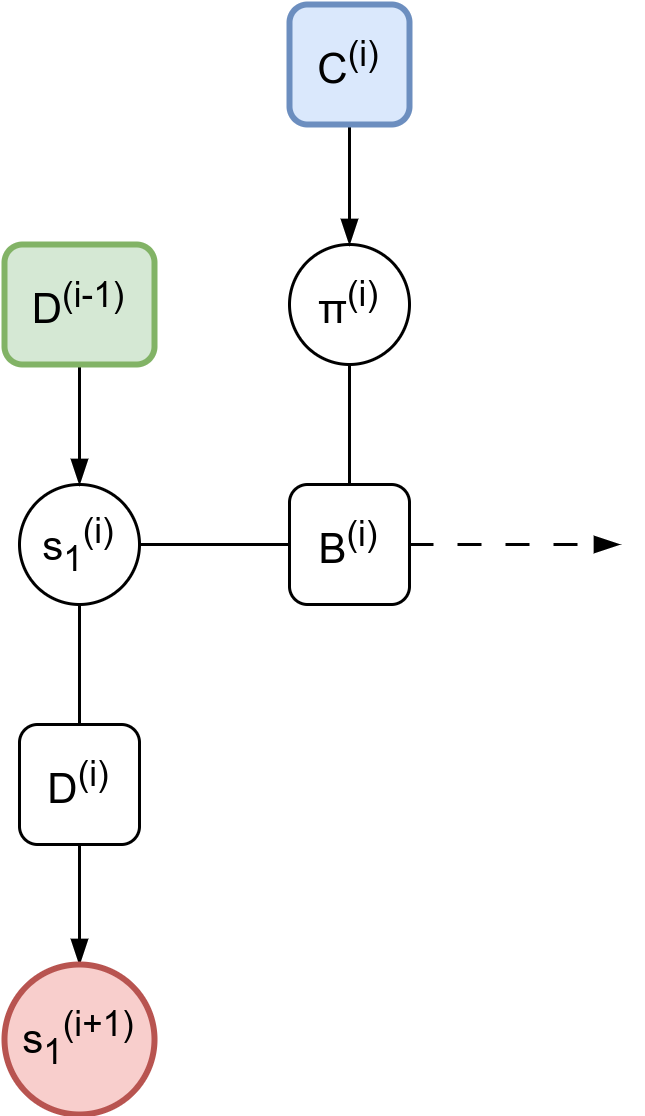}}
			\hspace{2em}
			\subfloat
			[]
			{\label{fig:hierarchies_pcn}\vspace{0em}\includegraphics[height=0.25\textheight]{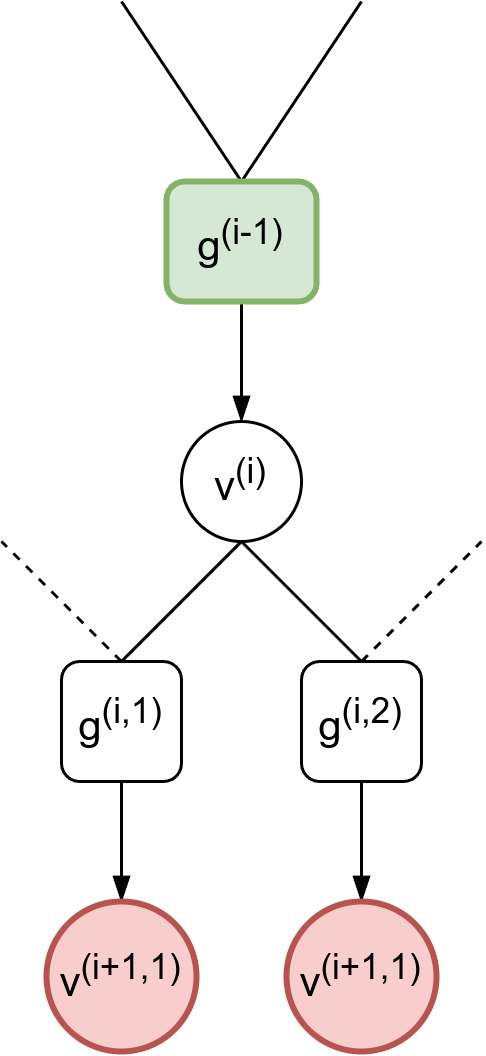}}
			\hspace{2em}
			\subfloat
			[]
			{\label{fig:hierarchies_new}\vspace{0em}\includegraphics[height=0.26\textheight]{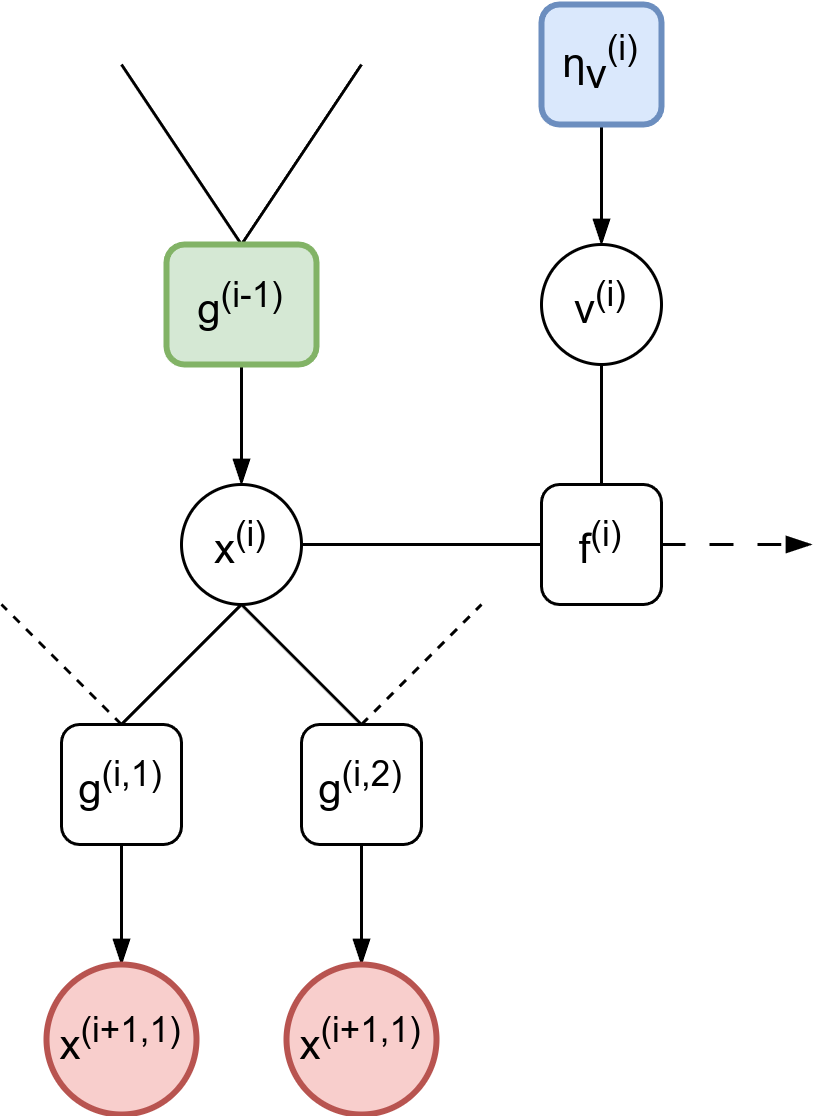}}
			\vspace{0em}
		\end{minipage}
		\begin{minipage}[b]{\textwidth}
			\centering
			\subfloat
			[]
			{\label{fig:ie_module}\vspace{0em}\includegraphics[width=0.75\textwidth]{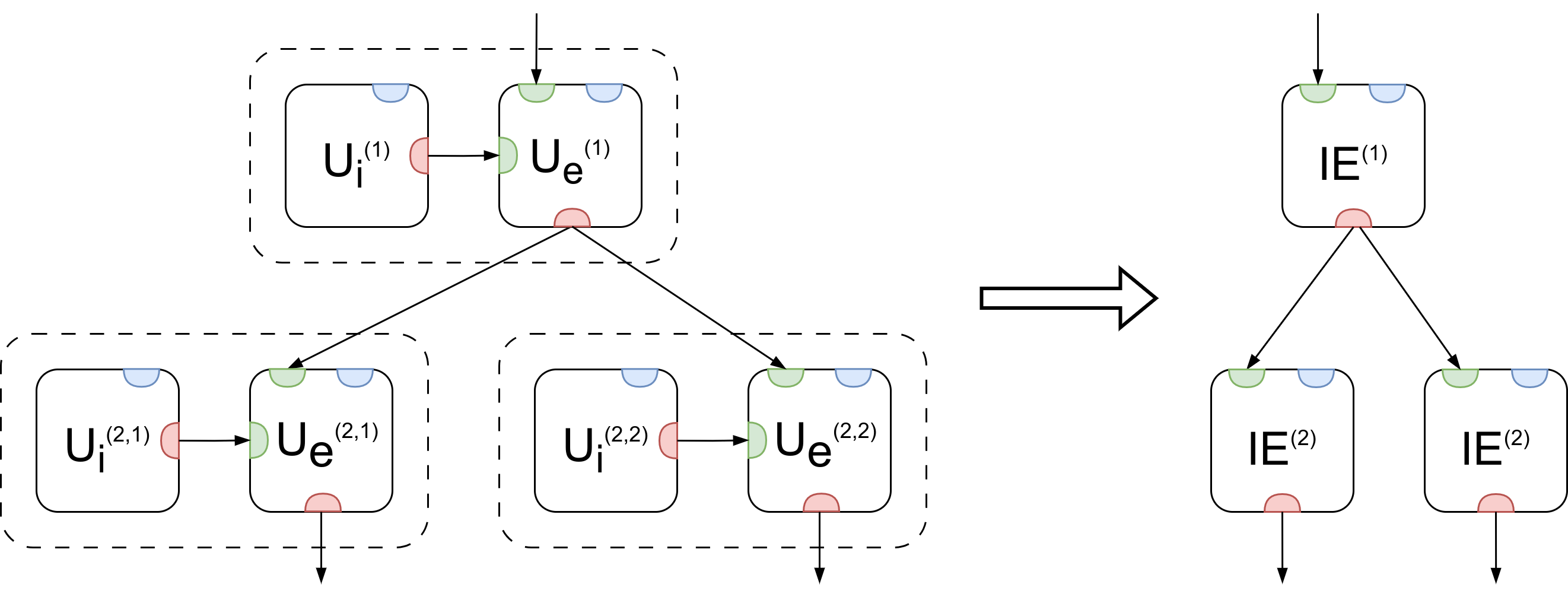}}
			\vspace{0em}
		\end{minipage}
		\begin{minipage}[b]{1.25\textwidth}
			\centering
			\subfloat
			[]
			{\label{fig:gen_model_hier}\vspace{0em}\includegraphics[width=\textwidth]{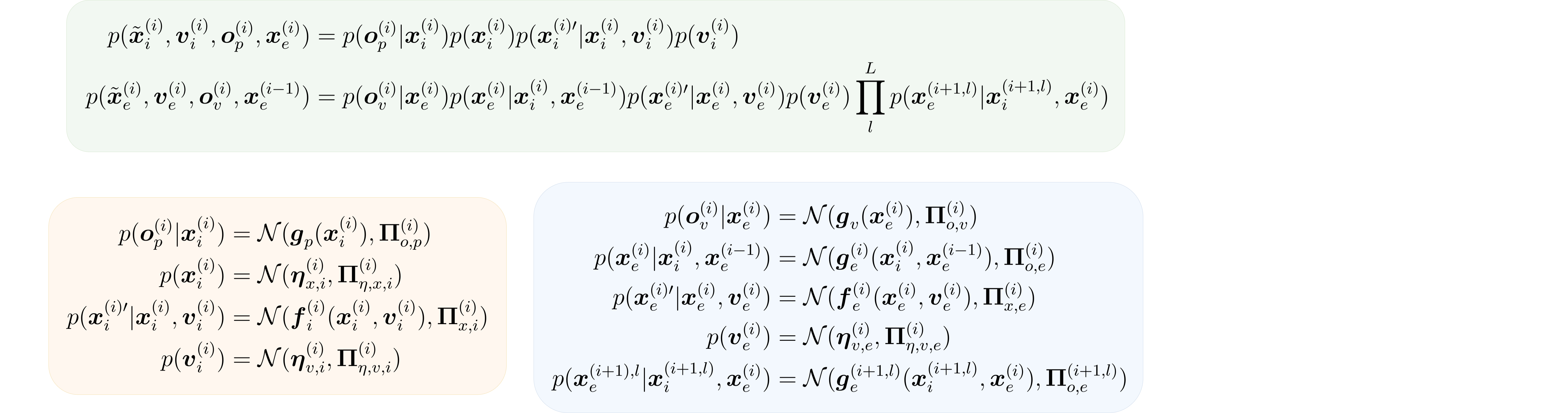}}
		\end{minipage}
		\caption{\textbf{(a)} Factor graph of the typical connection between two continuous levels. Hidden states $\bm{x}^{(i)}$ and hidden causes $\bm{v}^{(i)}$ of level $i$ generate -- through the likelihood function $\bm{g}^{(i)}$ -- the hidden causes $\bm{v}^{(i+1)}$ of level $i+1$. \textbf{(b)} Connections between two discrete levels. Hidden states $\bm{s}_1^{(i)}$ of level $i$ generate -- through the prior matrix $\bm{D}^{(i)}$ -- the hidden states $\bm{s}_1^{(i+1)}$ of level $i+1$. We will cover discrete models in the next chapter. \textbf{(c)} Connections between two levels in PCNs. The likelihood function $\bm{g}^{(i)}$ performs a simple combination of neurons, passed to a nonlinear activation function. Standard PCNs only permit representing the causal structure of a system, without modeling internal dynamics. \textbf{(d)} Factor graph of a level with multiple inputs and outputs from independent units, with similar connectivity to temporal predictive coding. The observation $\bm{o}^{(i,j)}$ becomes the 0th-order hidden state $\bm{x}^{(i+1,j)}$ of the level below, while the prior $\bm{\eta}_x^{(i,j)}$ becomes the 0th-order hidden state $\bm{x}^{(i-1,j)}$ of the level above. \textbf{(e)} A network of IE modules. An extrinsic $\bm{x}_e^{(i-1)}$, along with an intrinsic signal $\bm{x}_i^{(i,j)}$ (e.g., angle for rotation or length for translation), is passed to a function $\bm{g}_e^{(i,j)}$, generating a new extrinsic signal $\bm{x}_e^{(i,j)}$. \textbf{(f)} Generative model of (e).}
	\end{figure}

	The model in Figure \ref{fig:inv_kin} introduced a hierarchical dependency between two (intrinsic and extrinsic) levels, made possible by connecting hidden states. Instead, the typical approach in continuous-time active inference involves connections between hidden states and causes of a level and hidden causes (and not hidden states) of the subordinate level, as shown in Figure \ref{fig:hierarchies_cont}. While this allows one to impose a trajectory for the unit below, specifying fixed setpoints to the 0th-order hidden states is not as straightforward, since the dynamics prediction error generated from the hidden causes would have to travel back to the previous temporal orders. As clear from Figure \ref{fig:inv_kin}, a connection between hidden states is of high utility when designing hierarchical models. In fact -- as represented in Figure \ref{fig:hierarchies_disc} -- it is fundamental in defining the initial states of different temporal scales in discrete models, e.g., for pictographic reading \cite{Friston2020} or linguistic communication \cite{Friston2020}. Similar connections are used in standard PCNs, wherein each neuron of a level computes a combination of neurons of the level above passed to an activation function \cite{millidge2022predictive} -- as shown in Figure \ref{fig:hierarchies_pcn}.
	
	Recall that, in the previous kinematic model, the prediction $\bm{g}_e(\bm{x}_i)$ of the intrinsic hidden states acted as a prior for the extrinsic hidden states, while the latter acted as an observation for the intrinsic hidden states. Following this example, we use the observation of a level to bias the 0th-order hidden states of the level below directly:
	\begin{align}
		\bm{\eta}_x^{(i+1)} &\equiv \bm{g}^{(i)}(\bm{\mu}_x^{(i)}) \\
		\bm{o}^{(i)} &\equiv \bm{x}^{(i+1)}
	\end{align}
	As a result, the observation prediction error $\bm{\varepsilon}_{o}$ and the prior prediction error $\bm{\varepsilon}_{\eta,x}$ of Equation \ref{eq:pred_error} is expressed by the same variable:
	\begin{equation}
		\bm{\varepsilon}_o^{(i)} = \bm{\varepsilon}_{\eta,x}^{(i+1)} = \bm{\mu}_x^{(i+1)} - \bm{g}^{(i)}(\bm{\mu}_x^{(i)})
	\end{equation}
	where the hierarchical level is indicated with a superscript and lower levels are denoted by increasing numbers. We can then design a multiple-input and multiple-output system wherein a level imposes and receives priors and observations to several independent units, as in Figure \ref{fig:hierarchies_new}. The computation of the free energy in Equation \ref{eq:free_energy} remains unchanged, and the update of the hidden states turns into the following:
	\begin{equation}
		\dot{\tilde{\bm{\mu}}}_x^{(i,j)} = \begin{bmatrix} \bm{\mu}_x^{\prime{(i,j)}} - \bm{\Pi}_{o}^{(i-1)} \bm{\varepsilon}_{o}^{(i-1)} + \sum_l \partial_{x^{(i,j)}} \bm{g}^{(i,l)T} \bm{\Pi}_o^{(i,l)} \bm{\varepsilon}_o^{(i,l)} + \partial_{x^{(i,j)}} \bm{f}^{(i,j)T} \bm{\Pi}^{(i,j)}_x \bm{\varepsilon}^{(i,j)}_x \\ \\ - \bm{\Pi}^{(i,j)}_x \bm{\varepsilon}^{(i,j)}_x \end{bmatrix}
	\end{equation}
	where the superscript notation $(i,j)$ indicates the $i$th hierarchical level and the $j$th element within the same level. As evident, this is a similar connectivity to hierarchical dynamical models in temporal predictive coding \cite{Friston2009,Millidge2023}. Here, the forward prediction error $\bm{\varepsilon}_{o}^{(i-1)}$ combines the predictions from the units above and acts as prior (the gradient is absent since they are expressed in the same domain of the unit considered), while the backward prediction errors $\bm{\varepsilon}_o^{(i,l)}$ contain the observations from the units below.
	
	What advantages do deep hierarchical models carry compared to a shallow agent? Although the structure of Figure \ref{fig:inv_kin} affords a more advanced control with respect to the model in Figure \ref{fig:duplicated}, its uses are still limited to solving simple tasks, e.g., performing operations with the hand. While simultaneous coordination of multiple limbs is possible, it would require complex dynamics functions, with complexity increasing with the number of joints and ramifications of the kinematic chain. Critically, a shallow agent would not be capable of capturing the hierarchical causal relationships inherent to the generative process, allowing one to predict and anticipate the local exchange of forces that would unfold during movement. As mentioned in the Introduction, a deep model is also required if one has to use tools for manipulation tasks. Besides roto-translations in forward kinematics, iterative transformations are also essential in computer vision -- where an image can be subject to scaling, shearing, or projection -- and, more in general, whenever changing the basis of a coordinate vector.
	
	For these reasons, we can generalize the last model and construct an \textit{Intrinsic-Extrinsic} (or IE) module \cite{Priorelli2023b,Priorelli2023c,Priorelli2024a}. This module is composed of two units and its role is to perform iterative transformations between reference frames. In brief, a unit $\mathcal{U}_e^{(i-1)}$ encodes a signal in an extrinsic reference frame, while a second unit $\mathcal{U}_i^{(i)}$ contains a generic intrinsic transformation. Applying the latter to the first signal returns a new extrinsic reference frame embedded in a unit $\mathcal{U}_e^{(i)}$. Then, we can define a likelihood function $\bm{g}_e$ such that:
	\begin{equation}
		\label{eq:kin_lkh}
		\bm{x}_e^{(i)} = \bm{g}_e^{(i)}(\bm{x}_i^{(i)}, \bm{x}_e^{(i-1)}) + \bm{w}_{o,e}^{(i)} = \bm{T}^{(i)}(\bm{x}_i^{(i)})^T \bm{x}_e^{(i-1)} + \bm{w}_{o,e}^{(i)}
	\end{equation}
	where $\bm{w}_{o,e}^{(i)}$ is a noise term, while $\bm{T}^{(i)}$ here indicates a linear transformation matrix, which can express forward kinematics as in the previous examples or non-affine transformations. Backpropagating the extrinsic prediction error $\bm{\varepsilon}_{o,e}^{(i)} = \bm{\mu}_{x,e}^{(i)} - \bm{g}_e(\bm{\mu}_{x,i}^{(i)}, \bm{\mu}_{x,e}^{(i-1)})$ leads to simple belief updates:
	\begin{align}
		\begin{split}
			\partial_{x_{e}^{(i-1)}} \bm{g}_e^{(i)T} \bm{\varepsilon}_{o,e}^{(i)} &= \bm{T}^{(i)T} \bm{\varepsilon}_{o,e}^{(i)} \\\\
			\partial_{x_{i}^{(i)}} \bm{g}_e^{(i)T} \bm{\varepsilon}_{o,e}^{(i)} &= \partial_{x_{i}^{(i)}} \bm{T}^{(i)} \odot [\bm{\varepsilon}_{o,e}^{(i)} \bm{\mu}_{x,e}^{(i-1)T}]
		\end{split}
	\end{align}
	where $\odot$ is the element-wise product. 
	\begin{figure}[t]
		\centering
		\begin{subfigure}{0.3\textwidth}
			\centering
			\includegraphics[height=0.33\textheight]{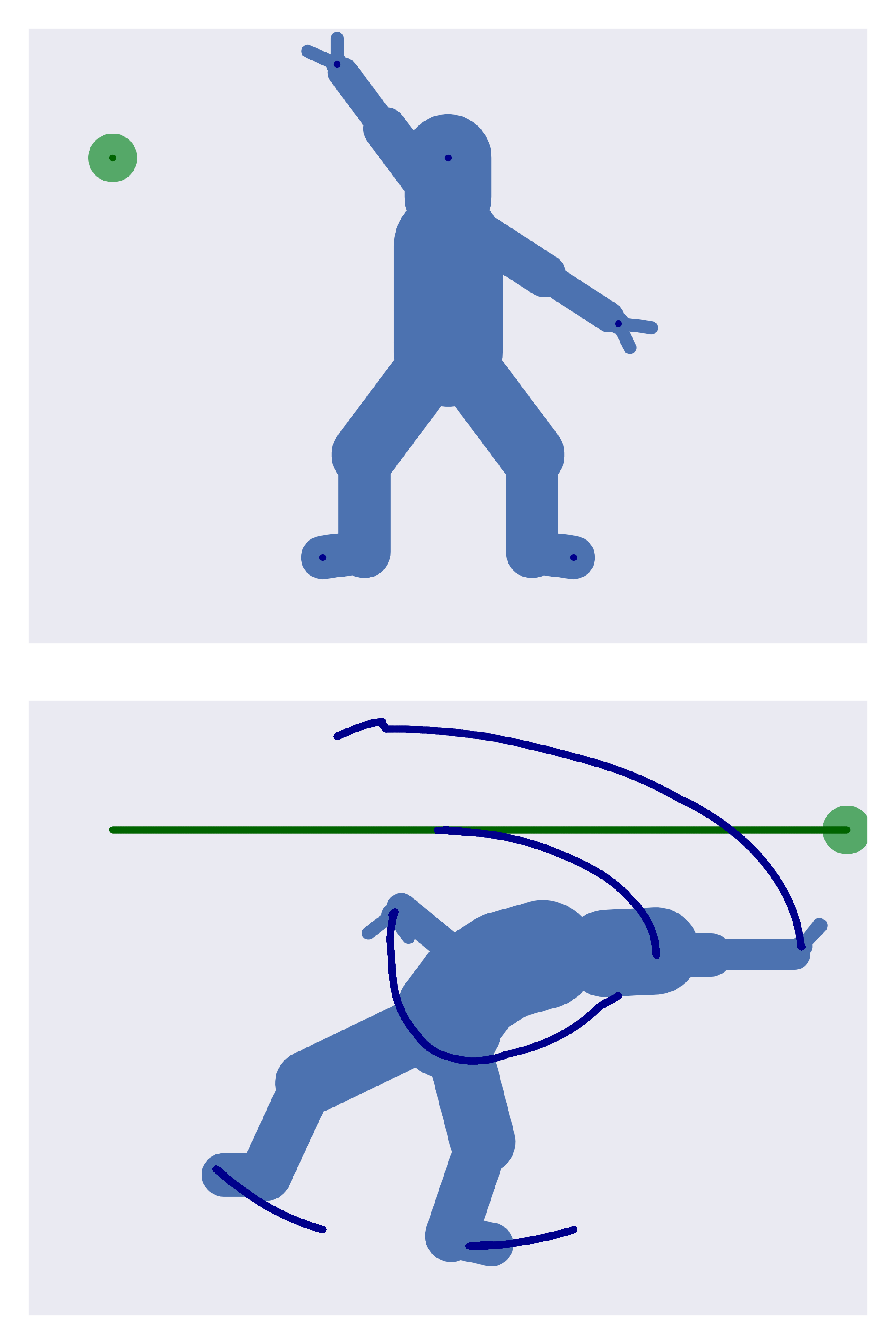}
			\caption{}
			\label{fig:example_matrix}
		\end{subfigure}
		\begin{subfigure}{0.3\textwidth}
			\centering
			\includegraphics[height=0.325\textheight]{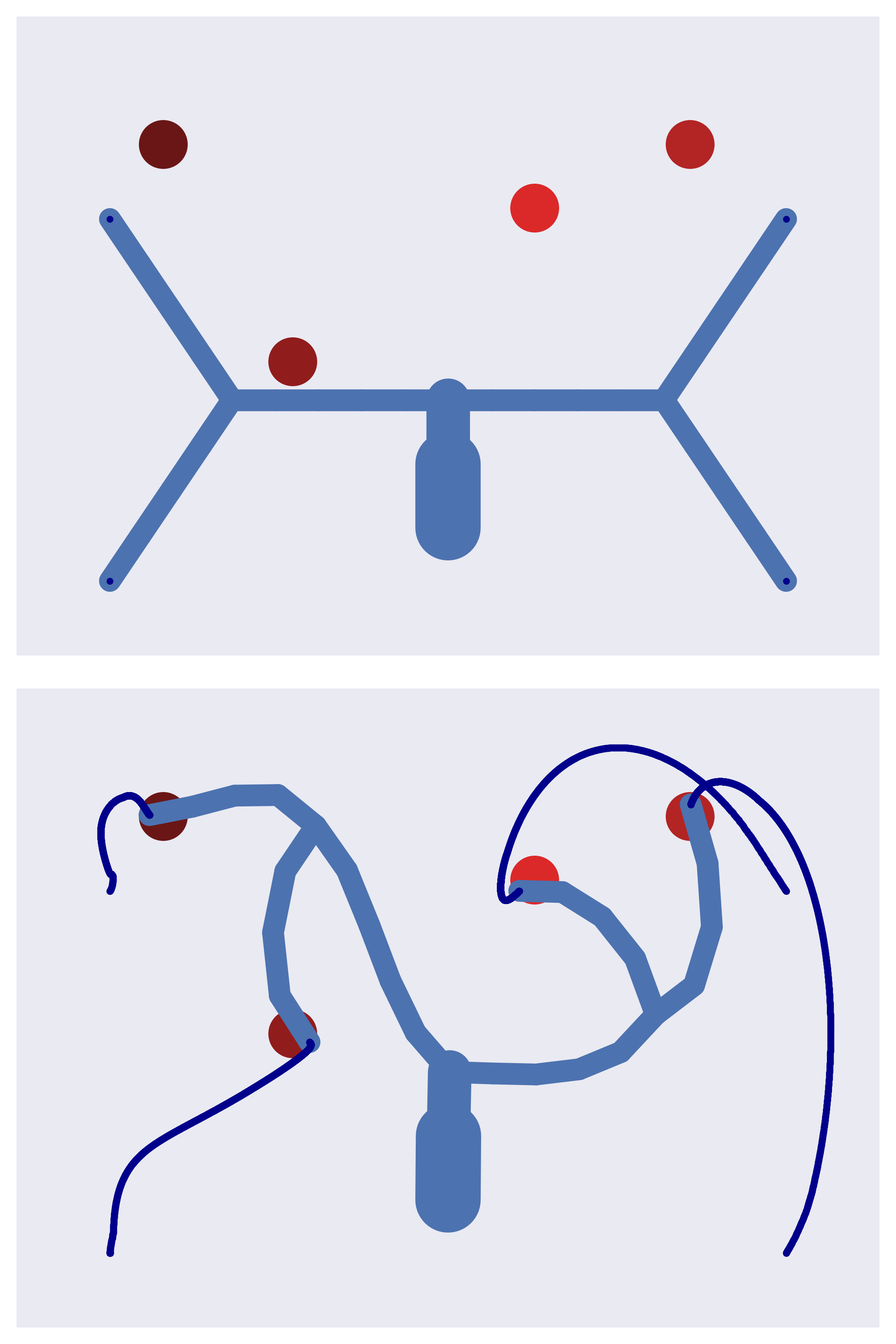}
			\caption{}
			\label{fig:example_octopus}
		\end{subfigure}
		\begin{subfigure}{0.3\textwidth}
			\centering
			\includegraphics[height=0.33\textheight]{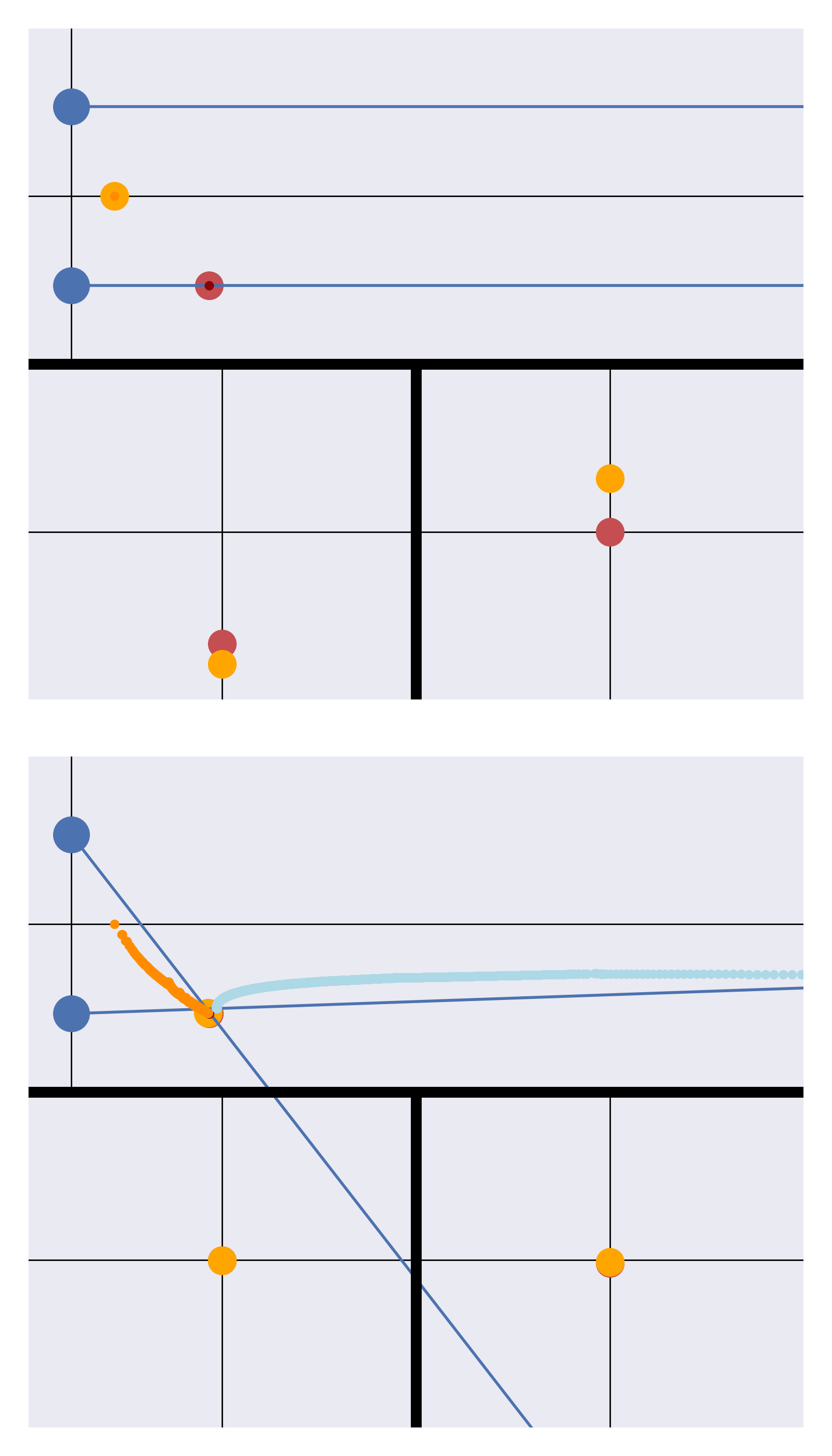}
			\caption{}
			\label{fig:example_camera}
		\end{subfigure}
		\label{fig:example_ie_tasks}
		\caption{\textbf{(a)} In this task, the agent (a 23-DoF human body) has to avoid a moving obstacle; this is possible by defining a repulsive attractor for each extrinsic level. \textbf{(b)} In this task, the agent (a 28-DoF kinematic tree) has to reach four target locations with the extremities of its branches. See \cite{Priorelli2023b} for more details. \textbf{(c)} In this task, the agent (the two blue eyes on the left of the top view) has to infer the depth of the red object while fixating on it. The inferred position (and its trajectory) is represented in orange, while the blue trajectory is the center of fixation of the eyes. The bottom two frames show the object projection in the eye planes. See \cite{Priorelli2023c} for more details.}
	\end{figure}
	These equations express the most likely intrinsic and extrinsic states that may have generated the new reference frame. As shown in Figure \ref{fig:ie_module}, modules are linked through the extrinsic units $\mathcal{U}_e^{(i)}$, while $\mathcal{U}_i^{(i)}$ performs an internal operation and does not contribute to the hierarchical connectivity. For motor control, we can realize a hierarchical multiple-output system, wherein the intrinsic hidden states $\bm{x}_i^{(i,j)}$ of a level encode a pair of joint angle and limb length of a single DoF. Iteratively applying roto-translations to an origin (e.g., body-centered) reference frame $\bm{x}_e^{(0)}$ -- consisting of a Cartesian position and an absolute orientation -- will determine the kinematic configuration of the agent in terms of extrinsic coordinates \cite{Priorelli2023b}. The generative model for the intrinsic and extrinsic units of a single IE module at level $i$ are displayed in Figure \ref{fig:gen_model_hier}. Compared to the single-level generative model of Figure \ref{fig:gen_model_ie} comprising every joint angle of the agent's body, here we have a dependency between levels encoding distinct joint angles. In particular, an additional term that links the extrinsic unit with the previous level, i.e., $p(\bm{x}^{(i)}_e | \bm{x}^{(i)}_i, \bm{x}_e^{(i-1)})$, and multiple distributions $p(\bm{x}_e^{(i+1,l)} | \bm{x}_i^{(i+1,l)}, \bm{x}_e^{(i)})$ generating extrinsic predictions for different modules at the next level.
	
	At this point, we can easily express how every single joint and limb would evolve, affording a highly advanced control as demonstrated by the simulations of Figures \ref{fig:example_matrix} and \ref{fig:example_octopus}. Besides modeling limb dynamics, the IE modules can also be applied to other linear transformations, e.g., perspective projections. As displayed in Figure \ref{fig:example_camera}, this can be useful for estimating the depth of an object via parallel predictions (e.g., from the eyes or multiple cameras) \cite{Priorelli2023c} -- a process that active inference casts in terms of target fixation and hypothesis testing \cite{Parr2018c}. The modularity of this architecture allows the agent to define dynamic attractors in the 2D projected planes, in the 3D reference frames of the eyes, or as simple vergence-accommodation angles. This approach also has some analogies with Active Predictive Coding \cite{rao2022active} and Recursive Neural Programs \cite{Fisher2023}, which addressed the part-whole hierarchy learning problem in computer vision by recursively applying reference frame transformations to parts of a scene.
	
	
	\subsection{The self, the objects, and the others}
	
	Describing Figure \ref{fig:inv_kin}, we passed over a critical mechanism introduced at the beginning: the characterization of object affordances. Recall that the hidden states encoded in parallel not only the self but also other environmental entities; however, the agent's model can now express the generative process hierarchically. This is described by the following likelihood function:
	\begin{equation}
		\bm{g}_e^{(i)}(\bm{x}_i^{(i)}, \bm{x}_e^{(i-1)}) = \begin{bmatrix} \bm{T}^{(i)}(\bm{x}_{i,0}^{(i)})^T \bm{x}_{e,0}^{(i-1)} & \bm{T}^{(i)}(\bm{x}_{i,1}^{(i)})^T \bm{x}_{e,1}^{(i-1)} & \dots & \bm{T}^{(i)}(\bm{x}_{i,N}^{(i)})^T \bm{x}_{e,N}^{(i-1)} \end{bmatrix}
	\end{equation}
	in which every IE module of the hierarchy has distinct factors for the self and every entity. For the self, this has a simple explanation, i.e., it just generates, one after the other, the positions of every segment of the kinematic chain depending on its joint angles. Concerning an object, we could encode its Cartesian position by attaching its visual observation to a second factor of the hidden states at a specific level. If the generative model has the same hierarchy for both the self and the object, backpropagating the extrinsic prediction errors of this second component will eventually infer a potential agent's configuration in relation to the object, as before. For instance, if the object is linked to the last (i.e., hand) level, this would represent the hand at the object location, while all the previous levels would represent appropriate intermediate positions and angles generating that final location. In other words, the additional factorizations of hidden states and likelihoods reflect here a (deep) hierarchical configuration of the self that the agent thinks to be suitable for object manipulation. Since each level can express some dynamics through its hidden causes, the inference of this potential configuration is steered to match the object's affordances and the agent's intentions. As will be shown in the next chapter, this permits flexible adaptation of the kinematic chain depending on the circumstances, as well as representing the hierarchical structures of objects (e.g., tools). The inferred beliefs would be subject only to exteroceptive information from the objects, while proprioceptive states would be used only to update the agent's belief of its current configuration.
	
	\begin{figure}[h]
		\centering
		\includegraphics[width=0.98\textwidth]{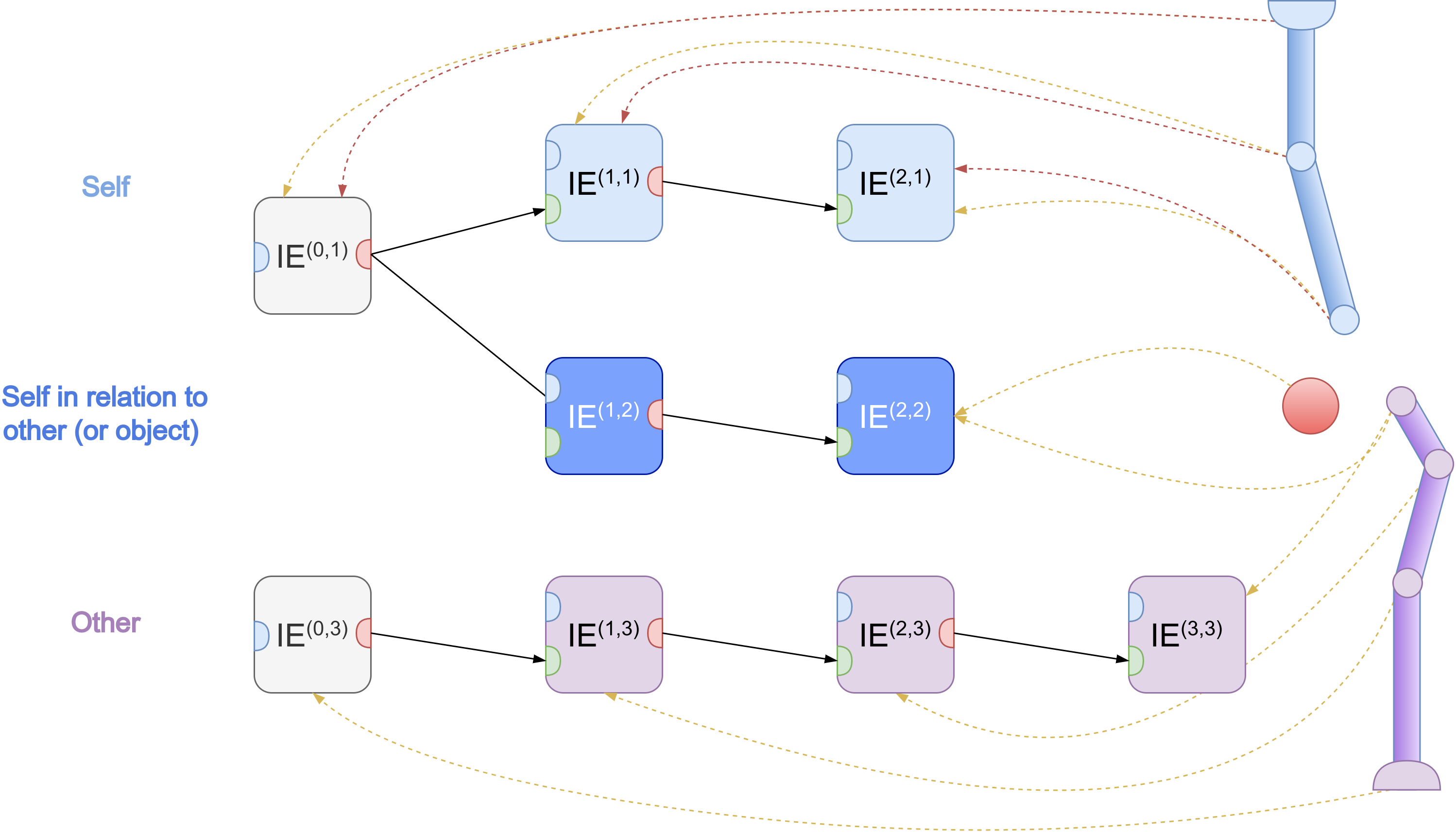}
		\caption{Interactions between an agent (a 2-DoF light-blue arm), another agent (a 3-DoF purple arm), and an object (a red ball). Generative model of the first agent, composed of three parallel pathways representing the kinematic structures of both agents. For clarity, lateral connections among the model components of each level are not shown. Both the self and the other (or object) in relation to the self depend on the same body-centered reference frame. The first component is subject to both proprioception (yellow dotted lines) and exteroception (red dotted lines), while the other two components are inferred via exteroception only. In this case, the interaction with the second agent just depends on the observation of its last level, leading, e.g., to a hand-shaking action.}
		\label{fig:others_body}
	\end{figure}
	
	Besides modeling object affordances, this strategy is also useful in multi-agent contexts. One could maintain a hierarchical generative model regarding the kinematic chain of another agent, which would be inferred by exteroceptive observations about all its positions and joint angles, starting from a different body-centered reference frame. As shown in Figure \ref{fig:others_body}, the goal-directed method used for external objects reflects in this case as well: the agent could represent, by a parallel hierarchical pathway, a second agent in relation to itself, expressing a particular kind of interaction (e.g., the hand of the second agent in terms of its own, resulting in a shaking action). These two cases could be interpreted, from a biological perspective, as simulating the functioning of mirror neurons, firing whenever a subject executes a voluntary goal-directed action or when that action is performed by other subjects \cite{Rizzolatti2004}. Building an internal model with the kinematic chain of the others -- both per se and in relation to the self -- could be critical to predict (thus, to understand) their intentions. In this view, neural activity results because the agent makes constant predictions over their kinematic structures depending on its hypotheses and the current context \cite{Kilner2007,Friston2011b}.
	
\begin{figure}[h]
		\begin{minipage}[b]{\textwidth}
			\centering
			\subfloat
			[]
			{\label{fig:example_objects}\vspace{0em}\includegraphics[height=0.3\textheight]{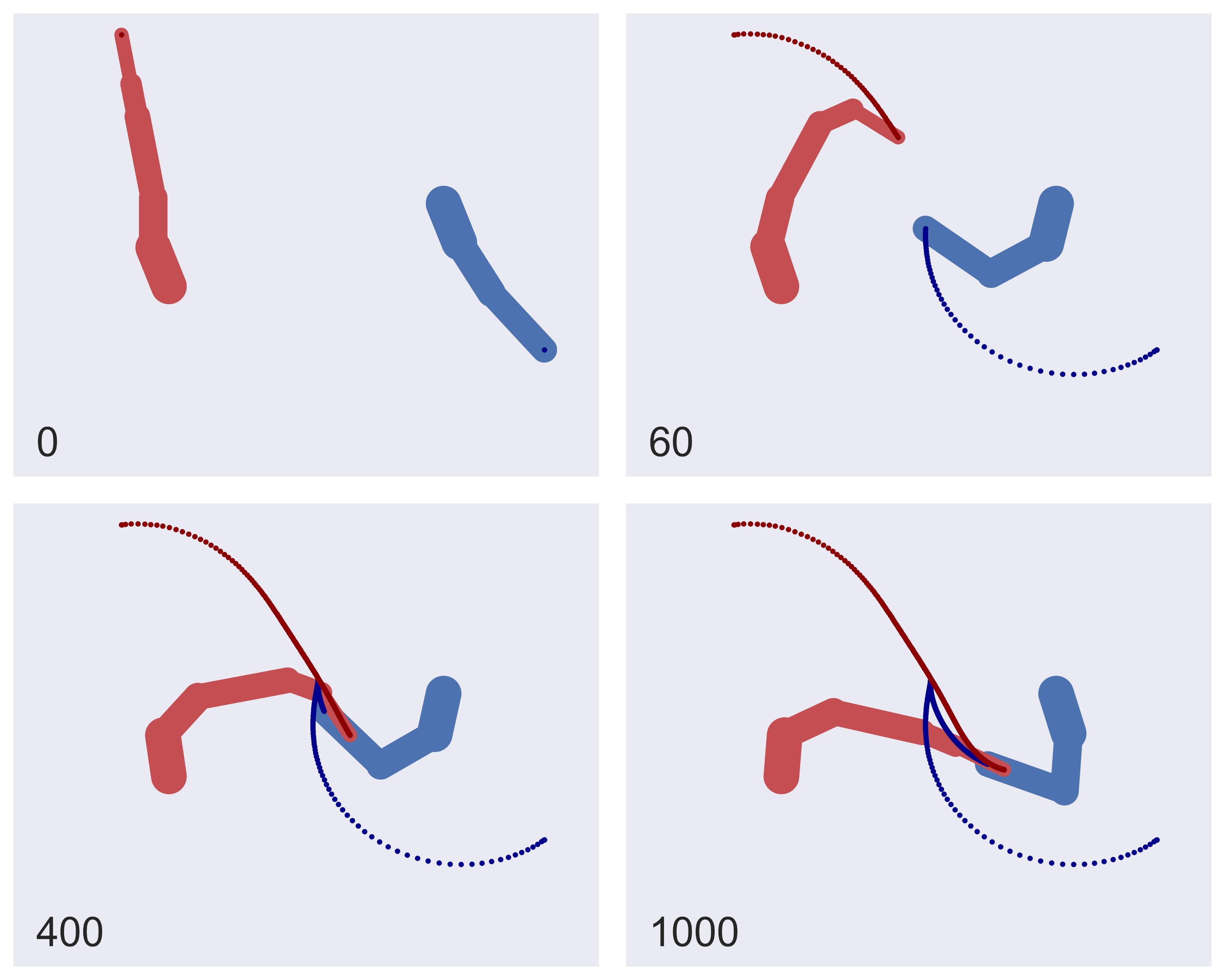}}
			\hspace{3em}
			\subfloat
			[]
			{\label{fig:example_relation}\vspace{0em}\includegraphics[height=0.3\textheight]{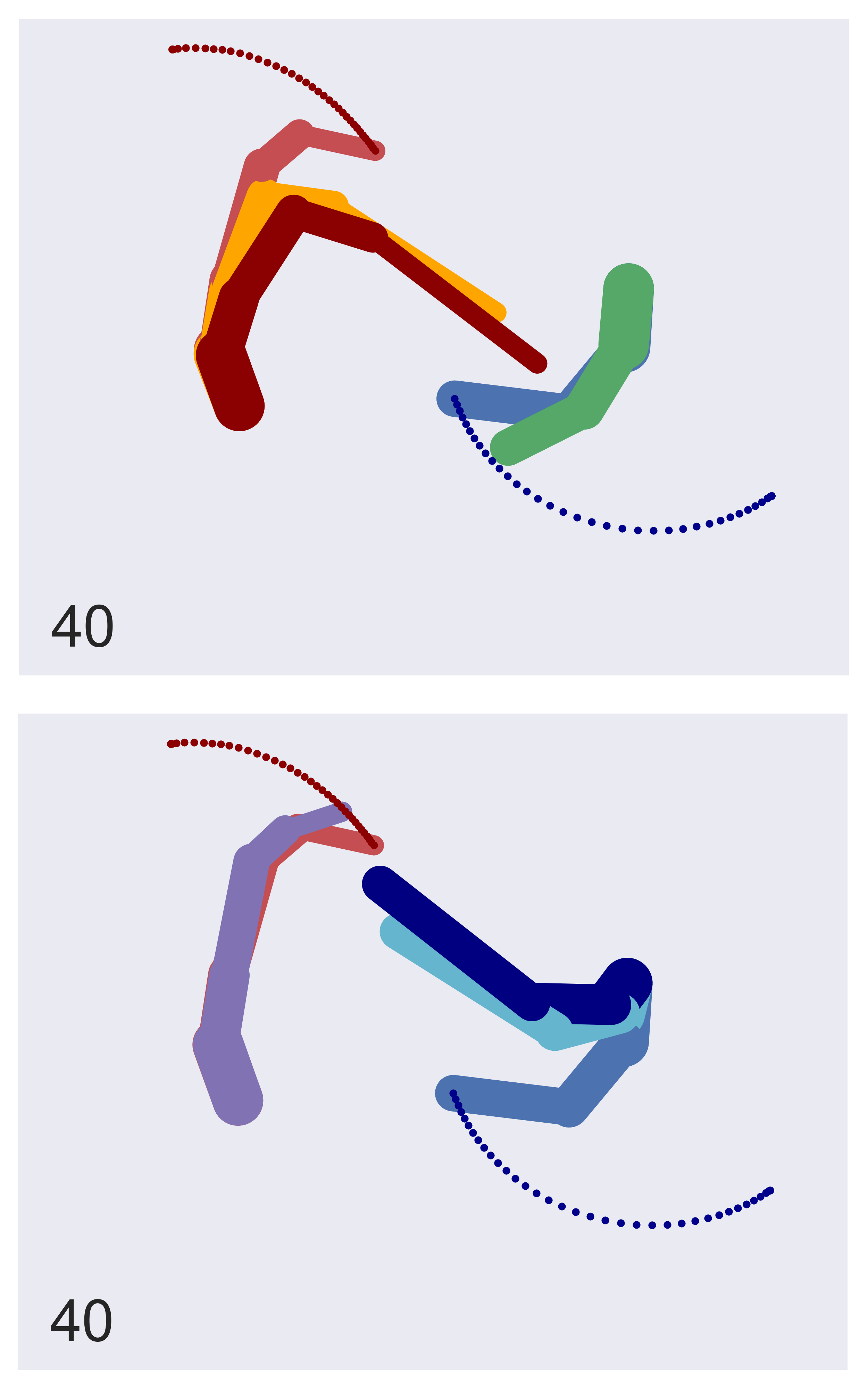}}
		\end{minipage}
		\caption{\textbf{(a)} In this task, two agents with different kinematic chains (respectively of 5 and 3 DoF) have two incompatible goals: the first agent (in red) has to reach the elbow of the second agent (in blue), while the second agent has to reach the hand of the first agent. Note that after an initial approaching phase from both, the second agent gradually retracts trying to touch the hand of the first agent. \textbf{(b)} Beliefs of both agents, compared with their actual configurations (displayed in light red for the first agent, and in light blue for the second). The top and bottom graphs respectively show the beliefs of the first and second agents. Specifically, the belief of the actual configuration of the self (in orange or cyan), the belief of the other in relation to the self (in dark red or dark blue), and the belief of the other (in green or purple). Note the belief succession during goal-directed behavior: the potential configuration pointing toward one's intentional state (either the elbow or the hand), followed by the current belief, and then by the real arm. Also note a slight delay about the inference of the configuration of the other agent.}
		\label{fig:frames_objects}
	\end{figure}

	The relationships between the self, the objects, and other agents under active inference may be better understood from the simulation of Figure \ref{fig:frames_objects}, showing two agents with incompatible goals that depend on each other. Here, both agents are able to infer parallel representations of different kinematic chains, using an effective decomposition of potential and real configurations. Note how one's current belief is always in between the intentional state to be realized and the actual configuration; this speaks of one of the fundamental aspects of active inference, i.e., that our beliefs never really reflect the state of the affairs of the world, but are always biased toward preferred states -- eventually driving action. In general, bodily states, objects, or other agents can all be manipulated in reference frames appropriate for a specific context; this is in line with the hypothesis that cortical columns use object-centered reference frames to encode external elements and more abstract entities \cite{Hawkins2017ATO}.
	
	\section{The hybrid unit}
	
	The continuous-time hierarchical models presented so far lack effective usability in the real world: although they can represent any future trajectory - which implies a degree of planning - they do not possess an explicit model of future states nor can they choose among alternative trajectories. In this chapter, we turn to the problem of how to integrate discrete decision-making into continuous motor control. In doing this, we revisit the basic unit of the first chapter, finally using the second input -- the prior over the hidden causes.
	
	Active inference in discrete state-space \cite{DaCosta2020,Smith2022} -- generally attributed to the cerebral cortex, especially prefrontal areas \cite{Parr2020}, along with corticostriatal loops -- exploits the structure of Partially Observable Markov Decision Processes (POMDPs) to plan abstract actions over expected sensations. This (active) inference relies on the minimization of the \textit{expected free energy}, i.e., the free energy that the agent expects to perceive in the future. The expected free energy can be unpacked into two terms resembling the two classical aspects of control theory, exploration and exploitation -- which here naturally arise; these respectively correspond to an uncertainty-reducing term, and a goal-seeking term that finds a sequence of actions toward the agent's prior belief.
	
	Further, so-called \textit{mixed} or \textit{hybrid} models \cite{Friston2017,Friston2017a} combine the potentialities of a discrete model with the inference of continuous signals, allowing robust decision-making in changing environments. While the theory of \textit{Bayesian model reduction} \cite{Friston2003e,Friston2011c,Friston2018a,Rosa2012} provides efficient communication between the two models, this unified approach has not enjoyed many practical implementations for the time being \cite{Parr2021,Friston2017,Friston2017a,Parr2018b,pf19,tbmbsp21,Parr2018c}. An open issue regards how to deal with highly dynamic environments: hybrid models usually perform a comparison between static priors, limiting the agent to realize, e.g., multi-step reaching movements through fixed positions. In \cite{Priorelli2023e}, a hybrid model in which the agent's hypotheses were generated at each time step from the system dynamics allowed to relate continuous trajectories with discrete plans. Besides these more conventional solutions, many other hybrid methods have been proposed recently. One study addressed the problem of realistic robot navigation in active inference, making use of bio-inspired SLAM methods \cite{ccatal2021robot}. Other studies proposed how to integrate active inference with imitation learning for autonomous vehicles, using a Dynamic Bayesian Network (DBN) to explain the interactions between an expert agent and dynamic objects \cite{Nozari2022, Nozari2023}. An enhanced DBN consisting of two continuous levels and two discrete levels was used to model the behavior of a UAV at different timescales for assisting wireless communications \cite{Krayani2023, Krayani2024,Obite2023, Krayani2022}. A study in robotics combined active inference with behavior trees for reactive action planning in dynamic environments \cite{Pezzato2023}. Finally, a hybrid model based on recurrent switching linear dynamical systems allowed to discover non-grid discretizations of a continuous Mountain Car task. \cite{collis2024learninghybridactiveinference}. 
	
	
	\subsection{\label{sec:dyn_inf}Dynamic inference by model reduction}
	
	Recall that in the unit of Figure \ref{fig:affordances}, some sort of multi-step behavior was achieved, based on higher-level priors over tactile sensations. In most cases, we need to switch intentions based on lower-level information, affording a more dynamic behavior. Considering a pick-and-place operation, an IE module would be more confident about the success of the first reaching movement if it could rely not just on a tactile belief but also on its kinematic configuration. In other words, hidden causes $\bm{v}$ should manage to effectively use both its prior $\bm{\eta}_v$ and the dynamics prediction error $\bm{\varepsilon}_x$. The latter assumes two different roles depending on which pathway it flows into: the gradient with respect to the hidden states infers the position that is most likely to have generated the current trajectory, needed for movement; conversely, the gradient with respect to the hidden causes infers the most likely combination of gains $v_m$, signaling the current status of the trajectory and resulting in a dynamic modulation of intentions. However, since the hidden causes are generated by Gaussian distributions, they do not encode proper probabilities, and the gradient $\partial_{v} \bm{f}_x$ infers just one over many possible combinations of gains. Instead, what we need is to infer \textit{the most likely intention} to have generated the current trajectory. Thus, to implement a correct intention selection, we assume that the hidden causes are generated from a categorical distribution:
	\begin{equation}
		\label{eq:pref_v}
		p(\bm{v}) = Cat(\bm{H}_v)
	\end{equation}
	where $\bm{H}_v$ is a prior preference like $\bm{\eta}_v$. In this way, each discrete element of $\bm{v}$ represents the probability that a specific continuous trajectory will be realized.
	
\begin{figure}[h]
		\begin{minipage}[b]{0.44\textwidth}
			\centering
			\subfloat
			[]
			{\label{fig:hybrid_conv}\vspace{0em}\includegraphics[width=\textwidth]{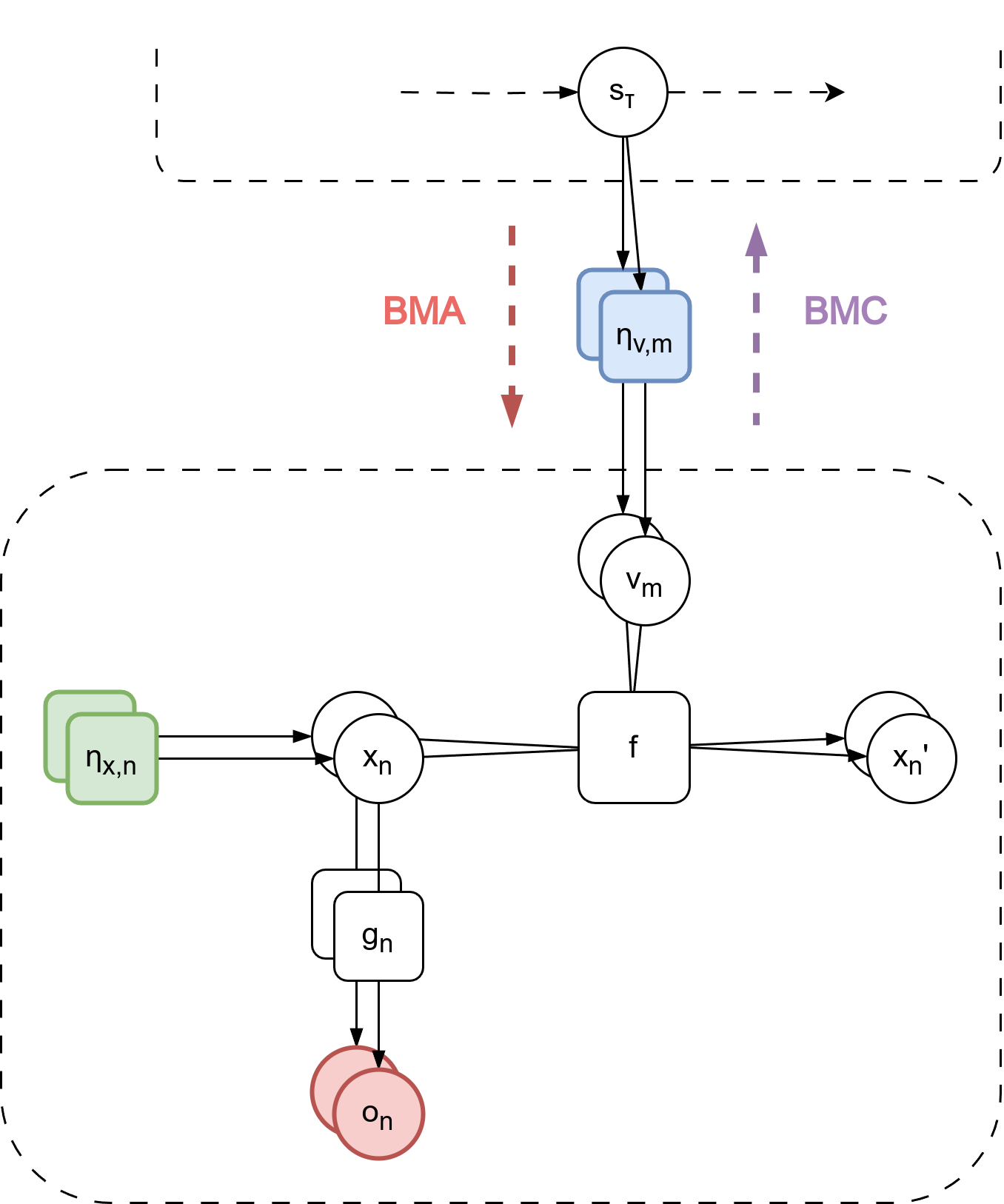}}
			\vspace{1em}
			\subfloat
			[]
			{\label{fig:bmc}\vspace{0em}\includegraphics[width=0.75\textwidth]{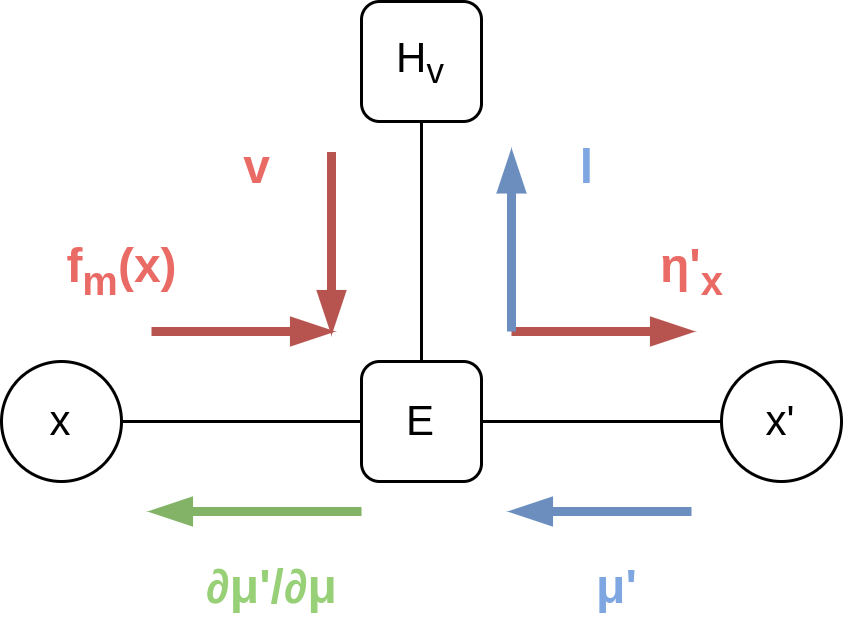}}
		\end{minipage}
		\hfill
		\begin{minipage}[b]{0.45\textwidth}
			\centering
			\subfloat
			[]
			{\label{fig:hybrid}\vspace{0em}\includegraphics[width=\textwidth]{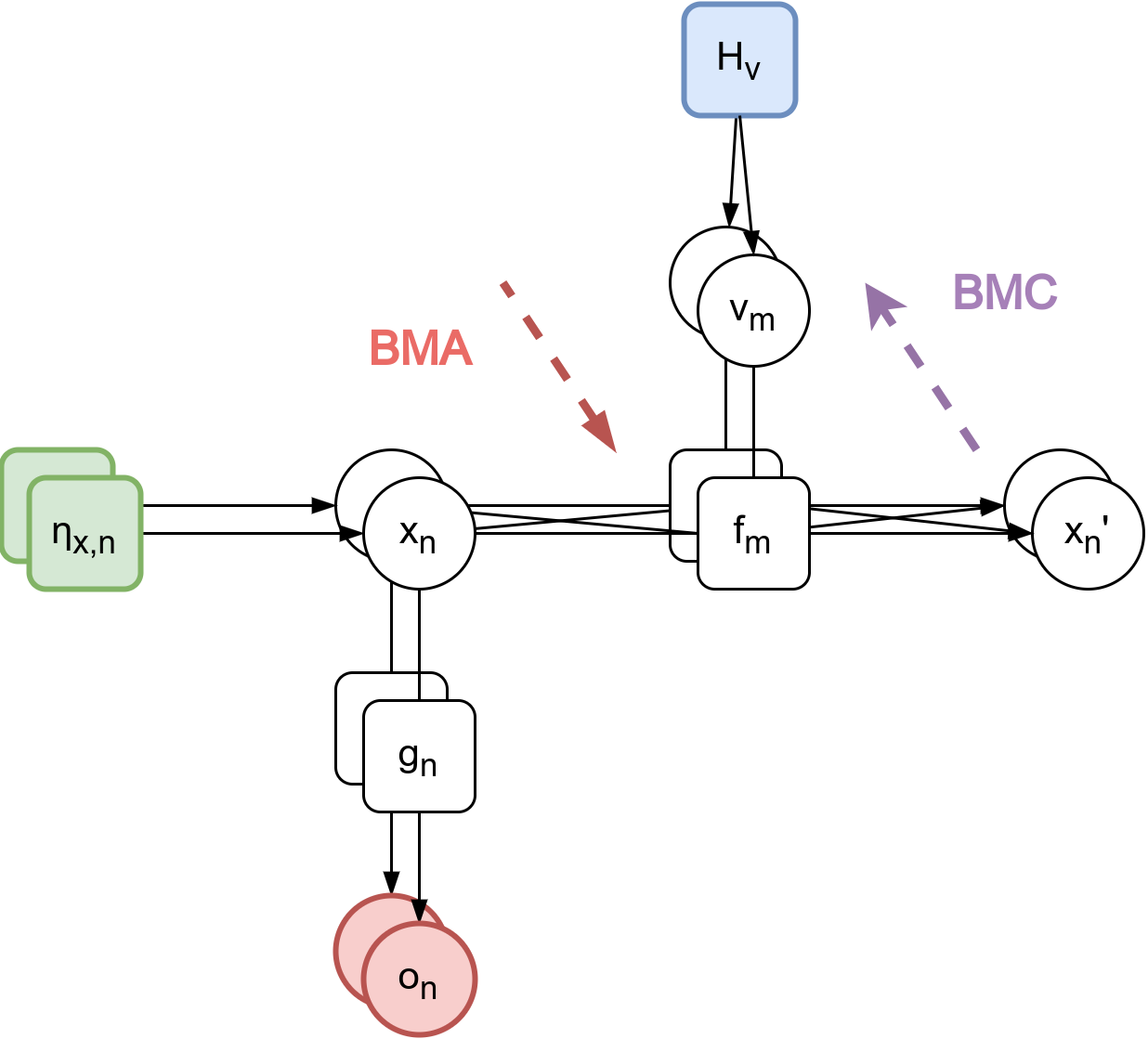}}
			\vspace{1em}
			\subfloat
			[]
			{\label{fig:gen_model_dyn}\vspace{0em}\includegraphics[width=0.9\textwidth]{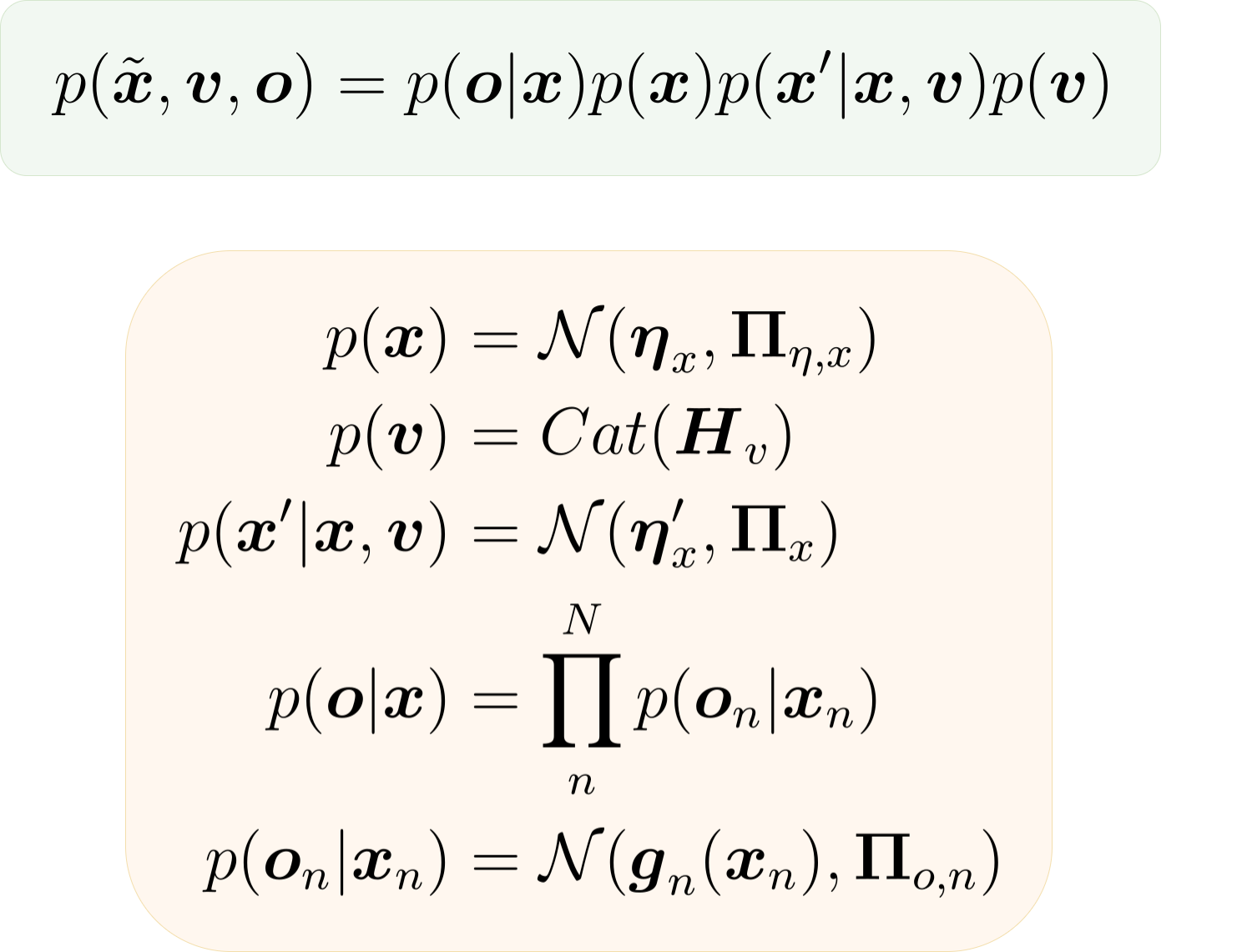}}
		\end{minipage}
		\caption{\textbf{(a)} A conventional hybrid model, composed of a discrete model at the top and a continuous model at the bottom. For simplicity, we assume that the continuous prior $\bm{\eta}_v$ is directly conditioned on discrete hidden states $\bm{s}$. We will cover discrete models in the following section. Here, top-down and bottom-up messages are computed by Bayesian model reduction of some static priors, without the possibility for dynamic planning. \textbf{(b)} A simplified graph displaying the exchange of top-down (red) and bottom-up (blue and green) messages of the hybrid control in \textbf{(c)}. \textbf{(c)} Factor graph of the unit with hybrid control. The hidden causes are now generated from a categorical distribution, such that instead of inferring a combination of continuous attractor gains, the model correctly infers the most likely intention associated with the current trajectory. This is done by computing the free energy $E_m$ corresponding to each intention. Through Bayesian model reduction between discrete hidden causes and continuous hidden states, the agent updates its reduced priors at each time step. Generative model of (c).}
	\end{figure}

	Conversion between discrete hidden causes to continuous hidden states (and vice versa) is done via Bayesian model reduction, a technique used to constrain the complexity of full posterior models into simpler and more restrictive (formally called \textit{reduced}) distributions \cite{Friston2011c,Friston2018a}. Reduced means that the likelihood of some data is equal to that of the full model and the only difference rests upon the specification of the priors; hence, the posterior of a reduced model $m$ can be expressed in terms of the posterior of the full model:
	\begin{equation}
		p(\tilde{\bm{x}}|\bm{o},m) = p(\tilde{\bm{x}}|\bm{o}) \frac{p(\tilde{\bm{x}}|m) p(\bm{o})}{p(\tilde{\bm{x}}) p(\bm{o}|m)}
	\end{equation}
	In our case, model reduction means to explain the infinite values of a continuous signal with a discrete set of hypotheses. A simplified version of a hybrid active inference model is shown in Figure \ref{fig:hybrid_conv}. We can cast this procedure into the usual message passing, where top-down and bottom-up messages between the two domains perform a Bayesian Model Average (BMA) of reduced priors and a Bayesian Model Comparison (BMC) of reduced sensory evidence, respectively. In conventional hybrid models, discrete hidden states generate priors for the continuous hidden causes by weighting a specific reduced prior with the probability of each discrete state; hence, the reduced priors represent alternative hypotheses over the true causes of the sensorium \cite{Friston2017}. The posterior over the hidden causes is then compared with such reduced priors to find which one among them could be the best explanation of the environment, taking into account their discrete probabilities before observing sensory evidence.
	
	Because the agent compares continuous alternatives that are fixed and determined a-priori, it cannot correctly operate in a changing environment. For instance, if the agent thinks to find an object in one of two locations, it will always reach either one or the other initial guesses, even if the object has been moved to a third location. How then to use the newly available evidence to update our reduced assumptions? By considering the hidden causes as generated from a categorical distribution -- as in Equation \ref{eq:pref_v} -- we can compare the posterior over the hidden states with the output of the dynamics functions $\bm{f}_m$, which thus act as the agent's reduced priors \cite{Priorelli2023e}. More formally, we define $M$ reduced prior probability distributions and a full prior model:
	\begin{align}
		\begin{split}
			\label{eq:precisions}
			p(\bm{x}' | \bm{x}, m) &= \mathcal{N}(\bm{f}_m(\bm{x}), \bm{\Pi}_{x,m}) \\
			p(\bm{x}' | \bm{x}, \bm{v}) &= \mathcal{N}(\bm{\eta}_{x}^\prime, \bm{\Pi}_x)
		\end{split}
	\end{align}
	where $\bm{\eta}_{x}^\prime$ is the full prior. Note here that the reduced priors have the same form of Equation \ref{eq:intention_dyn} but are not directly conditioned on the hidden causes:
	\begin{equation}
		\label{eq:red_dyn}
		\bm{f}_m(\bm{x}) = \bm{e}_{i,m} =  \bm{i}_m(\bm{x}) - \bm{x}
	\end{equation}
	Next, we define the corresponding posterior models:
	\begin{align}
		\begin{split}
			q(\bm{x}^{\prime}|m) &= \mathcal{N}(\bm{\mu}_{x,m}^{\prime}, \bm{P}_{x,m}) \\
			q(\bm{x}^{\prime}) &= \mathcal{N}(\bm{\mu}_x^{\prime}, \bm{P}_x)
		\end{split}
	\end{align}
	Now, we can find the full prior and its prediction error by averaging the continuous trajectories with their respective discrete probabilities:
	\begin{align}
		\begin{split}
			\label{eq:traj_prior}
			\bm{\eta}_{x}^\prime &= \sum_m^M v_m\bm{f}_m(\bm{x}) = \sum_m^M v_m \bm{e}_{i,m} \\
			\bm{\varepsilon}_x &= \bm{\mu}_x^\prime - \bm{\eta}_{x}^\prime
		\end{split}
	\end{align}
	which have the same form of Equations \ref{eq:intention_dyn} and \ref{eq:eps_x}. In fact, the hidden states still perceive a single dynamics prediction error containing the total contribution of every intention. Concerning the bottom-up messages $\bm{l}$, we first write the free energy of each reduced model in terms of the full model. As before, maximizing each reduced free energy makes it approximate the log evidence:
	\begin{equation}
		\mathcal{F}(m) = \mathcal{F} - \ln \int \frac{p(\tilde{\bm{x}}|m)}{p(\tilde{\bm{x}})} q(\tilde{\bm{x}}) d \tilde{\bm{x}} \approx \ln p(\bm{o}|m)
	\end{equation}
	As a result, the free energy related to each dynamics function $\bm{f}_m$ depends on the approximate posterior $q(\tilde{\bm{x}})$ of the full model, avoiding the computation of the reduced posteriors. Under a Gaussian approximation, the $m$th reduced free energy breaks down to a simple formula and the bottom-up messages $\bm{l}$ are found by accumulating the log evidence associated with every intention for a certain amount of continuous time $T$:
	\begin{align}
		\begin{split}
			\label{eq:log_ev}
			l_m &= \int_0^T \mathcal{L}_m dt \\
			\mathcal{L}_m &= \frac{1}{2} (\bm{\mu}_{x,m}^{\prime T} \bm{P}_{x,m} \bm{\mu}_{x,m}^{\prime} - \bm{f}_m(\bm{x})^T \bm{\Pi}_{x,m} \bm{f}_m(\bm{x}) - \bm{\mu}_x^{\prime T} \bm{P}_x \bm{\mu}_x^{\prime} + \bm{\eta}_{x}^{\prime T} \bm{\Pi}_x \bm{\eta}_{x}^\prime)
		\end{split}
	\end{align}
	Then, a BMC turns into computing the softmax of a vector comprising the free energy $E_m$ of every reduced model. This quantity compares the prior surprise $-\ln \bm{H}_v$ with the accumulated log evidence:
	\begin{equation}
		\label{eq:disc_v}
		\bm{v} = \sigma(-\bm{E}) = \sigma(\ln \bm{H}_v + \bm{l})
	\end{equation}
	where $\bm{l} = [l_1, \dots, l_M]$. See \cite{Friston2011c,Friston2018a} for a full derivation of BMC under the Laplace assumption, and \cite{Priorelli2023e} for more details about the presented approach. Equation \ref{eq:disc_v} is the discrete analogous of Equation \ref{eq:v_update}, but now the bottom-up message encodes a proper discrete distribution and can be used to infer the most likely intention associated with the current trajectory.
	
	The factor graph of this model, which we call a \textit{hybrid unit}, is displayed in Figure \ref{fig:hybrid}. Its inference process at each continuous step is better understood if we analyze separately the three different pathways shown in Figure \ref{fig:bmc}: (i) during the forward pass, the unit receives a discrete intention prior $\bm{H}_v$, performs a BMA with potential trajectories $\bm{f}_m(\bm{x})$, and imposes a prior $\bm{\eta}_{x}^\prime$ over the 1st order; (ii) through the first backward pass, the unit accumulates the most likely intention related to the current trajectory by comparing it to the ones generated by the dynamics functions; (iii) in the second backward pass, the unit propagates the dynamics prediction error back to the 0th order to infer the most likely continuous state associated with the trajectory, eventually generating biased observations. After a period $T$, the unit finally computes the difference between the discrete prior and the accumulated evidence, generates a new combination of intentions, and the process starts over.
	
	\begin{figure}
		\begin{subfigure}{0.42\textwidth}
			\centering
			\includegraphics[width=\textwidth]{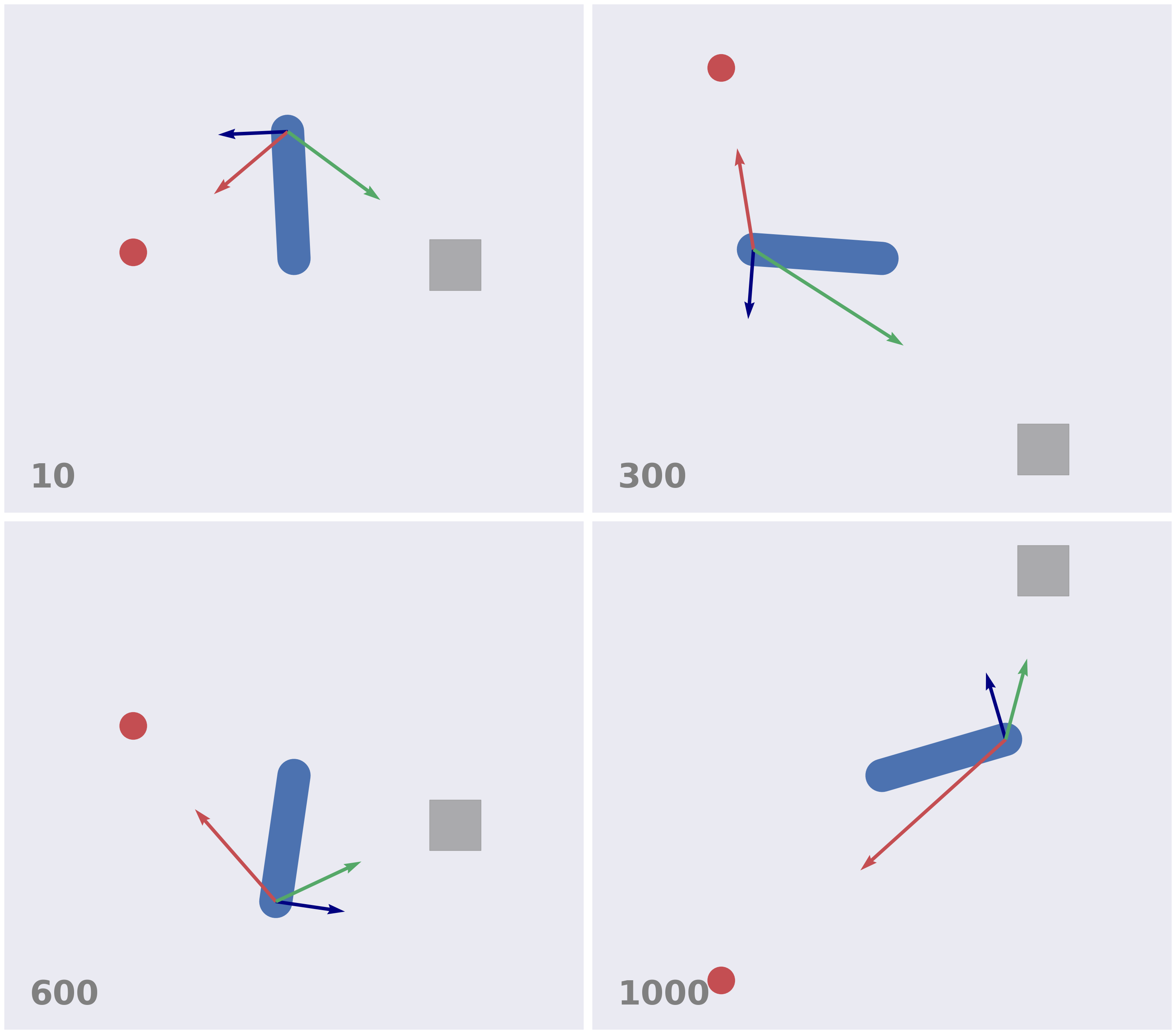}
			\caption{}
			\label{fig:frames_dyn_inf}
		\end{subfigure}
		\hfill
		\begin{subfigure}{0.55\textwidth}
			\centering
			\includegraphics[width=\textwidth]{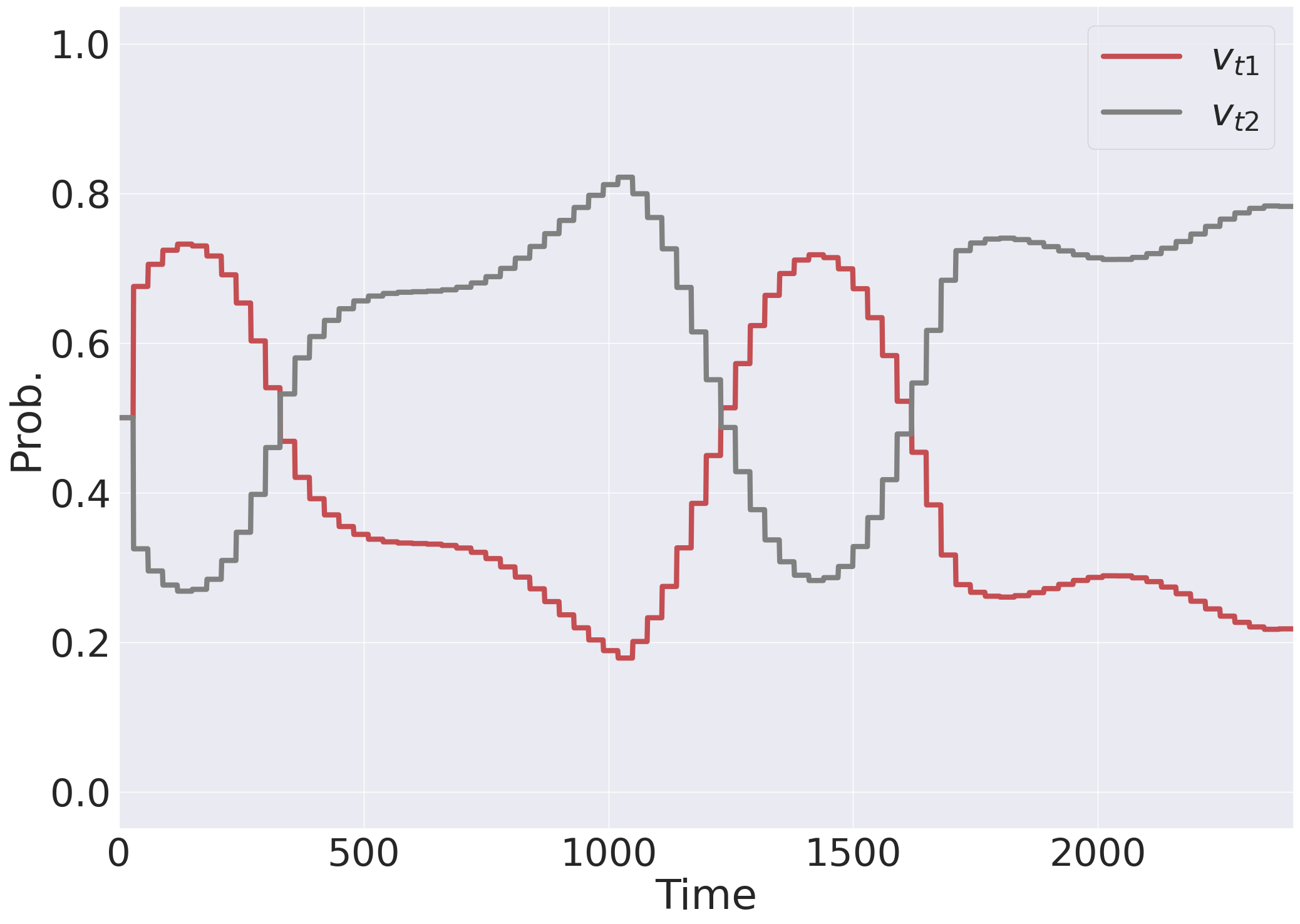}
			\caption{}
			\label{fig:dynamics_dyn_inf}
		\end{subfigure}
		\caption{\textbf{(a)} In this perception task, the agent has to infer which one among two objects (a red ball or a grey square) is following. The arm moves along a circular trajectory, while the two objects move linearly in opposite directions. \textbf{(a)} Sequence of time frames (left). The dark blue arrow represents the real agent's trajectory, while the red and green arrows represent potential trajectories associated with different reaching movements. \textbf{(b)} Dynamics of hidden causes. The two hidden causes $v_{t1}$ and $v_{t2}$ are associated with the two potential trajectories. As the hand moves away from the first target and approaches the second target, the dynamic accumulation of evidence leads to an increase in the second hidden cause. Here, we used a uniform prior $\bm{H}_v$, so the hidden causes only express the accumulated log evidence $\bm{l}$. Also, the continuous time $T$ for evidence accumulation was set to $30$ time steps. See \cite{Priorelli2023e} for more details.}
		\label{fig:example_inference}
	\end{figure}
	
	Now how this mechanism is self-reinforcing: a decision generates some movement, which in turn infers the decision itself to compute the next movement. Inferring a target from body movements is a fundamental aspect often neglected in decision-making studies, which can result in completely different behaviors. In particular, this component produces a commitment effect to the decision taken and stabilizes the actions, avoiding changes of mind that may lead to the loss of valid opportunities in dynamic environments \cite{Lepora2015, Priorelli2024b}. Further, this kind of dynamic inference has several utilities, e.g., it can be used to infer which one among multiple objects an agent is following -- as exemplified in Figure \ref{fig:example_inference} -- by generating trajectories for different objects and comparing them with the one perceived \cite{Priorelli2023e}. 
	
	An alternative that produces similar results is the model used in \cite{Isomura2019} in the context of social exchange. In this hybrid solution, a "student" bird maintains several continuous generative models for every teacher conspecifics. Discrete switching variables placed at the center of these hypotheses infer which teacher bird has generated the perceived birdsong. Learning of the generative models relies on two complementary methods, i.e., Bayesian model average of all possible teacher birds, or Bayesian model selection of a specific bird generating a song.
	
	As a last note, the dynamics precisions of Equations \ref{eq:precisions} and \ref{eq:log_ev} have here an interesting interpretation equivalent to the observation precisions $\bm{\Pi}_o$. Predictive coding assumes that whenever an agent perceives high noise about a sensory modality, the precision of that generative model will decrease because it cannot be trusted for understanding the state of affairs of the world \cite{Hohwy2013,Clark2016}. In addition, the dualism between action and perception inherent to the free energy principle tells us that the optimization of precisions -- which are thought to be encoded as synaptic gains -- could play a crucial role in attention mechanisms that selectively sample sensory data \cite{Feldman2010,Parr2018p}. Based on these assumptions, we note a dual interpretation of (reduced) dynamics precisions. A low precision $\bm{\Pi}_{x,m}$ related to a grasping action could mean that it is unreliable to explain the current context (e.g., an object far away from the hand). In addition, it could mean that the agent does not intend to rely on it to achieve a goal (e.g., grasping an object when it is out of reach). This perspective unveils an additional mechanism besides the fast inference of hidden causes that we mentioned before: a slow learning of reduced precisions that lets the agent score -- and, crucially, focus on -- those intentions that would be appropriate for a specific scenario \cite{Priorelli2023e}.
	
	
	\subsection{\label{sec:dyn_plan}A discrete interface for dynamic planning}
	
	Numerous studies have demonstrated that the brains of athletes are marked by a higher activation of posterior and subcortical regions that involves little or no conscious thinking, producing fluid transitions between different motions; in contrast, the brain of a novice requires a higher demand of prefrontal computations that results in lower performances \cite{Fattapposta1996,DiRusso2005,Graybiel2008}. From an active inference perspective, we can compare the proficiency of athletes with the continuous model of Figure \ref{fig:affordances}, corresponding to the subcortical sensorimotor loops. This model encodes a transition mechanism that is not very flexible, but that precisely for this can react much more rapidly to environmental stimuli, e.g., when grasping objects moving at high speed \cite{Priorelli2023d}. In general, this strategy can be very effective when the environment has limited uncertainty and the task to be solved comprises a rigid sequence of actions, which the agent has already correctly learned. However, suppose that the agent is introduced to a novel or complex task that requires careful thinking about the imminent future. In this case, it should be capable of replanning the correct sequence of actions if something goes unexpectedly, and a high-level belief always producing an a-priori-determined behavior for the hidden causes would fail to complete the task.
	
	\begin{figure}[h]
		\centering
		\includegraphics[width=0.95\textwidth]{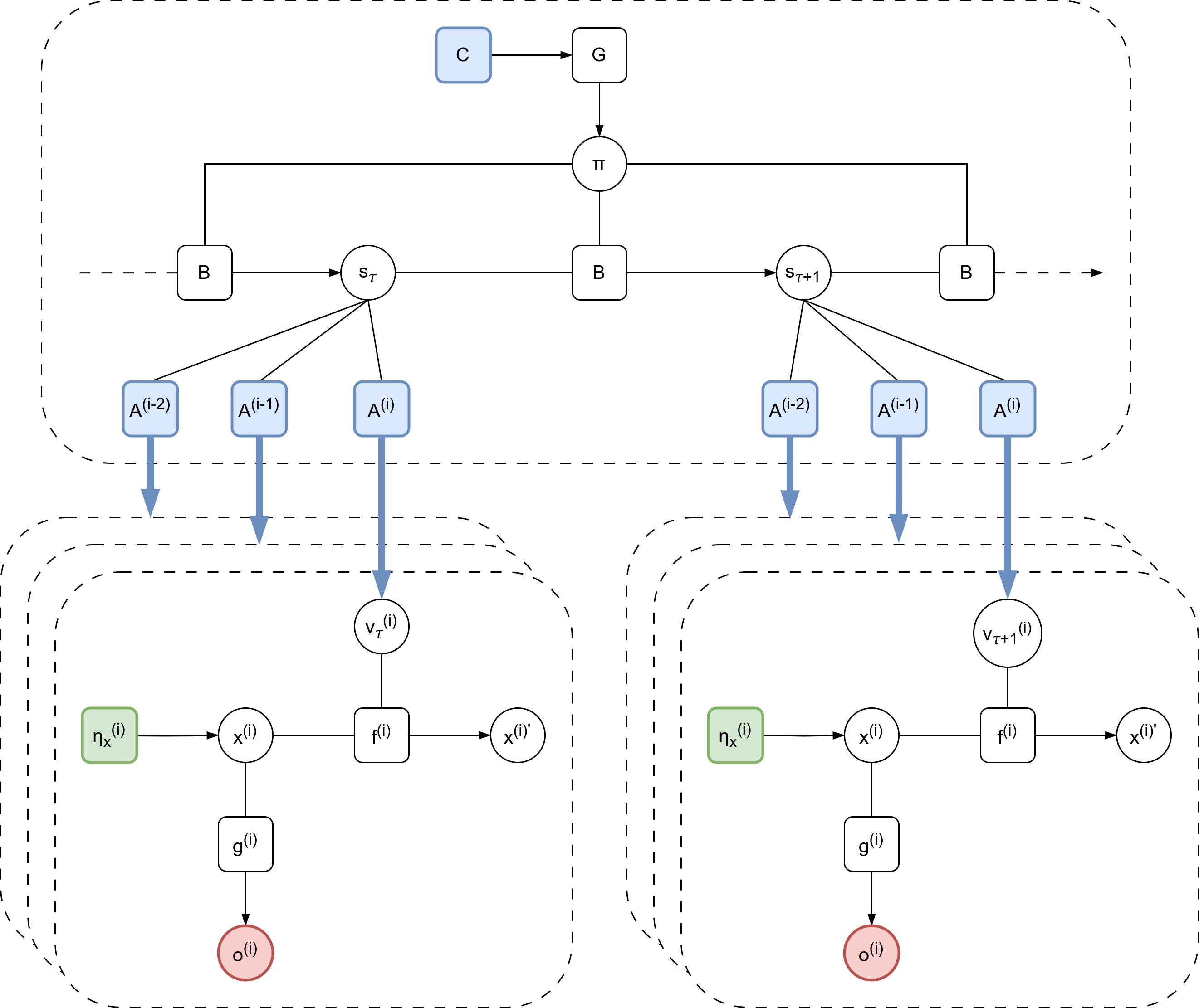}
		\caption{Interface between a discrete model (at the top) and several hybrid units (at the bottom). For clarity, the hidden states factorization of each unit is not displayed. The discrete hidden states $\bm{s}$ at time $\tau+1$ are computed from the current hidden states $\bm{s}_{\tau}$, by choosing some policy $\bm{\pi}_\pi$ related to a specific transition distribution encoded in $\bm{B}$. The best policy at any moment is the sequence of actions that most minimizes the expected free energy $\mathcal{G}$. As a result, the agent will try to sample those observations that conform to its preference $\bm{C}$. The discrete hidden causes $\bm{v}^{(i)}$ at time $\tau$ are directly generated from discrete hidden states $\bm{s}_{\tau}$ through likelihood matrices $\bm{A}^{(i)}$, thus affording dynamic planning, synchronized behavior, and inference with multiple evidences.}
		\label{fig:interface}
	\end{figure}
	
	Having replaced the continuous hidden causes of Figure \ref{fig:affordances} with discrete hidden causes in Figure \ref{fig:hybrid}, we can now endow the agent with planning capabilities through a discrete model composed of the following distributions -- as shown in Figure \ref{fig:interface}:
	\begin{equation}
		p(\bm{s}_{1:\mathcal{T}},\bm{v}_{1:\mathcal{T}},\bm{\pi}) = p(\bm{s}_1) p(\bm{\pi}) \prod_{\tau}^{\mathcal{T}} p(\bm{v}_\tau|\bm{s}_\tau) p(\bm{s}_\tau|\bm{s}_{\tau-1},\bm{\pi})
	\end{equation}
	where:
	\begin{align}
		\label{eq:disc_gen}
		\begin{aligned}
			p(\bm{s}_1) &= Cat(\bm{D}) \\
			p(\bm{\pi}) &= \sigma(-\mathcal{G})
		\end{aligned}
		&&
		\begin{aligned}
			p(\bm{v}_\tau|\bm{s}_\tau) &= Cat(\bm{v}_\tau | \bm{A} \bm{s}_\tau) \\
			p(\bm{s}_\tau|\bm{s}_{\tau-1},\bm{\pi}) &= Cat(\bm{s}_\tau | \bm{B}_{\pi,\tau} \bm{s}_{\tau-1})
		\end{aligned}
	\end{align}
	Here, $\bm{A}$, $\bm{B}$, $\bm{D}$ are the likelihood matrix, transition matrix, and prior of the hidden states, $\bm{\pi}$ are the policies (which are not state-action mappings as in RL but sequences of actions), $\bm{s}_\tau$ are the discrete hidden states at time $\tau$. These quantities have strict analogies with their continuous counterparts of Figure \ref{fig:gen_model_reaching}, i.e., the likelihood function $\bm{g}$, the dynamics function $\bm{f}$, the prior $\bm{\eta}_x$, hidden causes $\bm{v}$, and hidden states $\tilde{\bm{x}}$, with the difference that the discrete hidden states do not encode instantaneous paths expressed in generalized coordinates, but sequences of future states. Crucially, $\mathcal{G}$ is the \textit{expected free energy}, defined as the free energy that the agent expects to perceive in the future.
	
	Note here that the likelihood matrix $\bm{A}$ expresses a conditional probability over the discrete hidden causes $\bm{v}_\tau$. As in conventional hybrid models, the discrete hidden states are linked to the hidden causes, but now the latter directly act as discrete outcomes generated by the likelihood matrix, which thus replaces the prior $\bm{H}_v$ in Equation \ref{eq:pref_v}. Here, we wish to infer the posterior distribution:
	\begin{equation}
		p(\bm{s}_{1:\mathcal{T}}, \bm{\pi} | \bm{v}_{1:\mathcal{T}}) = \frac{p(\bm{v}_{1:\mathcal{T}}|\bm{s}_{1:\mathcal{T}}, \bm{\pi}) p(\bm{s}_{1:\mathcal{T}}, \bm{\pi})}{p(\bm{v}_{1:\mathcal{T}})}
	\end{equation}
	As before, this requires computing the intractable model evidence $p(\bm{v}_{1:\mathcal{T}})$, so we resort to a variational approach: we first express the approximate posterior by its sufficient statistics $\bm{s}_{\pi,\tau}$ and conditioning upon a specific policy:
	\begin{equation}
		p(\bm{s}_{1:\mathcal{T}}|\bm{v}_{1:\mathcal{T}},\bm{\pi}) \approx q(\bm{s}_{1:\mathcal{T}},\bm{\pi}) = q(\bm{\pi}) \prod_\tau^{\mathcal{T}} q(\bm{s}_\tau|\bm{\pi})
	\end{equation}
	where:
	\begin{align}
		\begin{split}
			q(\bm{\pi}) &= Cat(\bm{\pi}) \\
			q(\bm{s}_\tau|\bm{\pi}) &= Cat(\bm{s}_{\pi,\tau})
		\end{split}
	\end{align}
	We then compute the variational free energy $\mathcal{F}_\pi$ of that policy:
	\begin{equation}
		\label{eq:disc_fe}
		\mathcal{F}_\pi = \sum_{\tau=1}^{\mathcal{T}} \bm{s}_{\pi,\tau} \ln \bm{s}_{\pi,\tau} - \sum_{\tau=1}^{\mathcal{T}} \bm{v}_{\tau} \ln \bm{A} \bm{s}_{\pi,\tau} - \bm{s}_{\pi,1} \ln \bm{D} - \sum_{\tau=2}^{\mathcal{T}} \bm{s}_{\pi,\tau} \ln \bm{B}_{\pi,\tau} \bm{s}_{\pi,\tau-1}
	\end{equation}
	Differentiating Equation \ref{eq:disc_fe} allows us to infer the most likely discrete hidden states at time $\tau$ for policy $\pi$:
	\begin{equation}
		\bm{s}_{\pi,\tau} = \sigma( \ln \bm{B}_{\pi,\tau-1} \bm{s}_{\pi,\tau-1} + \ln \bm{B}_{\pi,\tau}^T \bm{s}_{\pi,\tau+1} + \ln \bm{A}^T \bm{v}_{\tau} )
	\end{equation}
	where we applied a softmax function to ensure that it is a proper probability distribution. In order to infer the policies, we additionally consider unobserved outcomes as random variables, and then condition the joint probability of the generative model upon some preferences $\bm{C}$. In this way, the agent can predict future outcomes and select the most likely sequence of actions that will lead it to its desired state. The expected free energy $\mathcal{G}_\pi$ under policy $\pi$ consists of a \textit{pragmatic} or goal-seeking term toward the agent's preferences, and an \textit{epistemic} or uncertainty-reducing term (see \cite{DaCosta2020} for more details):
	\begin{align}
		\begin{split}
			\label{eq:exp}
			\mathcal{G}_{\pi} &\approx \sum_\tau^{\mathcal{T}} D_{KL} [q(\bm{v}_\tau|\bm{\pi}) || p(\bm{v}_\tau|\bm{C})] + \E_{q(\bm{s}_\tau|\bm{s}_{\tau-1},\bm{\pi})} [H[p(\bm{v}_\tau|\bm{s}_\tau)]] \\
			&= \sum_\tau^{\mathcal{T}} \bm{v}_{\pi,\tau} (\ln \bm{v}_{\pi,\tau} - \bm{C}_\tau) + \bm{s}_{\pi,\tau} \bm{H}_A
		\end{split}
	\end{align}
	where:
	\begin{align}
		\bm{v}_{\pi,\tau} = \bm{A} \bm{s}_{\pi,\tau} &&
		\bm{C}_\tau = \ln p(\bm{v}_\tau|\bm{C}) &&
		\bm{H}_A = -diag(\bm{A}^T \ln \bm{A})
	\end{align}
	Here, $\bm{v}_{\pi,\tau}$ is the prediction of the hidden causes at time $\tau$ and under policy $\pi$, $\bm{C}_\tau$ is the logarithm of the prior preference over the hidden causes, and $\bm{H}_A$ is the ambiguity of the likelihood matrix $\bm{A}$. The policy $\bm{\pi}$ is then found by applying a softmax function to the expected free energy, as defined in Equation \ref{eq:disc_gen}. 
	
	Using the (deep) hierarchical model outlined in the previous section, we may connect several hybrid units to the discrete model by appropriate likelihood matrices $\bm{A}^{(i)}$. Each of them has an independent interface whereby the discrete model computes different signals and waits for the next step $\tau+1$, when it can infer its hidden states based on multiple accumulated evidences. Recall that in the combined structure of Figure \ref{fig:interface}, the role of the hybrid unit was to predict a trajectory from a discrete intention prior, and to infer the most likely intention in a continuous period $T$. But now the intention prior is generated from the discrete model:
	\begin{equation}
		\bm{v}_\tau = \sigma(\ln \bm{A} \bm{s}_{\tau}  + \bm{l}_\tau)
	\end{equation}
	where $\bm{l}_\tau$ is the bottom-up message at time $\tau$ found by Equation \ref{eq:log_ev}, and we compute $\bm{s}_{\tau}$ by marginalizing over all policies, i.e. $\bm{s}_{\tau}=\sum_\pi \pi_\pi \bm{s}_{\pi,\tau}$. 
	
	In sum, computing the posterior probability over policies $\bm{\pi}$ turns into finding the best action that conforms to the dual objective defined by $\mathcal{G}$. Here, the discrete actions are not intended as actual motor commands similar to Equation \ref{eq:motor}, but as abstract actions over high-level representations. In fact, the hierarchical nature of discrete models in active inference permits performing decision-making with a separation of temporal scales, wherein a specific level can generate and infer the states and the paths of the level below \cite{FRISTON2024105500,Toon2024,deTinguy2024}. Further evaluating the consequences of an action for a longer time horizon affords more advanced planning called \textit{sophisticated inference} \cite{Friston2021a}. Computing actions with the expected free energy is different from the motor control of continuous models, which only minimize the free energy of present states.
	
	\begin{figure}[h]
		\centering
		\includegraphics[width=0.9\textwidth]{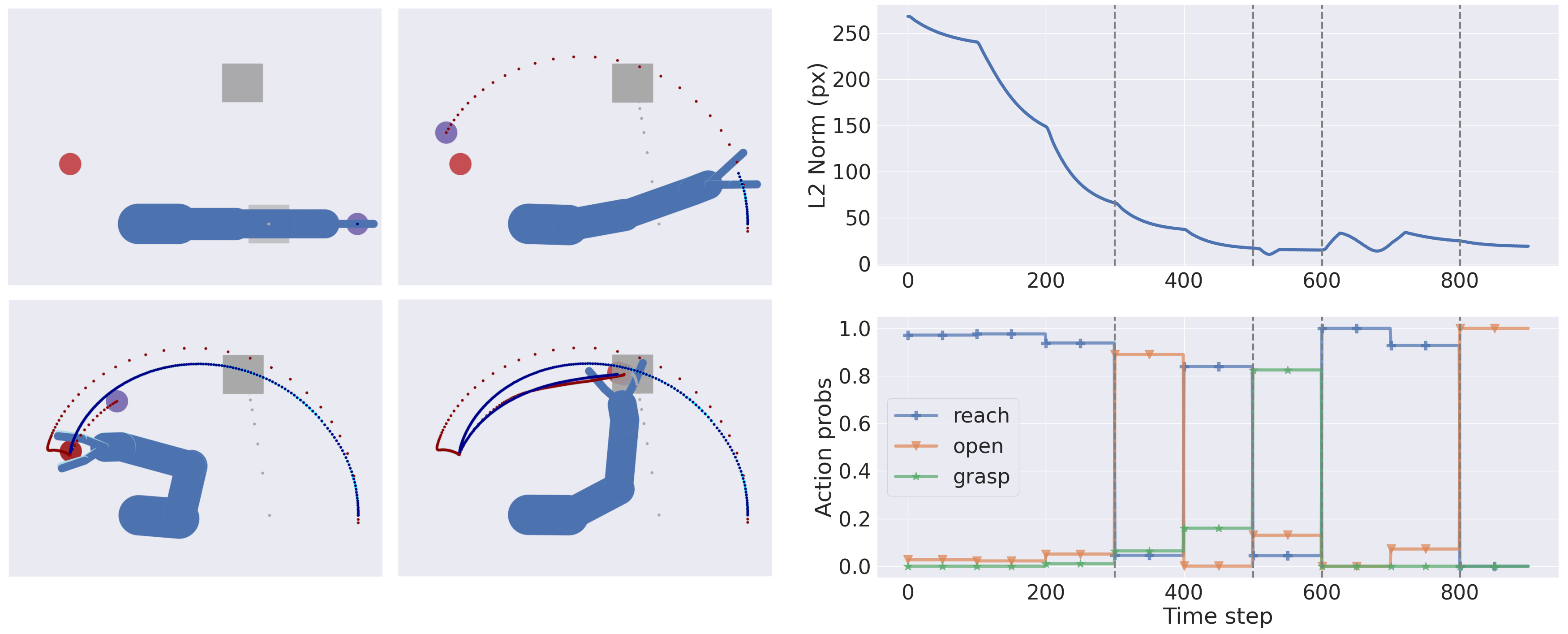}
		\caption{In this task, the agent (a 4-DoF arm with an additional 4-DoF hand composed of two fingers) has to pick a moving object, and place it at a goal position. The agent has a shallow structure with a single IE module computing the hand position from the arm joint angles. Besides the spatial dimension, the agent can easily integrate multisensory information such as touch, dimensions of objects, and other more abstract properties. Note that the object belief is rapidly inferred, and as soon as the picking action is complete, the belief is gradually pulled toward the goal position, resulting in a second reaching movement. The top right panel shows the hand-object distance over time, while the bottom right panel displays the dynamics of the discrete action probabilities used to infer the next discrete state. See \cite{Priorelli2023d} for more details.}
		\label{fig:example_grasping}
	\end{figure}	
	
	In addition to the previous agents, it is now possible to synchronize the behavior of different continuous signals based on the same high-level policy. For instance, one can realize a pick-and-place operation with a moving object -- as represented in Figure \ref{fig:example_grasping} -- producing smooth transitions between reaching and grasping actions, respectively performed in extrinsic and intrinsic domains \cite{Priorelli2023d}. Note how an intermediate phase between the two actions naturally arises, corresponding to a composite approaching movement. In principle, the learning of dynamics precisions $\bm{\Pi}_{x,m}$ might shed light on how motor skill learning occurs, via message passing between continuous trajectories and discrete policies.	Moreover, through this kind of dynamic planning the agent can infer and realize instantaneous trajectories even for the same continuous period $T$ within a discrete step $\tau$, useful, e.g., for grasping the moving object without waiting for the successive replanning step. Third, this configuration permits learning a discrete representation of the environment based on continuous evidence: as shown in \cite{Priorelli2024c}, learning the likelihood matrix $\bm{A}$ involves counting the coincidences between targets and body movements, thus adapting a response strategy (risky versus conservative) depending on the difficulty of the context. Additionally, through learning of the prior $\bm{D}$, a habitual behavior toward the chosen decision emerges. Similarly, one could update preferred states encoded in the matrix $\bm{C}$ based on current observations and actions.
	
	Since the agent plans with trajectories and not positions, to correctly maintain a goal state we need to introduce a hidden cause loosely corresponding to the \textit{stay} action commonly used in discrete tasks \cite{Smith2022}. This hidden cause can be linked to an identity map, i.e., $\bm{i}_{stay}(\bm{x}) = \bm{x}$, which can be interpreted as the agent's desire to maintain the current state of the world \cite{Priorelli2023e}. This hidden cause also relates to the initial state of the task, and translates into a phase of pure perceptual inference.
	
	
	\subsection{Flexible hierarchies}
	
\begin{figure}[h]
		\centering
		\includegraphics[width=0.8\textwidth]{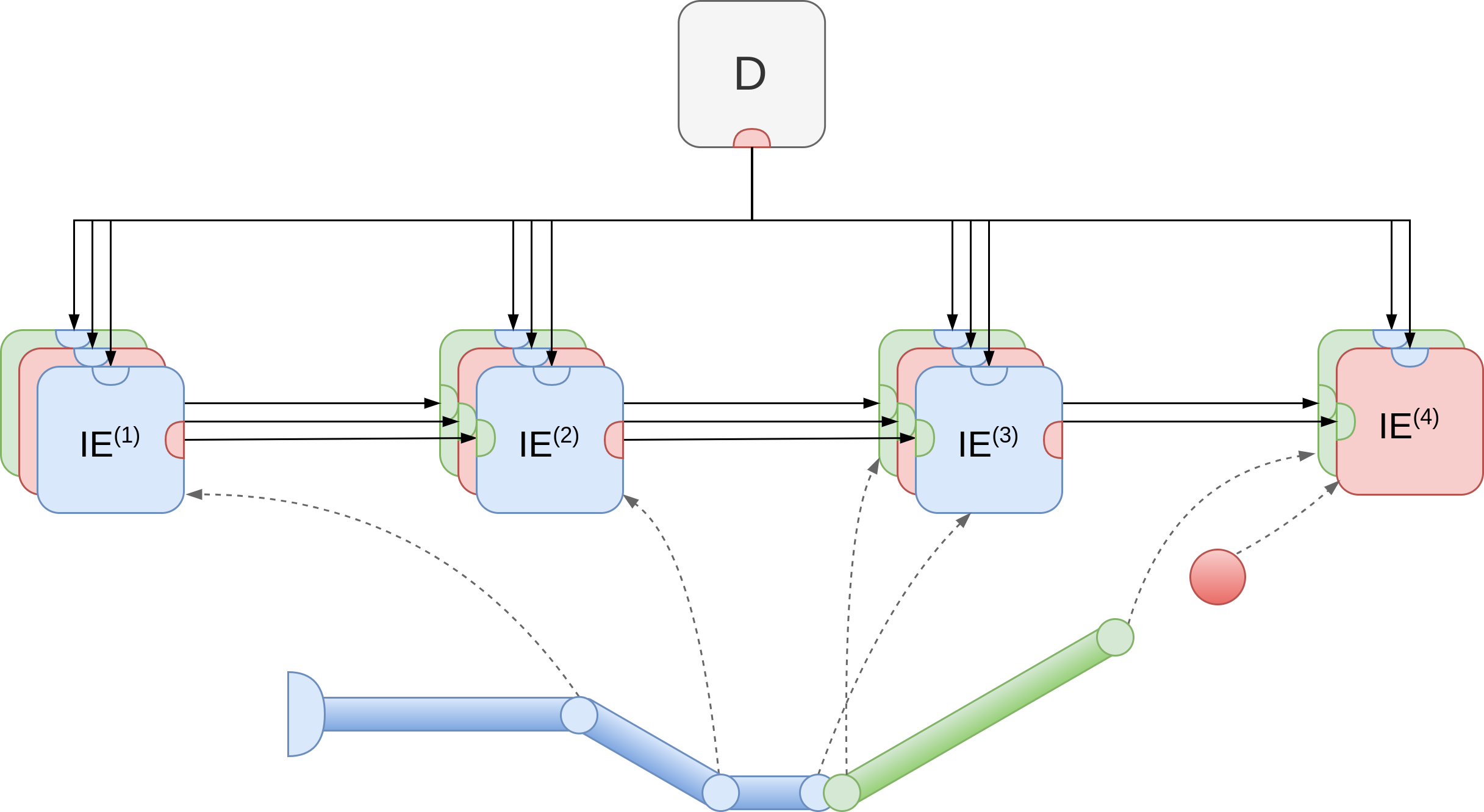}
		\caption{Graphical representation of a deep hybrid model for tool use, composed of a discrete model at the top and several IE modules. Every module is factorized into three elements, which are respectively linked to the observations of the agent's arm (in blue), a tool (in green), and a ball (in red). Note that the last level only encodes the tool's extremity and the ball.}
		\label{fig:hhm}
	\end{figure}

	Figure \ref{fig:hhm} portrays a \textit{deep hybrid model} designed for solving a tool-use task \cite{Priorelli2024a}. It combines the expressivity of a (deep) hierarchical formulation, the advantages of planning trajectories inherent to a hybrid unit, and the possibility of encoding object affordances and other agents. As in Figure \ref{fig:interface}, the IE modules communicate with a discrete model at the top, but now they are combined in a hierarchical fashion recapitulating the agent's kinematic chain. As a consequence, two different goal-directed strategies arise. Considering a simple reaching movement, an attractor imposed at the hand level would generate a cascade of extrinsic prediction errors flowing back to the previous levels and finding a suitable kinematic configuration with the hand over the target. This corresponds to a \textit{horizontal} hierarchical depth occurring along the hybrid units, and can be compared to the process of motor babbling typical of infants \cite{Caligiore2008UsingMB}, whereby random attractors are generated at different hierarchical levels to identify the correct body structure. In addition to this naive strategy, since a discrete model can now generate trajectories for every IE module (in both intrinsic and extrinsic domains), a more advanced behavior can be achieved once inverse kinematics is correctly performed, which imposes a specific path to the whole kinematic chain. This corresponds to a \textit{vertical} hierarchical depth with two (discrete and continuous) temporal scales, steering the lower-level inference in a direction that, e.g., avoids singularities or gets out from local minima generated by repulsive attractors.
	
	Considering the update rule for the (extrinsic) hybrid unit at level $i$:
	\begin{equation}
		\dot{\bm{\mu}}_{x,e}^{(i)} \propto - \bm{\Pi}_{o,e}^{(i-1)} \bm{\varepsilon}_{o,e}^{(i-1)} + \partial \bm{g}_e^T \bm{\Pi}_{o,e}^{(i)} \bm{\varepsilon}_{o,e}^{(i)} + \partial \bm{\eta}^{\prime (i)T}_{x,e} \bm{\Pi}_{x,e}^{(i)} \bm{\varepsilon}_{x,e}^{(i)}
	\end{equation}
	where $\partial \bm{\eta}^{\prime (i)T}_{x,e}$ is the gradient of the trajectory prior of Equation \ref{eq:traj_prior}, we note a delicate balance between forward and backward extrinsic likelihood, and the top-down modulation of the discrete model. From the discrete model's perspective, the discrete hidden states produce a specific combination of hidden causes for each hybrid unit; this combination generates a composite trajectory in the continuous domain weighting distinct potential trajectories, taking into account dynamic elements for the whole discrete step $\tau$. After this period, evidence is accumulated for every hybrid unit, eventually inferring the most likely discrete state that may have generated the actual trajectory, related to the self and the environment.
	
\begin{figure}
		\centering
		\includegraphics[width=0.85\textwidth]{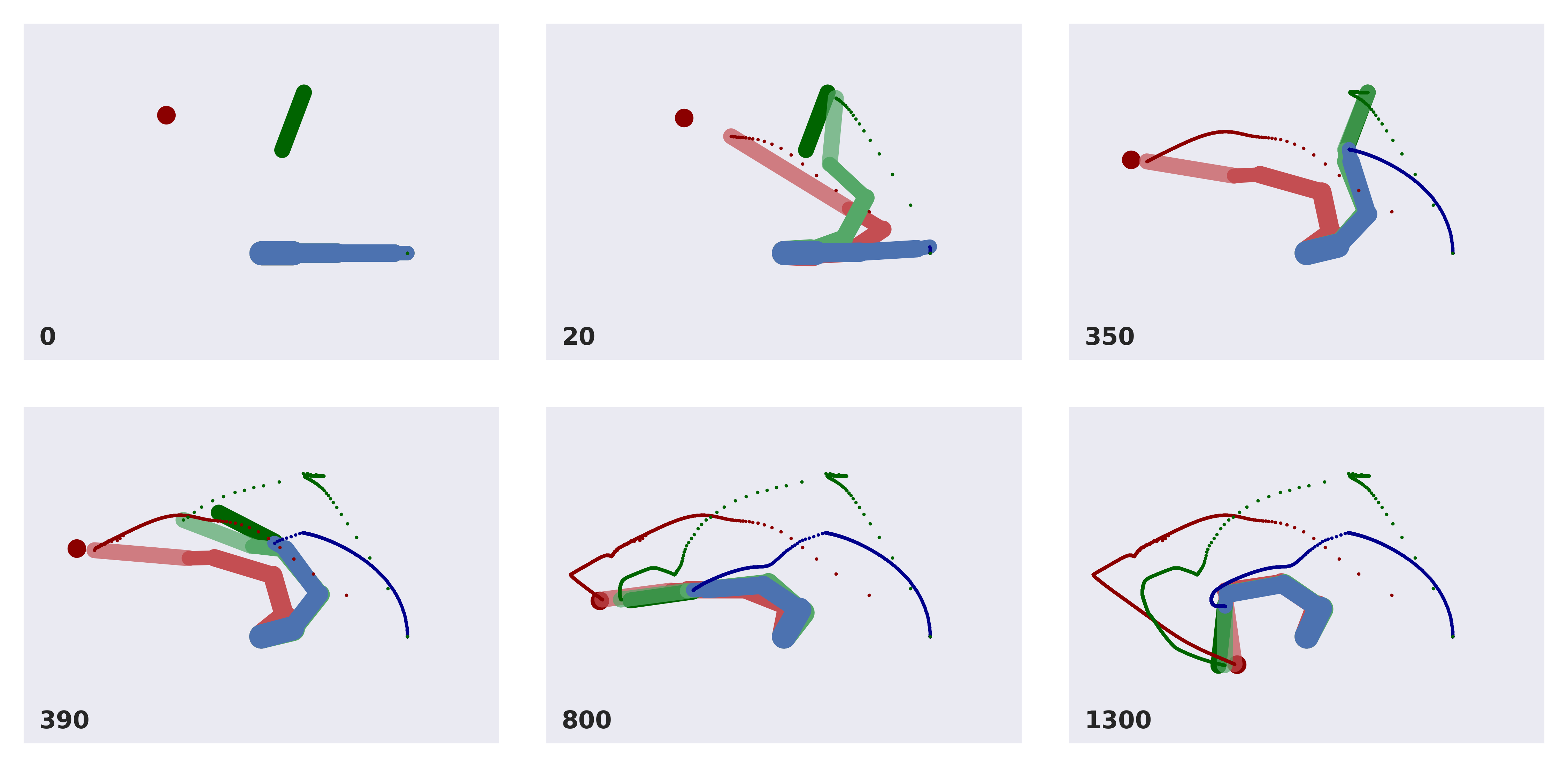}
		\caption{In this task, the agent (a 4-DoF arm) has to grasp a green tool and track a moving red ball with the tool's extremity. The real arm configuration is represented in blue, while the light green and light red arms correspond to the potential agent's configurations in relation to the tool and the ball, respectively. Note that the last two levels of the tool's belief gradually match the real tool, while the ball's belief makes no distinction between the arm and the tool, and is only defined by the tool's extremity. The (deep) hierarchical factorization allows the agent to infer a potential configuration for the ball even during the first reaching movement. See \cite{Priorelli2024a} for more details.}
		\label{fig:example_tool_use}
	\end{figure}

	A non-trivial issue exists in tasks requiring tool use, e.g., reaching a ball with the extremity of a stick. Much as other agents may have different kinematic structures than the self, a tool may have its own hierarchy (e.g., even a simple stick is represented by two Cartesian positions and an angle) that must somehow be integrated into the agent's generative model. Specifically, reaching an object with the extremity of a tool means defining a potential body configuration augmented by a new \textit{virtual} level. This new level does not exist in the generative process and the belief of the actual body configuration $\bm{x}_0$. Nonetheless, if we consider the two potential configurations, the agent thinks of the tool as an extension of its arm, hence flexibly modifying its body schema as discussed in the Introduction. This is possible by linking the two visual observations of the tool to the hand and virtual levels in a second pathway of the hidden states, as shown in Figure \ref{fig:hhm}. Since the intrinsic units of the IE modules also encode information about limb lengths, the agent can not only infer its kinematic structure through visual observations, but also the actual length of the tool \cite{Priorelli2023f}. While this second pathway is still marked by a clear distinction between the tool and the arm since the hand level receives observations from both elements, a third pathway is constructed such that the observation of the ball is only linked to the virtual level. As a result, this new potential configuration views the arm and the tool as part of the same kinematic chain. The interactions between these three pathways (shown in Figure \ref{fig:example_tool_use}) may shed light on how the remapping of the motor cortex gradually occurs with extensive tool use \cite{lriki1996,Obayashi2001}, modifying the boundaries between the self and the environment.
	
	\section{Discussion}
	
	Despite the many advances that have been made in active inference, with increasing popularity among different scientific domains, a current drawback is that studies about motor control and decision-making have been somewhat separated so far, making use of two highly similar but distinct frameworks. As a result, there is no consensus on how to achieve dynamic planning (i.e., how to perform decision-making in constantly changing environments), and state-of-the-art solutions to tackle complex tasks generally couple active inference with traditional methods in machine learning or optimal control. From a theoretical perspective, a few works prescribed an efficient and elegant way for combining the capabilities of discrete and continuous representations into a single generative model \cite{Friston2017,Friston2017a}; however, this hybrid approach has not reached as much maturity, with the consequence that there are far fewer studies on the subject in the literature.
	
	For this reason, we tried here to give a comprehensive view of this yet unexplored direction, comparing several design choices regarding tasks of increasing complexity, with the intent of bringing motor control and behavioral studies closer. As a practical example, we described the modeling of tool use \cite{Priorelli2024a}, a task that inevitably calls for both discrete and continuous frameworks, and that requires taking two additional aspects into account, i.e., object affordances and hierarchical relationships. In a simple scenario, considering a target to reach as the cause of some hidden states is a reasonable assumption and permits operating in dynamic contexts. But when there are multiple objects, how does the agent decide what the cause of a particular action will be? How can it account for different object affordances? And what if the target moves along a non-trivial path? The hidden states may be factorized into independent distributions encoding multiple entities in intrinsic coordinates, hence expressing \textit{potential body configurations}. Further, the hidden causes may be linked to \textit{potential trajectories} related to the agent's intentions \cite{Priorelli2023, Priorelli2023b, Priorelli2023d}. Beliefs over each entity would have their own dynamics, allowing the agent to predict, e.g., the trajectory of a moving ball. Then, we described how this unit could be scaled up to construct complex (deep) hierarchical structures, e.g., for simulating human body kinematics \cite{Priorelli2023b}, and to perform more general transformations of reference frames, e.g., perspective projections \cite{Priorelli2023c}. A hierarchical factorization of the hidden states now assumes a broader perspective that can also account for multi-agent interaction -- an aspect that has been analyzed in the discrete framework as well \cite{10288543}. Finally, we described the design of a hybrid unit with discrete hidden causes and continuous hidden states, affording dynamic inference via Bayesian model reduction \cite{Priorelli2023e}; then, a higher-level discrete model made it possible to simulate dynamic tasks involving online planning of actions. This showed further parallelisms between the inference of trajectories in continuous models and policies in discrete models.
	
	One challenge in maintaining a deep representation of the kinematic chain is the associated computational complexity and the time required to infer a body posture from visual input, both increasing with the DoF. This is because an extrinsic prediction error generated at the distal (e.g., hand) levels would have to climb the hierarchy toward the (root) body-centered reference frame. This directly affects behavioral accuracy and movement time, which critically depend on the correct inference of the intrinsic state. A performance comparison during inference and action is shown in \cite{Priorelli2023b}, for increasing DoF. In contrast, handling multiple objects does not lead to a significant increase in inference time, since the (deep) representations are computed in parallel, and are only limited by the complexity of the object dynamics. Communication with the discrete model is equally efficient, as a single discrete state can infer the body trajectories based on multiple units concurrently \cite{Priorelli2023e}. Still, the behavior of other objects in the environment requires predicting not only how the agent will behave after a specific action but also how the other objects will behave and their influence on the agent's behavior. This requires a rich discrete representation that can model the interactions of every object, and evaluating complex policies may not be as effective for high-dimensional settings \cite{Paul2024}. A last issue regards the time $T$ needed for accumulating continuous evidence for a single discrete step $\tau$. As shown in \cite{Priorelli2023d}, Figure 5b, a narrow sampling time interval permits efficient control in highly dynamic environments but at the cost of increased computational time.
	
	A limitation of the models reviewed here is their fixed structure. The critical question, then, is how tool use can be realized without embedding prior knowledge into the agent’s generative model. In other words, how an agent can perform using active inference when starting from a blank memory or under the assumption that the surrounding environment will remain unchanged. A common criticism of continuous-time active inference models is that their generative models are a-priori defined and fixed, with intricate and hardcoded dynamics functions that raise some concerns about biological plausibility. In contrast, one appealing characteristic of PCNs is that they simulate brain processing with extremely simple functions typical of the connectivity of neural networks (e.g., linear combinations of weights and biases passed to a non-linear activation function). This allows PCNs to easily adapt to high-dimensional data, with a few critical advantages compared to deep learning (e.g., about top-down modulation) \cite{millidge2022predictive}. While much of the research with PCNs involves static representations, some studies addressed how predictive coding could be used to learn temporal sequences \cite{Jiang2022,Millidge2023}, or to solve RL tasks \cite{Ororbia2022, rao2022active, Millidge2019}. Here, we demonstrated how generative models in active inference could be realized by simple likelihood and dynamics functions, showing some analogies with the inference of PCNs. Based on these findings, a promising research direction would be to imitate their (deep) hierarchical architectures (as in Figure \ref{fig:hierarchies_new}), so that an agent could not only flexibly adapt its body schema to interact with objects with different hierarchical structures, but also learn the system dynamics and act over them to conform to prior beliefs.
	
	Learning policies in continuous environments is not an easy challenge, but addressing it with strategies different from traditional methods might be key for advancing with current intelligent agents, realizing the full theoretical potential at the basis of active inference and the free energy principle. On this matter, the state-of-the-art is to approximate the likelihood and transition distributions by deep neural networks \cite{Ueltzhoffer2017,atal2019BayesianPS,Ferraro2023,Yuan2023, Millidge2020, Fountas2020, champion2023deconstructing, 9514227}. While several benefits arise compared to deep RL, this still relegates the deep structure within the neural network, generally using a single-level active inference agent. One study used a more biologically plausible PCN as a generative model \cite{Millidge2019}, but relied on a similar approach. As extensively analyzed in \cite{Friston2008}, neural networks can be seen as static generative models with infinitely precise priors at the last level and no hidden states. This architecture can be used to perform sparse coding or Principal Components Analysis (PCA); however, it fails to account for dynamic variables, as in deconvolution problems or filtering in state-space models. Temporal depth -- either discrete or continuous -- is thus key to inferring the most accurate representation of the environment, and indeed it seems that cortical columns are able to express model dynamics (e.g., the prefrontal cortex is constantly involved in predicting future states, and motion-sensitive neurons have been recorded in the early visual cortex as well \cite{Grossberg2008}). While it is true that temporal sequences can be easily handled by deep architectures such as recurrent neural networks or transformers \cite{NIPS2017_3f5ee243}, their passive generative mechanism could still be reflected in the behavior of the active inference agent. In contrast to such a passive AI, being grounded on sensorimotor experiences and actively modifying the environment could be fundamental to the emergence of genuine understanding \cite{Pezzulo2024}. Taken together, these facts suggest that a hierarchy of actions upon generalized coordinates of motion or discrete future states could bring several advantages in solving RL tasks. For instance, representing an agent in a hierarchical fashion afforded highly advanced control over its whole body structure that would not have been possible by a single level generating only the hand position \cite{Priorelli2023b, Priorelli2023f}.
	
	How to learn dynamic planning in deep hierarchical models? The importance of being discrete when considering structure learning has been stressed in \cite{friston2023supervised}. Indeed, hierarchical discrete models afford much more expressivity compared to their continuous counterparts, above all, deriving from the simplicity of computing the expected free energy. Nonetheless, as Friston and colleagues note, whether using continuous or discrete representations depends on the model evidence. Specifically, the former may have better performances when the evidence has contiguity properties, e.g., when dealing with time series or with Euclidean space. Indeed, the task exemplified in Figure \ref{fig:example_tool_use} is effective because the Bayesian model reduction performs a dynamic evidence accumulation over the extrinsic space in which the agent operates. Hence, coupling the hierarchical depth of the hybrid units in Figure \ref{fig:hhm} with a hierarchical discrete architecture (and not just a single discrete level) could bring efficient structure learning also in constantly changing environments. A successful Bayesian approach to learning discrete structures involves growing discrete distributions using an infinite Dirichlet process and the Chinese restaurant prior \cite{Sanborn2010}. This method assumes a potentially infinite mixture of basis distributions and builds a structure starting from an empty model wherein novel configurations are either assigned to popular existing states, or occasionally, used to create novel states. This approach has shown success in learning structures that support complex goal-directed behavior \cite{Stoianov2016}, hierarchical spatial organization \cite{Stoianov2022}, and spatial navigation \cite{Stoianov2018a} among other applications. An alternative to hierarchical discrete models would be to combine units composed of a joint discrete-continuous model -- as in Figure \ref{fig:interface} -- which would allow to perform dynamic planning within each single unit. While this solution may not be supported by empirical evidence from biological agents, it could be an encouraging direction to explore from a machine learning perspective, contrasting the hypothesis of central discrete decision-making with a distributed network of local decisions.
	
	A third interesting topic regards motor intentionality. Although multi-step tasks are typically tackled at the discrete level, we showed here that, under appropriate assumptions, a non-trivial behavior could be achieved and analyzed also at the continuous level. The flexible intentions we defined could be compared to an advanced stage of motor skill learning, consisting of autonomous and smooth movements that do not necessitate conscious decision-making \cite{Priorelli2023d}. Still, even in this case the model structure was predefined. How do such intentions emerge during repeated exposure to the same task? How does the agent score which intentions will be appropriate for a specific context? As mentioned in the last chapter, optimization of dynamics precisions is likely to involve the free energy of reduced models (see Equation \ref{eq:log_ev}). This process may shed light on how discrete actions arise from low-level continuous trajectories and, conversely, how the latter are generated from a composite discrete action. Last, a few studies proposed additional connections between policies unfolding at different timescales, either directly \cite{Toon2024,deTinguy2024} or through discrete hidden states \cite{FRISTON2024105500}. Such approaches could be adopted in hybrid and continuous contexts as well, so that flexible intentions could be propagated via local message passing between hidden causes along the whole hierarchy.
	
	
	\section*{Data availability}
	
	Code and data are deposited in GitHub (https://github.com/priorelli/dynamic-planning).
	
	
	\section*{Acknowledgments}
	
	This research received funding from the European Union’s Horizon H2020-EIC-FETPROACT-2019 Programme for Research and Innovation under Grant Agreement 951910 to I.P.S. The funders had no role in study design, data collection and analysis, decision to publish, or preparation of the manuscript.
	
	
	\bibliographystyle{unsrt}  
	\bibliography{references}
	
\end{document}